\documentclass{article}
\usepackage[a4paper, total={6in, 8in}]{geometry}

\RequirePackage{amsthm,amsmath,amsfonts,amssymb}
\RequirePackage[authoryear]{natbib}
\RequirePackage[colorlinks,citecolor=blue,urlcolor=blue,backref=page,backref=page]{hyperref}
\RequirePackage{graphicx}

\usepackage{todonotes}
\usepackage{bm}
\usepackage{graphicx}
\usepackage{subcaption}
\usepackage{mathtools} 
\usepackage{booktabs} 
\usepackage{tikz} 
\usepackage[ruled]{algorithm2e}
\usepackage{amssymb,amsthm}
\usepackage{amsmath}
\usepackage[noend]{algpseudocode}
\usepackage{enumitem}
\usetikzlibrary{shapes.geometric, arrows}

\def\startlocaldefs{\makeatletter}
\def\endlocaldefs{\makeatother}

\startlocaldefs

\usepackage{xspace}
\usepackage{arydshln}
\usepackage{xurl}

\newcommand{\ignore}[1]{}

\newcommand{\x}{\mathbf{x}}

\newcommand{\xx}{\mathbf{X}}
\newcommand{\f}{\mathbf{f}}
\newcommand{\m}{\mathbf{m}}
\renewcommand{\u}{\mathbf{u}}
\newcommand{\s}{\mathbf{s}}
\newcommand{\y}{\mathbf{y}}
\newcommand{\z}{\mathbf{z}}
\newcommand{\zz}{\mathbf{Z}}
\newcommand{\mmu}{\bm{\mu}}
\newcommand{\vmu}{\bm{\mu}}

\newcommand{\X}{\mathcal{X}}

\newcommand{\Q}{\mathbf{Q}}

\newcommand{\K}{\mathbf{K}}
\renewcommand{\S}{\mathbf{S}}
\newcommand{\nparams}{D}
\newcommand{\hyp}{\bm{\xi}}

\newcommand{\gskl}{GsKL\xspace}

\newcommand{\normpdf}[3]{\mathcal{N}\left({#1}; {#2}, {#3 } \right)}

\newcommand{\erf}{\text{erf}}
\newcommand{\data}{\mathcal{D}}
\newcommand{\vtheta}{\bm{\theta}}

\newcommand{\like}{p(\data|\x)}
\newcommand{\post}{p(\x|\data)}

\newcommand{\qparams}{\bm{\phi}}

\newcommand{\qp}{q_{\qparams}}
\newcommand{\qx}{q(\x)}
\newcommand{\qpx}{q_{\qparams}(\x)}

\newcommand{\vgp}{\bm{\psi}}

\newcommand{\VBMC}{\textsc{vbmc}\xspace}
\newcommand{\SGPR}{\textsc{sgpr}\xspace}
\newcommand{\SVGP}{\textsc{svgp}\xspace}

\newcommand{\VSBQ}{\textsc{vsbq}\xspace}
\newcommand{\BBVI}{\textsc{bbvi}\xspace}
\newcommand{\ADVI}{\textsc{advi}\xspace}
\newcommand{\NNR}{\textsc{nnr}\xspace}
\newcommand{\LA}{\textsc{laplace}\xspace}

\newcommand{\MCMC}{\textsc{mcmc}\xspace}

\newcommand{\SVI}{\textsc{svi}\xspace}
\newcommand{\cmaes}{\textsc{cma-es}\xspace}

\newcommand{\ELBO}{\textsc{elbo}\xspace}
\newcommand{\bads}{\textsc{bads}\xspace}

\newcommand{\MMTV}{MMTV\xspace}

\DeclareMathOperator{\Tr}{Tr}

\newcommand{\NA}{N/A}

\algnewcommand\algorithmicinput{\textbf{Input:}}
\algnewcommand\algorithmicoutput{\textbf{Output:}}
\algnewcommand\INPUT{\item[\algorithmicinput]}
\algnewcommand\OUTPUT{\item[\algorithmicoutput]}
\algloop{Do}
\algblock{ParFor}{EndParFor}
\algnewcommand\algorithmicparfor{\textbf{parfor}}
\algnewcommand\algorithmicpardo{\textbf{do}}
\algnewcommand\algorithmicendparfor{\textbf{end\ parfor}}
\algrenewtext{ParFor}[1]{\algorithmicparfor\ #1\ \algorithmicpardo}
\algrenewtext{EndParFor}{\algorithmicendparfor}

\newtheorem{theorem}{Theorem}[section]

\newtheorem*{definition*}{Definition}

\newtheorem{lemma}[theorem]{Lemma}

\newtheorem*{remark*}{Remark}

\newtheorem*{remarks*}{Remarks}

\newtheorem*{notation*}{Notation}

\newtheorem*{ex*}{Example}

\newtheorem*{exs*}{Examples}

\newtheorem*{app*}{Application}
\newtheorem{conjecture*}{Conjecture}
\endlocaldefs

\title{Fast post-process Bayesian inference with \\ 
Variational Sparse Bayesian Quadrature}

\author{Chengkun Li$^{\dag}$, Grégoire Clarté$^{\ddag}$, Martin Jørgensen$^{\dag}$, Luigi Acerbi$^{\dag}$}

\date{
    $^\dag$Department of Computer Science, University of Helsinki, \texttt{chengkun.li@helsinki.fi, martin.jorgensen@helsinki.fi
    luigi.acerbi@helsinki.fi} \\
    $^\ddag$ Department of Statistics, University of Edinburgh, \texttt{gclarte@ed.ac.uk}\\
}

\begin{document}
\maketitle

\begin{abstract}
In applied Bayesian inference scenarios, users may have access to a large number of pre-existing model evaluations, for example from maximum-a-posteriori (MAP) optimization runs. However, traditional approximate inference techniques make little to no use of this available information. We propose the framework of \emph{post-process Bayesian inference} as a means to obtain a quick posterior approximation from existing target density evaluations, with no further model calls.
Within this framework, we introduce Variational Sparse Bayesian Quadrature (\VSBQ), a method for post-process approximate inference for models with \emph{black-box} and potentially noisy likelihoods. \VSBQ reuses existing target density evaluations to build a sparse Gaussian process (GP) surrogate model of the log posterior density function. Subsequently, we leverage sparse-GP Bayesian quadrature combined with variational inference to achieve fast approximate posterior inference over the surrogate. We validate our method on challenging synthetic scenarios and real-world applications from computational neuroscience. The experiments show that \VSBQ builds high-quality posterior approximations by post-processing existing optimization traces, with no further model evaluations.\\

\end{abstract}
\section{Introduction}
\label{sec:intro}
Bayesian inference is a well-founded approach to uncertainty quantification and model selection, widely adopted in data science and machine learning \citep{robert2007bayesian,gelman2013bayesian,ghahramani2015probabilistic}. 
Key quantities in Bayesian inference are the \emph{posterior distribution} of model parameters, useful for parameter estimation and uncertainty quantification, and the \emph{marginal likelihood} or model evidence, useful for model selection. 
In practice, Bayesian inference is particularly challenging when dealing with models with `black-box' features common in science and engineering, such as lack of gradients or mildly-to-very expensive and possibly noisy evaluations of the likelihood, e.g., arising from simulation-based estimation \citep{diggleMonteCarloMethods1984,wood2010statistical,van2020unbiased,priceBayesianSyntheticLikelihood2018}.

Due to the cost of inference, parameter estimation and the workflow of Bayesian analyses often start with a preliminary exploration phase via simpler and cheaper means \citep{gelman2020bayesian}. A popular choice consists of performing maximum likelihood estimation (MLE) or maximum a posteriori (MAP) estimation~\citep[Chapter~13]{gelman2013bayesian}, i.e., finding the (global) \emph{mode} of the posterior density,\footnote{For a fixed parameterization, MLE can be viewed as MAP estimation with (improper) uniform/flat priors.} often via multiple runs of black-box optimization algorithms (e.g., \citealp{hansen2003reducing, acerbi2017practical}), which can easily require tens of thousands likelihood or posterior density evaluations over distinct optimization runs to satisfactorily explore the parameter landscape (e.g., \citealp{acerbi2018bayesian,norton2019human, cao2019causal, zhou2020role, yoo2021uncertainty,heald2021contextual}). In fact, due to prohibitive costs, many analyses stop here, with a point estimate instead of a full posterior, as advocated for example in modeling tutorials and textbooks in applied domains such as computational and cognitive neuroscience \citep{wilson2019ten, ma2023bayesian}.

%

In this paper, we propose the framework of \emph{post-process Bayesian inference} as a solution to the waste of potentially expensive likelihood and posterior density evaluations, with the goal of making `black-box' Bayesian inference cheaper and more widely applicable by modeling practitioners.
Namely, we aim to build an approximation of the full Bayesian posterior by recycling \emph{all} previous evaluations of the posterior density achieved by various means. In principle, our proposed method has no restrictions for the source of evaluations, as long as the evaluation points form a representative set of the underlying posterior. In this paper, we focus on re-utilizing evaluations obtained from MAP optimization runs of black-box optimization methods such as \cmaes \citep{hansenCMAEvolutionStrategy2016} and \bads \citep{acerbi2017practical,singh2024pybads}. We chose these methods for their popularity among practitioners, available implementations in multiple programming languages (e.g., MATLAB and Python), as well as their capability of robustly handling both exact and noisy objective functions.

As an instantiation of our framework, we introduce Variational Sparse Bayesian Quadrature (\VSBQ). \VSBQ builds a Gaussian process (GP; \citealp{rasmussen2006gaussian}) surrogate of the log density, starting from existing log-likelihood or log-density evaluations. In particular, we use a \emph{sparse} GP to deal with a potentially large number of model evaluations \citep{titsias2009variational} and develop the \emph{noise shaping} technique for efficient posterior modeling. A tractable posterior approximation is then obtained by performing variational inference over the surrogate, thus without further evaluations of the original model. Fitting the flexible variational posterior is particularly efficient by utilizing Bayesian quadrature \citep{ohagan1991bayes, acerbi2018variational}. We validate our approach on synthetic targets and real datasets and models from computational neuroscience \citep{acerbi2012internal, acerbi2018bayesian}.
Overall, we find that post-process Bayesian inference via \VSBQ is not only feasible but can yield high-quality approximations, providing the applied modeler with a new approximate inference tool that can easily fit in existing modeling pipelines as a refinement step. At the end of the paper, we discuss the limitations of the method and future work.

\subsection{Related work}

There are several works related to post-processing existing density evaluations to construct an approximate posterior. \citet{zhangPathfinderParallelQuasiNewton2022} find the best multivariate normal approximation along the optimization paths generated by a quasi-Newton optimization algorithm, L-BFGS~\citep{liuLimitedMemoryBFGS1989},  in terms of Kullback-Leibler (KL) divergence to the true posterior. Unlike our method, the L-BFGS algorithm requires the gradients of the log-likelihood or log-density function to build approximate estimates of the Hessian along the optimization trajectory and is brittle to noise in the target function. 
More closely aligned with our approach, \citet{bliznyukBayesianCalibrationUncertainty2008} locates the high posterior density region through derivative-free MAP optimization and augments the evaluation set with additional design points to build a surrogate of the log density using radial basis functions. Both methods above require additional evaluations of the log-density function. \citet{yao2022stacking} propose to reuse parallel -- and possibly incomplete -- runs of different inference algorithms such as Markov Chain Monte Carlo (\MCMC; \citealp[see e.g.][]{robertMonteCarloStatistical2004}) or variational inference \citep{blei2017variational} by combining them in a weighted average via `Bayesian stacking'. Similar in spirit to our approach, the computation of the stacking weights only requires a post-processing step. However, Bayesian stacking requires the samples to be approximately drawn from the posterior and does not make use of all available target evaluations.

Our work also connects to simulation-based inference (SBI), a broad framework for estimating posteriors when likelihoods are intractable or computationally expensive  \citep{diggleMonteCarloMethods1984,cranmerFrontierSimulationbasedInference2020}. SBI methods include classical approaches such as rejection sampling \citep{sissonOverviewABC2018}, parametric and nonparametric likelihood approximations \citep{diggleMonteCarloMethods1984, price2018bayesian, gutmannBayesianOptimizationLikelihoodFree2016}, and modern neural density estimation techniques \citep{lueckmannBenchmarkingSimulationbasedInference2021, radevBayesFlowLearningComplex2020}. Our approach is distinct in that it post-processes a fixed set of (noisy) log-density evaluations to estimate a single posterior, without requiring additional simulations. Related to the post-processing idea, \citet{yaoSimulationBasedStacking2024} proposes to improve posterior approximations via stacking multiple SBI results.

Finally, a related technique that merits mention is \emph{offline} black-box optimization \citep{krishnamoorthyDiffusionModelsBlackBox2023, trabuccoDesignBenchBenchmarksDataDriven2022a}. In offline black-box optimization, the objective is to identify a parameter input that maximizes a black-box function utilizing pre-existing offline function evaluations. Much like our approach, offline black-box optimization capitalizes on leveraging existing evaluations. However, a key distinction lies in our primary objective, which is to construct an approximation of the unknown posterior density, a task inherently more challenging than finding a point estimate.

\subsection{Outline}

We first recap the essential background methods---(sparse) Gaussian processes (GPs), variational inference, and Bayesian quadrature, in Section \ref{sec:background}. In Section \ref{sec:vsbq}, we describe the details of our proposed framework for post-process inference, Variational Sparse Bayesian Quadrature (\VSBQ). In the process, we introduce a simple and principled heuristic to make the sparse GP representation focus on regions of interest, \emph{noise shaping} (Section \ref{sec:noiseshaping}). Section \ref{sec:experiments} validates \VSBQ on challenging synthetic and real-world examples. Finally, Section \ref{sec:discussion} discusses the strengths and limitations of our approach. Proofs, implementation details, additional results, and extended explanations can be found in the Supplementary Material.

\section{Background}
\label{sec:background}

In this section, we present the core concepts and techniques used in the paper. We recall that our objective is to compute a tractable approximation of the posterior density, given the evaluations (observations) of the unnormalized log-density function. We provide an overview of key techniques, including the variational inference method for posterior approximation, the Gaussian process as a regression surrogate,  Bayesian quadrature, Variational Bayesian Monte Carlo \citep{acerbi2018variational}, and sparse Gaussian processes.

\subsection{Notation}
Throughout the paper, we denote with $f_0(\x) \equiv \log p(\data | \x) p(\x)$ the target (unnormalized) log posterior density or log joint, where $p(\data | \x)$ is the likelihood of the model of interest under the data $\data$, $p(\x)$ is the prior, and $\x \in \X \subseteq \mathbb{R}^D$ is a vector of model parameters. The dimension of the parameter space is denoted as $D \in \mathbb{N}$. We indicate with $(\x_n, y_n)$ pairs of observed locations and values of the log-density, i.e., $y_n = f_0(\x_n)$. For noisy evaluations of the target arising from simulation-based estimates, $\sigma_\text{obs}^2(\x_n)$ denotes the variance of the observation, assumed to be known---in practice, it is typically estimated (see, for example, \citealp{acerbi2020variational,jarvenpaa2021parallel}). $\left( \mathbf{X}, \y, \s \right) \equiv \{ \x_n, y_n, \sigma_\text{obs}(\x_n) \}_{n=1}^{N}$ denote the set of observations. We denote with $\S$ the diagonal matrix of observation noise variances, with $\S \equiv \text{diag}\left[ \sigma_\text{obs}^2(\x_1), \ldots, \sigma_\text{obs}^2(\x_N) \right]$.\footnote{Later, this will be the total observation noise, which includes noise shaping introduced in Section~\ref{sec:noiseshaping}.} 

\begin{figure*}[tp]
    \centering
    \includegraphics[]{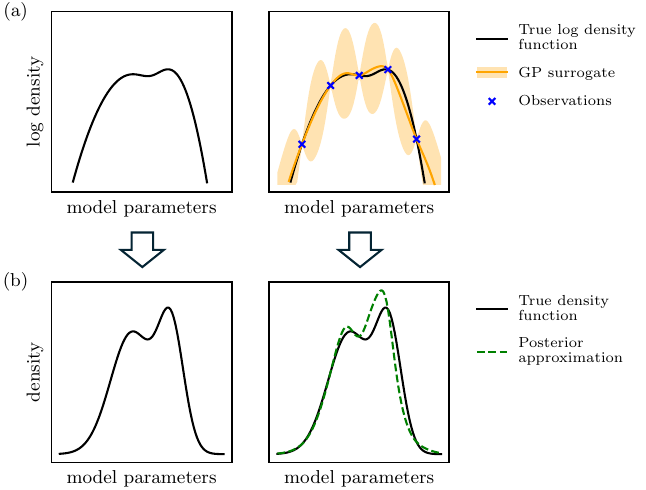}
    \caption{\textbf{Log-density function modeling and posterior approximation.} (a) The left panel depicts the ground-truth (unnormalized) posterior log-density function. The right panel illustrates the Gaussian Process (GP) approximation to the log-density function (mean and 95\% credible interval), given observed log-density evaluations (blue crosses). (b) The left panel depicts the ground-truth posterior density function, corresponding to the log density in (a). The right panel illustrates the posterior density approximation corresponding to the GP surrogate mean.}
    \label{fig:density_modeling}
\end{figure*}

\subsection{Variational inference}
\label{sec:vi}

Variational inference is a popular approach to approximate the intractable posterior density $\post$ via a simpler distribution $\qx \equiv \qpx$ that belongs to a parametric family indexed by $\qparams$ \citep{jordan1999introduction,bishop2006pattern,blei2017variational,kingma2013auto}. 
The goal of variational inference is to find $\qparams$ for which the variational posterior $\qp$ best approximates the true posterior, as quantified by the reverse Kullback-Leibler (KL) divergence,
\begin{equation} \label{eq:KL}
D_{\text{KL}}\left[\qpx || \post\right] = \mathbb{E}_{\qparams} \left[\log \frac{q_{\qparams}(\x)}{\post} \right],
\end{equation}
where $\mathbb{E}_{\qparams} \equiv \mathbb{E}_{\x \sim q_{\qparams}(\x)}$. Crucially, $D_{\text{KL}}(q || p) \ge 0$ and $D_{\text{KL}}(q || p) = 0$ if and only if $q = p$. 
Minimizing the KL divergence casts Bayesian inference as an optimization problem. This optimization consists of finding the variational parameters, $\qparams$, that maximize the objective:
\begin{equation} \label{eq:supp_elbo}
\begin{split}
    \text{\ELBO}\left( q_{\qparams} \right) &=  \mathbb{E}_{\qparams} \left[\log \frac{\like p(\x)}{q_{\qparams}(\x)} \right] \\
    &= \mathbb{E}_{\qparams} \left[f_0(\x) \right] + \mathcal{H}[q_{\qparams}(\x)],
\end{split}
\end{equation}
with $f_0(\x) = \log \like p(\x) = \log p(\data, \x)$ the log joint density, and $\mathcal{H}[q]$ the entropy of $q$.
Eq. \ref{eq:supp_elbo} is the \emph{evidence lower bound} (\ELBO), a lower bound to the log marginal likelihood $\log p(\data)$ (also called model evidence), with equality holding if $\qx = \post$.

Variational inference with a flexible variational family $q_{\qparams}$ can approximate the target arbitrarily well \citep{miller2016variational}, but at the cost of a large number of evaluations of the target -- and its gradient -- to optimize the \ELBO in Eq. \ref{eq:supp_elbo}. This is particularly problematic if the target likelihood or joint density is a black-box function, such that evaluations may be limited due to computational resources and the gradient unavailable.

\subsection{Gaussian processes}
\label{sec:gps}

When a computational model of interest has an expensive black-box likelihood, a proven approach for efficient Bayesian inference consists of building a (cheaper) \emph{surrogate model} to emulate either the log-likelihood $\log p(\data | \x)$ or directly the log joint density function $f_0(\x)$; see Figure~\ref{fig:density_modeling}. There is a long tradition of using Gaussian processes (GPs) as surrogate models for Bayesian inference \citep{rasmussen2003gaussian, gunter2014sampling, gutmann2016bayesian, nemeth2018merging, wang2017adaptive, acerbi2018variational, jarvenpaa2021parallel,desouza2022parallel, gammalFastRobustBayesian2022}. 

GPs are stochastic processes that can be thought of as distributions over functions. We refer the reader to \citet{rasmussen2006gaussian,garnett2023bayesian} for an introduction to GPs in applied machine learning contexts. GPs are determined by a prior mean function $m: \X \rightarrow \mathbb{R}$; a positive definite covariance function $\kappa: \X \times \X \rightarrow \mathbb{R}$ (also called a kernel); and a likelihood or observation model. The mean function represents the average behavior and trend of a GP far from observed data. In the case of log-density modeling, it is both beneficial and necessary to consider specific mean functions other than the usual constant mean function \citep{acerbi2019exploration}. We detail the mean function design in Section \ref{sec:vbmc}.


\paragraph{Observation model}
GPs are characterized by a likelihood or observation noise model, commonly assumed to be Gaussian to afford GP posterior computations in closed form. In this paper, we consider both noiseless and noisy observations of the target log joint. For noiseless targets, we use a Gaussian likelihood with a small variance $\sigma_\text{obs}^2 = 10^{-5}$, also known as \emph{nugget}, for numerical stability \citep{gramacy2012cases}. Noisy observations of the target often arise from using stochastic estimators of the log-likelihood via simulation \citep{wood2010statistical,van2020unbiased}.
 
\paragraph{GP posterior} 
The GP posterior given function observations $(\xx, \y, \s)$ is also a Gaussian process with the posterior mean function $\mu_\text{p}$ and posterior covariance function $\kappa_\text{p}$ \citep{rasmussen2006gaussian}. For $ \tilde{\x}, \tilde{\x}^\prime \in \mathcal{X}$, 
\begin{align}
\mu_\text{p} (\tilde{\x}) &= \kappa(\tilde{\x}, \xx) 
    \left(\kappa(\xx, \xx) + \S\right)^{-1} \left(\y - m(\xx)\right) \nonumber\\
    &\quad + m(\tilde{\x}),
\\
\kappa_\text{p} (\tilde{\x},\tilde{\x}^\prime) &= 
    \kappa(\tilde{\x},\tilde{\x}^\prime) - 
    \kappa(\tilde{\x},\xx)\left(\kappa(\xx, \xx) + \S\right)^{-1}\nonumber \\
    &\quad \times \kappa(\xx,\tilde{\x}^\prime).
\end{align}

\paragraph{Tractable surrogate density approximation}

A key observation is that a stochastic GP surrogate of the log density (Figure~\ref{fig:density_modeling}a) does not immediately yield a usable approximate posterior. The normalization constant is unknown and we cannot directly sample from the density associated with the GP surrogate. Another layer of approximation is needed to go from the log density surrogate to posterior approximation (Figure~\ref{fig:density_modeling}). One straightforward approach is to use the posterior mean function of the GP as a deterministic surrogate for the log density and apply MCMC methods to sample from the resulting approximate posterior~\citep{jarvenpaa2021parallel,gammalFastRobustBayesian2022}. Alternatively, several techniques leverage Bayesian quadrature~\citep{ohagan1991bayes} to solve the integrals involving the stochastic GP surrogate \citep{rasmussenBayesianMonteCarlo2002, osborne2012active, gunter2014sampling, acerbi2018variational, adachi2022fast}, as described next. 

\subsection{Bayesian quadrature}
\label{sec:bq}

Many key computations in Bayesian inference require the estimation of intractable integrals, for example the \ELBO seen in Eq. \ref{eq:supp_elbo}.
Bayesian quadrature \citep{ohagan1991bayes,rasmussenBayesianMonteCarlo2002} is a technique to obtain Bayesian estimates of intractable integrals of the form 
\begin{equation} \label{eq:integral}
\mathcal{J} = \int_{\X} f(\x) \pi(\x) d\x,
\end{equation}
where $f$ is a function of interest and $\pi$ a known probability distribution. Here we consider the domain of integration $\mathcal{X} = \mathbb{R}^\nparams$. 
When a GP prior is specified for $f$, since integration is a linear operator, the integral $\mathcal{J}$ is also a Gaussian random variable whose posterior mean and variance are \citep{rasmussenBayesianMonteCarlo2002}
\begin{align} 
    \mathbb{E}_{f}[\mathcal{J}] &= \int \mu_\text{p}(\x) \pi(\x) d\x, \label{eq:bq}\\
    \mathbb{V}_{f}[\mathcal{J}] &= \int \int \kappa_\text{p}(\x, \x^\prime) \pi(\x) \pi(\x^\prime) d\x d\x^\prime,\label{eq:bq-var}
\end{align}
where $\mu_\text{p}$ and $\kappa_\text{p}$ are the GP posterior mean and covariance function.
Importantly, if $f$ has a Gaussian kernel and $\pi$ is a Gaussian or mixture of Gaussians (among other functional forms), the integrals in Eqs. \ref{eq:bq} and \ref{eq:bq-var} have closed-form solutions.

\subsection{Variational Bayesian Monte Carlo}
\label{sec:vbmc}
Variational Bayesian Monte Carlo (\VBMC; \citealp{acerbi2018variational, acerbi2020variational, huggins2023pyvbmc}) is a framework that combines variational inference (Section \ref{sec:vi}), Gaussian processes (Section \ref{sec:gps}), and Bayesian quadrature (Section \ref{sec:bq}) with the goal of approximating the posterior density. \VBMC employs a Gaussian process as a surrogate to the log-density function and then performs variational inference on the GP surrogate as opposed to using the original expensive target.  The surrogate \ELBO in variational inference is efficiently estimated via Bayesian quadrature. Moreover, \VBMC introduces acquisition functions for actively sampling new evaluations of the log density to iteratively refine the posterior approximation.

\paragraph{Surrogate ELBO}
Using the GP model $f$ as the surrogate to the log joint density $f_0$, and for a given variational posterior $q_{\qparams}$, the posterior mean of the surrogate \ELBO (see Eq. \ref{eq:supp_elbo}) can be estimated as 
\begin{equation} \label{eq:elbo}
\overline{\ELBO} = \mathbb{E}_{f}\left[\text{\ELBO}(q_{\qparams})\right] =  \mathbb{E}_{f}\left[\mathbb{E}_{\qparams} \left[f \right]\right] + \mathcal{H}[q_{\qparams}],
\end{equation}
where $\mathbb{E}_{f}\left[\mathbb{E}_{\qparams} \left[f \right]\right]$ is the posterior mean of the expected log joint under the GP model, and $\mathcal{H}[q_{\qparams}]$ is the entropy of the variational posterior \citep{acerbi2018variational}. In particular, the expected log joint takes the form
\begin{equation} \label{eq:logjoint}
\mathbb{E}_{\qparams}\left[ f \right] = \int q_{\qparams}(\x) f(\x) d \x.
\end{equation} 
Specific choices of variational family and GP representation afford \textit{closed-form} solutions for the posterior mean and variance of Eq. \ref{eq:logjoint} (and of their gradients) by means of Bayesian quadrature (see Section \ref{sec:bq}). The entropy of $q_{\qparams}$ and its gradient can be estimated via simple Monte Carlo and the reparameterization trick \citep{kingma2013auto,miller2016variational}, such that Eq. \ref{eq:elbo} can be optimized via stochastic gradient ascent \citep{kingma2014adam}. In Figure~\ref{fig:density_modeling}b, we depict the variational posterior obtained by maximizing the surrogate \ELBO in Eq.~\ref{eq:elbo}, based on the GP surrogate of the log-density depicted in Figure~\ref{fig:density_modeling}a. 


\paragraph{Variational posterior}
\VBMC takes the variational posterior family to be a flexible mixture of multivariate Gaussians. 
When coupled with a Gaussian process with the exponentiated quadratic kernel and a negative quadratic mean function (see Section~\ref{sec:overview} for the detailed form), this choice of variational posterior enables closed-form computation of the expected log joint in Eq.~\ref{eq:logjoint} via Bayesian quadrature. Consequently, it affords efficient and robust optimization of the variational objective.\footnote{Note that the entropy of a Gaussian mixture does not admit a closed form, requiring the surrogate \ELBO to be optimized using stochastic gradient descent.} 

\subsection{Sparse Variational Gaussian Processes}
\label{sec:svgp}
A major limitation of the framework described so far is that
standard ``exact'' GPs scale badly to large numbers of training points $N$, due to the cubic complexity of the matrix inversion used to evaluate the GP posterior or marginal likelihood, when fitting the GP to observations \citep{rasmussen2006gaussian}. 
Therefore, fast GP surrogate modeling of the log-density function (or the log-likelihood function) is restricted to regimes with $N \approx 10^3$ log-density evaluations. To address the issue of GP scalability, \emph{sparse} GPs have been proposed to reduce the computational burden of exact full GPs \citep{snelsonSparseGaussianProcesses2005,titsias2009variational,hensman2013gaussian}. 
In this paper, we adopt \emph{sparse variational GP regression} (\SGPR; \citealp{titsias2009variational}).

\paragraph{Sparse Gaussian process regression}

In a nutshell, many sparse GP methods can be interpreted as approximating the full GP via a ``smaller'' GP defined on a set of \emph{inducing points}
$\zz \equiv \left(\z_1, \ldots, \z_M\right)$ with $M \ll N$, and \emph{inducing variables} $\u$ representing the values of the sparse GP at the inducing points $\zz$ \citep{snelsonSparseGaussianProcesses2005, titsias2009variational}. The key difference between sparse GP methods is in how the (smaller) sparse GP posterior is constructed to best approximate the full GP posterior.

To start with the construction of a sparse GP, let $\f$ denote the latent function values corresponding to the observations $\y$.
By first assuming that the inducing values $\u$ are the result of the same Gaussian process as $\f$, we can write their joint distribution as a multivariate Gaussian distribution (for simplicity, here in the \emph{zero-mean} case):
\begin{equation}
p(\f,\u) = \mathcal{N}
\left(  \cdot \left\vert \,
\mathbf{0}, \left[\begin{pmatrix} \K_{\xx,\xx} & \K_{\xx,\zz} \\ \K_{\zz,\xx} & \K_{\zz,\zz} \end{pmatrix} \right] \right.
\right),
\end{equation}
where $\K_{\zz,\xx}$ and $\K_{\xx,\zz}$ are the cross-covariance matrices for the GP prior evaluated at the points in $\xx$ and $\zz$.

In turn, we can write the full joint distribution $p(\y,\f,\u)$ as~\citep{titsias2009variational, hensman2015scalable}:
\begin{equation}
p(\y,\f,\u) = p(\y \mid \f)p(\f \mid \u)p(\u),
\end{equation}
and find the best approximate distribution $q (\f, \u)$ in the KL-divergence sense for the posterior $p(\f, \u | \y)$. The form of the approximate distribution $q (\f, \u)$ is chosen to be $p(\f | \u) \tilde{p} (\u)$, where $\tilde{p} (\u)$ is the variational distribution for inducing variables $\u$. The target \ELBO for the sparse GP can be written as:
\begin{equation} \label{eq:sgp_objective}
    \text{GP-\ELBO} = \mathbb{E}_{q(\f)}[\log p(\mathbf{y} \mid \mathbf{f})] - \operatorname{KL} \left[ \tilde{p} (\u) || p(\u) \right],
\end{equation}
where $q(\f) \equiv \int p (\f | \u) \tilde{p} (\u) d\u$. Note that Eq.~\ref{eq:sgp_objective} is the \ELBO of the variational sparse GP approximation to the full GP. We refer to it as GP-$\ELBO$ to differentiate it from the \ELBO of the variational approximation in Sections~\ref{sec:vi} and~\ref{sec:vbmc}, which is the (surrogate) \ELBO of the approximate posterior over model parameters. 

For the case of sparse GP regression with Gaussian likelihood, \citet{titsias2009variational} proved that the optimal variational distribution that maximizes Eq. \ref{eq:sgp_objective} for fixed GP hyperparameters and inducing point locations $\zz$ is given by $\tilde{p}(\u) = \mathcal{N}(\m_\u,\mathbf{R}_{\u\u})$, with
\begin{align}
     \m_\u& = \K_{\zz,\zz}\bm{\Sigma} \K_{\zz,\xx}\mathbf{\S}^{- 1}\mathbf{y}, \label{eq:opt_mu_sgpr} \\
 \mathbf{R}_{\u\u}& = \K_{\zz,\zz}\bm{\Sigma} \K_{\zz,\zz}, \label{eq:opt_var_sgpr}
\end{align}
where $\bm{\Sigma} \equiv (\K_{\zz,\xx} \mathbf{\S}^{-1} \K_{\xx,\zz} +\K_{\zz,\zz})^{-1}$. We denote by $\vgp = (\m_\u, \mathbf{R}_{\u\u})$ the optimal variational parameters and  $p(\mathbf{f}, \mathbf{u} \mid \vgp) = q(\mathbf{f}, \mathbf{u})$ the joint variational posterior of the sparse GP under this optimal setting. Note that equality between the sparse and full GP posterior is obtained for $\zz = \xx$. 



Detailed derivations for the non-zero mean case and numerical implementation details are provided in Supplementary Material A.1-A.3.

\section{Variational Sparse Bayesian Quadrature}
\label{sec:vsbq}

In this section, we present our method for \emph{post-process Bayesian inference}, named Variational Sparse Bayesian Quadrature (\VSBQ). As mentioned, statistical analysis in computational modeling studies often relies on \emph{maximum a posteriori} (MAP) estimation,\footnote{Often with a uniform, noninformative prior, effectively reducing to maximum-likelihood estimation.} typically involving multiple runs of a numerical optimization algorithm to identify the MAP estimate from the highest log-density value. Crucially, the evaluation traces from MAP optimization, which hold valuable information about the posterior density, are usually discarded.
The \emph{post-process} inference framework we propose aims to reuse these many existing evaluations to efficiently construct a good approximation of the posterior distribution, which is particularly beneficial in scenarios with computationally expensive or noisy model evaluations. Our approach recycles valuable information, effectively converting a point estimate into a posterior approximation with minimal computational expense.

\subsection{Overview of the algorithm}
\label{sec:overview}
\begin{figure}
    \centering
    \resizebox{0.5\columnwidth}{!}{ 
        \begin{tikzpicture}[node distance=0.5cm, every node/.style={align=center}]
            \tikzstyle{process} = [rectangle, minimum width=1cm, minimum height=1cm, text centered, text width=5cm, draw=black]
            \tikzstyle{io} = [trapezium, 
            trapezium stretches=true,
            trapezium left angle=70, 
            trapezium right angle=110, minimum height=1cm,
            minimum width=0cm,
            text centered, draw=black]
            \tikzstyle{arrow} = [thick,->,>=stealth]
            \node (input) [io] {\textbf{Input}: MAP optimization traces $(\x_n, y_n, s_n)_{n=1}^{N}$};

            \node (trim) [process,below=of input] {Trim the provided evaluations};
            \node (fitgp) [process, below=of trim] {Fit a sparse GP surrogate $f$};
            \node (vi) [process, below=of fitgp] {Variational inference with Bayesian quadrature};
            \node (output) [io, below=of vi] {\textbf{Output}: Posterior approximation $q_\phi$, $\overline{\ELBO}$,  $\ELBO_\textrm{sd}$};

            \draw [arrow] (trim) -- (fitgp);
            \draw [arrow] (fitgp) -- (vi);
            \draw [arrow] (vi) -- (output);
            \draw [arrow] (input) -- (trim);
        \end{tikzpicture}
    }
    \caption{\textbf{Overview of VSBQ algorithm.} See text for details.}
    \label{fig:algo_overview}
\end{figure}

Our proposed algorithm consists of three main steps summarized in  Figure~\ref{fig:algo_overview}. 
\begin{enumerate}[nosep]
\item[a.] We first collect and ``trim'' the target evaluations $( \x_n, y_n, s_n)_{n=1}^{N}$ from MAP optimization or other sources. 
\item[b.] We fit a sparse GP surrogate to the remaining log-density evaluations. 
\item[c.] We perform efficient variational inference over the surrogate via Bayesian quadrature. 
\end{enumerate}
The output of the procedure is an approximate posterior $q_\phi$, along with the estimated surrogate \ELBO mean $\overline{\ELBO}$ (see Section~\ref{sec:vbmc}) and its standard deviation $\ELBO_\text{sd}$. We briefly describe each step below and discuss additional details in the following sections. A complete algorithmic description is available in Supplementary Material C.1.

\paragraph{Trimming of the provided evaluations}

We remove from the provided log-density evaluations all points with \emph{extremely} low log-density values relative to the maximum observed value. Such points are at best weakly informative (indication of near-zero density value) for approximating the posterior and at worst induce instabilities in the GP surrogate \citep{acerbi2018bayesian, jarvenpaa2021parallel, gammalFastRobustBayesian2022}. Note that we keep points with low density values, as those are useful for anchoring the GP surrogate \citep{desouza2022parallel} -- we just remove the \emph{extremely} low ones, as explained below. The remaining points after removal are typically still of the order of many thousands, too many to be handled efficiently by exact GPs. We call this preprocessing step \textit{trimming}.

Trimming works as follows. Consider an evaluation $\left(\x_n, y_n, s_n\right)$, for $n \in [1, N]$, where $s_n \equiv \sigma_\text{obs}(\x_n)$ is the estimated standard deviation of the observation noise. For each evaluation, we define the lower/upper confidence bounds of the log-density value as, respectively,
$\text{\textsc{lcb}}(\x_n) \equiv y_n - \beta s_n$, 
$\text{\textsc{ucb}}(\x_n) \equiv y_n + \beta s_n$,
where $\beta > 0$ is a confidence interval parameter.
We remove from our evaluation set all points $\x_n$ for which
\begin{equation}
\max_{n^\prime}(\text{\textsc{lcb}}(\x_{n^\prime})) - \text{\textsc{ucb}}(\x_n) > \eta_\text{trim}.
\end{equation}
In other words, we trim all points whose difference in underlying log-density value compared to the highest log-density value is larger than a threshold with high probability, accounting for the observation noise. A detailed discussion on how to set $\beta$ and $\eta_\text{trim}$ is provided in Supplementary Material C.1.


\paragraph{Sparse GP fitting} 
The sparse GP surrogate model uses an exponentiated quadratic kernel, 
\begin{equation} \label{eq:cov}
\begin{split}
    &\kappa(\x,\x^\prime; \sigma_f^2, \bm{\ell}) \\
    &= \sigma_f^2 \exp\left[-\frac{1}{2}\left(\x - \x^\prime\right) \mathbf{\Sigma}_\ell^{-1} \left(\x - \x^\prime\right)^\top\right],
\end{split}
\end{equation}
where $\sigma_f$ is the output scale, $\bm{\ell} \equiv (\ell_1, \ldots, \ell_\nparams)$ is the vector of input length scales, and $\mathbf{\Sigma}_\ell = \text{diag}\left[\ell_1^2, \ldots, \ell_\nparams^2\right]$.
This choice of kernel imposes a smoothness prior on the functions and affords closed-form expressions for Bayesian quadrature (see Section~\ref{sec:bq}). As depicted in Figure~\ref{fig:density_modeling}a, a GP with this covariance kernel smoothly interpolates between observations of the log-density function while providing estimates of uncertainty. 

The mean function is chosen to be a negative quadratic mean function (Eq.~\ref{eq:neg_quadratic_mean}) to ensure compatibility with Bayesian quadrature and integrability of the exponentiated surrogate, same as \citet{acerbi2018variational},
\begin{equation}
\label{eq:neg_quadratic_mean}
m(\x; m_0, \bm{\mu}, \bm{\omega}) = m_0 - \frac{1}{2} \sum_{i=1}^{\nparams} \frac{\left(x_i - \mu_i\right)^2}{\omega_i^2},
\end{equation}
where $m_0$ denotes the maximum, $\bm{\mu} \equiv \left(\mu_1, \ldots, \mu_\nparams\right)$ is the location vector, and $\bm{\omega} \equiv \left(\omega_1, \ldots, \omega_\nparams\right)$ is a vector of scale parameters. Concretely, since the GP falls back to the prior mean function when far from observed data, a negative quadratic mean function ensures that $\int_\X \exp(\mu_\text{p}(\x)) d\x$ is finite, where $\mu_\text{p}$ is the posterior mean function of the GP. A negative quadratic mean function can also be interpreted as a prior assumption that the target density is a multivariate normal. However, note that the GP can model deviations from this assumption and represent multimodal and non-Gaussian distributions as well. 

The sparse GP posterior given function observations $(\xx, \y, \s)$ and inducing point locations $\zz$ is also a Gaussian process with mean and covariance: 
%
\begin{align}\mu_{\vgp} (\tilde{\x}) 
=& \kappa(\tilde{\x},\zz) \K_{\zz,\zz}^{-1} ( \m_\u - m(\zz) ) +m(\tilde{\x}), \label{eq:variat_post_mean} \\
\kappa_{\vgp}(\tilde{\x},\tilde{\x}^\prime) =& \kappa(\tilde{\x},\tilde{\x}^\prime) - \kappa(\tilde{\x},\zz)(\K_{\zz,\zz}^{-1} - \bm{\Sigma} ) \kappa(\zz,\tilde{\x}^\prime),\label{eq:variat_post_var} \end{align}
where $\m_\u$ and $\bm{\Sigma}$ are defined in Eqs.~\ref{eq:opt_mu_sgpr} and~\ref{eq:opt_var_sgpr}.

Fitting a sparse GP to the log-density observations involves two critical components: the selection of inducing points and the sparse GP hyperparameters. Our chosen approach is detailed in Section \ref{sec:ip_selection}.

\paragraph{Variational posterior}

As per the \VBMC method described in Section~\ref{sec:vbmc}, we take the variational posterior $q_\phi$ to be a mixture of $K$ multivariate Gaussians,
\begin{equation}
\begin{split}
    q_{\qparams}(\x) = \sum_{k = 1}^K w_k \normpdf{\x}{\mmu_k}{\sigma_k^2 \mathbf{\Sigma}_{\bm{\lambda}}}
\end{split}
\end{equation}
where $\mathbf{\Sigma}_{\bm{\lambda}} \equiv \text{diag}[{{\lambda}^{(1)}}^2,\ldots,{\lambda^{(\nparams)}}^2]$ and $\bm{\lambda}=({\lambda}^{(1)},\ldots,\lambda^{(\nparams)})$ is a vector of parameter scales shared across mixture components; $w_k$, $\mmu_k$, and $\sigma_k$ are, respectively, the mixture weight, mean, and global scale of the $k$-th component. This choice of variational posterior is both flexible and conducive to enabling (sparse) Bayesian quadrature in the subsequent steps.

\paragraph{Sparse Bayesian quadrature}

Bayesian quadrature is used in \VBMC to efficiently optimize the variational objective (\ELBO) when fitting the variational posterior $q_{\qparams}$ using the exact GP surrogate (see Section \ref{sec:vbmc}). In this work, we are interested in Bayesian quadrature formulae for the \emph{sparse} GP $f$ integrated over a mixture of Gaussians $q_{\qparams}$. We call it \textit{sparse Bayesian quadrature}.\footnote{Sparse Bayesian quadrature was introduced in an earlier preprint version of this paper \citep{liFastPostprocessBayesian2023}, and more recently by \citet{warrenFastFourierBayesian2024} under the name of \emph{low-rank Bayesian quadrature}.} For multivariate normal distributions of the form $\mathcal{N}(\cdot \, ; \vmu_k,\bm{\Sigma}_k)$, with $1 \le k \le K$, the integrals of interest are Gaussian random variables $\{\mathcal{I}_k\}_{k=1}^{K}$ that depend on the inducing variables $\u$ and take the form:
\begin{equation}
\begin{split}
\mathcal{I}_k[\u] = & \int \mathcal{N}(\tilde{\x} ; \vmu_k,\bm{\Sigma}_k) f(\tilde{\x} \mid \textbf{u}) \mathrm{d} \tilde{\x}. 
\end{split}
\end{equation}
Denoting with $\psi(\textbf{u})$ the optimal variational distribution of $\u$ in \SGPR, the posterior mean of each integral is:
\begin{equation}
\begin{split}
\mathbb{E}\left[\mathcal{I}_k\right] = & \int \int \mathcal{N}(\tilde{\x} ; \vmu_k,\bm{\Sigma}_k) f(\tilde{\x} \mid \textbf{u}) \psi(\textbf{u}) \mathrm{d} \tilde{\x} \mathrm{d}\textbf{u} \\
= & \int \mathcal{N}(\tilde{\x} ; \vmu_k,\bm{\Sigma}_k) \mu_{\vgp}(\tilde{\x}) \mathrm{d} \tilde{\x}, \\
\end{split}
\end{equation}
where $\mu_{\vgp}$ is the sparse GP posterior mean function as in Eq.~\ref{eq:variat_post_mean}.
Similarly, the posterior covariance between integrals $\mathcal{I}_i$ and $\mathcal{I}_j$ is:
\begin{equation}
\label{eq:cov_sbq}
\begin{split}
\text{Cov}(\mathcal{I}_i, \mathcal{I}_j) 
&= \int \int 
    \mathcal{N}(\tilde{\mathbf{x}}; \mu_i, \Sigma_i) 
    \mathcal{N}(\tilde{\mathbf{x}}'; \mu_j, \Sigma_j)  \\
&\quad \times \text{Cov}(f(\tilde{\mathbf{x}}), f(\tilde{\mathbf{x}}')) 
    \, d\tilde{\mathbf{x}} \, d\tilde{\mathbf{x}}'  \\
&= \int \int 
    \mathcal{N}(\tilde{\mathbf{x}}; \mu_i, \Sigma_i) 
    \mathcal{N}(\tilde{\mathbf{x}}'; \mu_j, \Sigma_j)  \\
&\quad \times \kappa_\psi(\tilde{\mathbf{x}}, \tilde{\mathbf{x}}') 
    \, d\tilde{\mathbf{x}} \, d\tilde{\mathbf{x}}',
\end{split}
\end{equation}
where $\kappa_{\vgp}$ is the sparse GP posterior covariance function as in Eq.~\ref{eq:variat_post_var}. Both the posterior mean and variance of the integral can be obtained in closed form as integrals for the product of Gaussians. Therefore, we can compute the $\ELBO$ and its standard deviation $\ELBO_\textrm{sd}$ efficiently. Thus, once the sparse GP fitting is done, obtaining a tractable posterior approximation is computationally cheap and robust. Derivations are provided in Supplementary Material A.4.

\subsection{Inducing points and hyperparameter selection}
\label{sec:ip_selection}

The selection of hyperparameters $\hyp$ for the GP kernel and mean functions, as well as the placement of $M$ inducing points $\zz$, is critical for the approximation quality of a sparse GP. While in principle we could jointly optimize $\hyp$ and $\zz$, this optimization can be extremely expensive and inefficient and is not recommended by modern practice \citep{burt2020jmlr}.
Therefore, we follow \citet{burt2020jmlr, Maddox2021ConditioningSV} and adaptively select inducing points that minimize an empirical error term, namely the trace of the error of a rank-$M$ Nystr\"om approximation. 


\paragraph{Inducing point location}

The \ELBO in \SGPR \citep{titsias2009variational}, extended in this paper to the heteroskedastic case (observation-dependent noise), takes the closed form:
\begin{equation} \label{eq:gpelbo}
\begin{split}
    \text{GP-\ELBO}(\zz, \hyp) = & \log\mathcal{N}\left(\mathbf{y};\m({\xx}), \Q_{\xx,\xx} + \S \right) \\
    &- \frac{1}{2}{\Tr}\left(\left(\K_{\xx, \xx} - \Q_{\xx, \xx}\right) \mathbf{\S}^{- 1}\right),
\end{split}
\end{equation}
where $\Q_{\xx,\xx} \equiv \K_{\xx,\zz} \K_{\zz,\zz}^{- 1} \K_{\zz,\xx}$. Note that the GP-\ELBO depends on the location of inducing points $\zz$ and GP hyperparameters $\hyp$, and consists of the log probability density function of a multivariate normal term, minus the trace of an error term. 

For known GP hyperparameters, \citet{burt2020jmlr} show that sampling from an $M$-determinantal point process ($M$-DPP) to select inducing points $\zz \subset \xx$ will make the trace error term of the GP-\ELBO (Eq. \ref{eq:gpelbo}) close to its optimal value. Since the DPP approach is intractable, we follow \citet{burt2020jmlr} and \citet{Maddox2021ConditioningSV} who recommend instead to sequentially select inducing points that greedily maximize the diagonal of the error term,
\begin{equation} \label{eq:greedyselection}
\z^\star = \arg\max_\z \text{diag}\left[ \left(\K_{\xx,\xx} - \Q_{\xx,\xx}\right) \S^{-1} \right].
\end{equation}
In turn, this reduces the trace error in Eq. \ref{eq:gpelbo} and aims to maximize the GP-\ELBO under the current (fixed) GP hyperparameters. Eq. \ref{eq:greedyselection} can be interpreted as weighted \emph{greedy variance selection} \citep{burt2020jmlr}, in that it selects points with maximum prior conditional marginal variance at each point in $\xx$ (conditioned on the inducing points selected so far), weighted by the precision (inverse variance) of the observation at that location. Overall, this selection procedure can be achieved with complexity $\mathcal{O} (N M^2)$ by Algorithm 1 in \citet{chen2018fast}.

\paragraph{GP hyperparameters}

The procedure for the selection of inducing points mentioned above requires known GP hyperparameters $\hyp$. In practice, it is often enough to start with a reasonable estimate for $\hyp$ and then iterate the process multiple times~\citep{burt2020jmlr}. To obtain an initial estimate for the GP hyperparameters, we fit an exact GP on a small subset of the data and use the exact GP hyperparameters for initial inducing point selection. The subset is chosen via stratified $K$-means, where we ensure that the chosen subset is representative of the full set in terms of both location and log-density values.
After the above initialization, we iterate sparse GP hyperparameter optimization and inducing points selection using the current sparse GP hyperparameters, until no improvement on the GP-\ELBO is found. Similarly to other block-optimization procedures, this process is not guaranteed to find the global optimum of the GP-\ELBO, but it often works well in practice~\citep{burt2020jmlr}.

\subsection{Noise shaping}
\label{sec:noiseshaping}

We recall that while we use a surrogate of the \emph{log density} (Figure~\ref{fig:density_modeling}a), our end goal is to accurately estimate the Bayesian posterior \emph{density} (Figure~\ref{fig:density_modeling}b). This goal can be formalized from a decision-theoretical perspective as minimizing the integrated $L^p$ error between our approximate posterior density and the true posterior;\footnote{See \citet{jarvenpaa2021parallel} for a similar analysis.} a formulation which is however generally intractable.
In practice, this means that, given a limited-resource surrogate model, we want the surrogate to spend resources to accurately represent high log-density regions and allocate fewer resources to low log-density regions, since the latter areas will map close to zero density regardless of the exact log-density value, with near-zero influence on the reconstruction error of the density.

\begin{figure*}[t!]
    \centering
    \subcaptionbox{True density function (black line) and the observations (gray circles).\label{fig:true_pdf_noise_shaping}}{
        \includegraphics[scale=1]{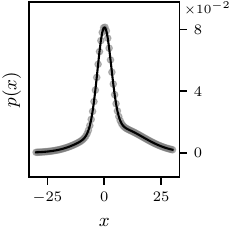}
    }\hspace{1em}
    \subcaptionbox{Sparse GP approximation (orange line) to the true log-density function (black line) \textit{without} noise shaping. 
\label{fig:without_noise_shaping_gp}}{
        \includegraphics[scale=1]{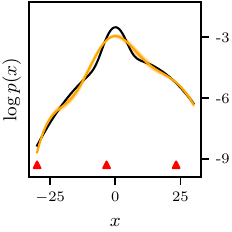}
    }\hspace{1em}
    \subcaptionbox{Variational posterior (green dashed line) \textit{without} noise shaping.\label{fig:without_noise_shaping_vp}}{
        \includegraphics[scale=1]{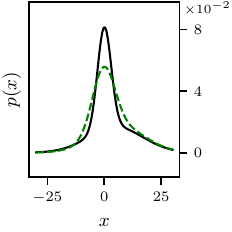}
    } \\
    \subcaptionbox{Noise shaping curve corresponding to Eq.~\ref{eq:noiseshaping}.\label{fig:noisehsaping_fun}}{
        \includegraphics[scale=1]{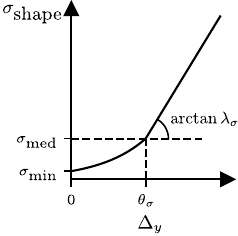}
    }\hspace{1em}
    \subcaptionbox{Sparse GP approximation \textit{with} noise shaping.\label{fig:with_noise_shaping_gp}}{
        \includegraphics[scale=1]{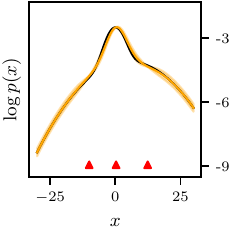}
    }\hspace{1em}
    \subcaptionbox{Variational posterior \textit{with} noise shaping.\label{fig:with_noise_shaping_vp}}{
        \includegraphics[scale=1]{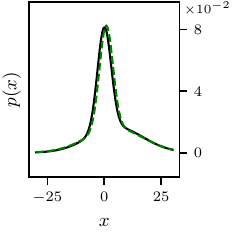}
    }

    \caption{\textbf{Illustration of noise shaping effect.} The red triangles are locations of inducing points selected via Eq.~\ref{eq:greedyselection}. Noise shaping (bottom row) improves the selection of inducing points and the approximation of the sparse GP in the high posterior density region, compared to a sparse GP \emph{without} noise shaping (top row). A better sparse GP consequently leads to an improved variational posterior.  
    }
    \label{fig:noise_shaping}
\end{figure*}
This scenario is exemplified in our case in Figure~\ref{fig:noise_shaping}.
Here, the surrogate model is a sparse GP whose main resource is the inducing points and their location, as discussed in Section \ref{sec:svgp}. Given a target density (Figure~\ref{fig:true_pdf_noise_shaping}), if we naively use a sparse GP to model the target log density, the inducing points are allocated equally over the region (Figure~\ref{fig:without_noise_shaping_gp}), yielding an inaccurate approximation of the density (Figure~\ref{fig:without_noise_shaping_vp}).
Our proposed solution consists of \emph{noise shaping} (Figure~\ref{fig:noisehsaping_fun}), a simple motivated heuristic that increases the likelihood noise of lower log-density observations, effectively downweighing the contribution of these points to the sparse GP objective (details below). With noise shaping in place, the sparse GP automatically favors the allocation of inducing points (Eq.~\ref{eq:greedyselection}) to better capture higher-density regions (Figure~\ref{fig:with_noise_shaping_gp}), yielding a highly accurate representation of the posterior density (Figure~\ref{fig:with_noise_shaping_vp}).

Noise shaping consists of adding an artificial `shaping' noise term, $\sigma_\text{shape}(y)$, to the Gaussian likelihood of each observation in the model, without changing the actual observation $y$. We assume shaping noise to be Gaussian and independent, such that the total likelihood variance for observation $(\x_n, y_n, \sigma_\text{obs}(\x_n))$ becomes:
\begin{align}
\sigma^2_\text{tot}(\x_n, y_n) &= \sigma^2_\text{obs}(\x_n) + \sigma^2_\text{shape}(\Delta y_n),
\end{align}
where $\sigma^2_\text{obs}(\x_n)$ is the estimated measurement variance at $\x_n$,
and $\Delta y_n \equiv y_\text{max} - y_n$, with $y_\text{max}$ the maximum observed log density.

Note that the added variance depends on $y_n$, the \emph{observation} at $\x_n$. We design $\sigma^2_\text{shape}(\Delta y)$ to add minimal noise to relatively high-valued observations of the log-density, and increasingly larger noise to lower-density points. Specifically, we define
\begin{align} \label{eq:noiseshaping}
    \sigma_\text{shape}(\Delta y) =& \exp((1-\rho) \log \sigma_{\text{min}} + \rho \log \sigma_{\text{med}}) \nonumber\\
    &+ \mathbf{1}_{\Delta y  \geq \theta_\sigma} {\lambda_\sigma} (\Delta y - \theta_\sigma),
\end{align}
where $\rho = \min(1,\Delta y/\theta_\sigma )$;  $\theta_\sigma$ is a threshold for `very low density' points, at which we start the linear increase; $\lambda_\sigma$ is the slope of the increase; and $\sigma^2_\text{min}$ and $\sigma^2_{\text{med}}$ are two shape parameters. $\sigma_{\text{med}}$ is the added noise at the low density threshold $\theta_\sigma$. Noise shaping is designed to be small for $\Delta y < \theta_\sigma$ (below $\sigma_\text{med}^2$) and only then it starts taking substantial values. Figure \ref{fig:noisehsaping_fun} shows an example of how the added shaping noise $\sigma_\text{shape}$ increases with $\Delta y$. The design principles for noise shaping and specific values of these parameters are provided in Supplementary Material C.1.

As a further motivation for noise shaping, we show that noise shaping is mathematically equivalent to downweighing observations in the sparse GP objective of \SGPR.
With noise shaping, the GP observation likelihood $p(\mathbf{y} \mid \mathbf{f})$ becomes $\tilde{p}(\mathbf{y} \mid \mathbf{f})$, with
\begin{equation}
\begin{split}
  \log \tilde{p}(\mathbf{y} \mid \mathbf{f}) &= \sum_{n=1}^N \log \tilde{p}(y_n | f_n ) \\
  &= C -\sum_{n=1}^N \frac{(y_n - f(\x_n))^2}{2\sigma^2_\text{tot}(\x_n, y_n)},  
\end{split}
\end{equation}
where $C = - \sum_{n} \log\sqrt{2\pi\sigma^2_\text{tot}(\x_n, y_n) }$ is a constant that does not depend on the GP $\f$.
According to \citet{hensman2015scalable}, the expected log-likelihood part of the sparse GP objective (Eq. \ref{eq:sgp_objective}) can be equivalently written as a sum over individual data points,
\begin{equation}
\mathbb{E}_{q(\f)}[\log \tilde{p}(\mathbf{y} \mid \mathbf{f})] = \sum_{n=1}^N \mathbb{E}_{q(f_n)}\left[\log \tilde{p}(y_n | f_n )\right],
\end{equation}
where $q(\f)$ is the variational GP posterior.
Thus, with noise shaping, the expected log-likelihood term of the GP-\ELBO becomes,  
\begin{equation}
\begin{split}
    &\mathbb{E}_{q(\f)}[\log \tilde{p}(\mathbf{y} \mid \mathbf{f})]  \\
    &=\text{const} + \sum_{n=1}^N w_n \mathbb{E}_{q(f_n)}[\log p(y_n | f_n )], 
\end{split}
\end{equation}
where $w_n \equiv \frac{\sigma^2_\text{obs}(\x_n)}{\sigma^2_\text{tot}(\x_n, y_n)} \le 1$.
That is, by assigning larger `shaping' noise to low-density observations, we are downweighing their role in the sparse GP representation, guiding the sparse GP to better represent higher-density regions. Noise shaping also helps during inducing point selection (Eq.~\ref{eq:greedyselection}), as shown in Figure~\ref{fig:noise_shaping}.


\subsection{Approximation error}

In this paper, we use a sparse GP as a surrogate model for the log-density function, which significantly reduces the complexity of the algorithm and makes it possible to post-process a large number of evaluations. At the same time, a sparse GP introduces new approximation errors compared to an exact GP. The approximation error in representing the log-density function via the sparse GP surrogate further leads to errors in the variational posterior. 
In this section, we present a theoretical result that bounds the (additional) approximation error induced by using a sparse GP in \VSBQ. 

We introduce Lemma \ref{dist_predict} and \ref{lemma_int_exp} first, whose proofs are provided in Supplementary Material B. Recall from Section \ref{sec:svgp} that $\f$ denotes the latent function values corresponding to the observations $\y$, $\u$ represents values at the inducing points $\zz$, $\vgp$ denotes the optimal variational parameters of the sparse GP and $p(\mathbf{f}, \mathbf{u} \mid \vgp)$ is the joint posterior distribution of the sparse GP given the optimal variational parameters. Denote with $f$ and $f_e$ the posterior predictive functions of the sparse GP and exact GP, respectively. The posterior mean functions of the sparse GP and exact GP are represented as $\bar{f}$ and $\bar{f}_e$. Finally, $\Vert \cdot \Vert_\text{TV}$ denotes the total variation distance.

\begin{lemma}
\label{dist_predict}
Assume that $D_\text{KL}(p(\mathbf{f},\mathbf{u} \mid \vgp) \Vert   p(\mathbf{f},\mathbf{u} \mid \mathbf{y})) < \gamma$.
Then, for any $\ell>0$ there exists $K_\ell$ such that, for any $\x^*$, $\vert \mathbb{E}[{f}(\x^*)^\ell] - \mathbb{E}[{f}_e(\x^*)^\ell] \vert < K_\ell \sqrt{\gamma/2}$. There also exists $K_\mathrm{e}$ such that, for any $\x^*$, $\vert \mathbb{E}[\mathrm{exp}({f}(\x^*))] - \mathbb{E}[\mathrm{exp}({f}_e(\x^*))] \vert < K_\mathrm{e} \sqrt{\gamma/2}$.
\end{lemma}
\begin{lemma}
 \label{lemma_int_exp}
 Let $a$ and $b$ be two functions associated with two distributions defined on $\mathcal{X}$, $\pi_a \propto \exp(a(\cdot)) $ and $\pi_b \propto \exp(b(\cdot))$. If $\forall x, \ \vert a(x) - b(x) \vert < K$, then:
 \begin{equation}\Vert \pi_a - \pi_b \Vert_\text{TV} \leq  1-\exp(-K).\end{equation}
 \end{lemma}
Given the two lemmas above, we can now state a theorem that bounds the distance between the variational posterior $q_\phi$ constructed from the sparse GP and the variational posterior $q_{\phi_e}$ from the exact GP.
\begin{theorem}
\label{th:variat_post}
Let $f$ and $f_e$ be the sparse GP and exact GP approximation of the target log-density function $f_0$, respectively. Let $q_\phi$ and $q_{\phi_e}$ be the variational posteriors obtained from the exact GP and sparse GP. Let $\pi$ be the normalized posterior density associated with $\exp(\bar{f})$ (resp., $\pi_e$ and $\exp(\bar{f}_e)$). Assume that $D_\text{KL}(q_\phi \Vert \pi)<\gamma_1$ and $D_\text{KL}(q_{\phi_e} \Vert \pi_e)<\gamma_2$, then there exist constants $K_l$ and $\gamma$, such that:
\begin{equation}
\begin{split}
    \Vert q_\phi - q_{\phi_e}\Vert_\text{TV}  <& \sqrt{\gamma_1/2}+\sqrt{\gamma_2/2} \\
    &+ (1 - \mathrm{exp}(-K_l \sqrt{\gamma/2})). 
\end{split}
\end{equation}
\end{theorem}
\begin{proof}
By the triangle inequality, with $\pi$ and $\pi_e$ the distributions associated with $\bar{f}$ and $\bar{f}_e$: $\Vert q_\phi - q_{\phi_e} \Vert_\text{TV} \leq \Vert q_\phi - \pi\Vert_\text{TV} + \Vert \pi_e - q_{\phi_e} \Vert_\text{TV} + \Vert \pi_e - \pi \Vert_\text{TV} $.
The first two terms are bounded by the assumptions and Pinsker's inequality, the last one by Lemma \ref{dist_predict} and \ref{lemma_int_exp}.
\end{proof}


Theorem~\ref{th:variat_post} provides a theoretical justification for the quality of the variational posterior obtained from a sparse GP surrogate. While the bound is not directly useful for empirical tuning---due to its dependence on constants such as $K_l$ and $\gamma$---it reveals how the total error decomposes into two main interpretable and controllable sources: the error introduced by approximating the exact GP with a sparse GP, and the error due to variational inference over the surrogate.

\section{Experiments}
\label{sec:experiments}

We empirically investigate the performance of \VSBQ with both synthetic and real-world benchmark problems. Each problem is represented by a target posterior density assumed to be a \emph{black box}: gradients are unavailable and evaluations of the log-density function may be (mildly) expensive and noisy. We measured the quality of the posterior approximation by comparing (1) the variational posterior with the ground-truth posterior and (2) the estimated log normalizing constant (via the \ELBO) with the ground-truth log marginal likelihood. The ground-truth posterior is represented by samples from well-tuned and extensive \MCMC sampling~\citep{emcee} or rejection sampling for the 2D synthetic problem.

\subsection{Baseline methods}
\label{sec:baseline}

We recall that with black-box inference we mean that the target lacks gradients and may be expensive to evaluate and possibly noisy.
As few other methods afford post-process and black-box inference of the posterior, we compare \VSBQ against three baselines: black-box variational inference (\BBVI), neural network regression (\NNR), and the popular Laplace approximation (\LA). Of these, only \NNR is also a post-process method. Further implementation details of \VSBQ are provided in Supplementary Material C.1.

\paragraph{Black-box variational inference} 
When the target posterior density is a black-box, the reparameterization trick~\citep{kingma2013auto} cannot be applied to estimate the \ELBO gradient for stochastic variational inference. Instead, one needs to resort to other techniques for computing the gradient of the \ELBO, such as the score function estimator, also known as the REINFORCE estimator~\citep{ranganath2014black}, which often has higher variance compared to the reparameterization trick~\citep{Gal2016Uncertainty, xuVarianceReductionProperties2019}. Specifically, by differentiating Eq.~\ref{eq:supp_elbo}, the \ELBO gradient $\nabla_{\qparams} \ELBO\left[ q_{\qparams} \right]$ can be written as,
\begin{equation}
    \begin{split}
    & \nabla_{\qparams} \mathbb{E}_{\qparams} \left[\log \frac{\like p(\x)}{q_{\qparams}(\x)} \right] \\ 
    &= \mathbb{E}_{\qparams} \left[ \nabla_{\qparams} \log q_{\qparams}(\x) \left( \log \like p(\x) - \log q_{\qparams}(\x) \right) \right]. \\
\end{split}
\end{equation}    
In addition, we leverage the control variates technique to reduce the gradient variance (see the supplement for details). We experiment with the following variational distributions for \BBVI: a Gaussian with \emph{diagonal} covariance matrix, a Gaussian with \emph{full-rank} covariance matrix, and a mixture of Gaussians with $K=5$ and $K=50$ components, respectively, where the mixture of Gaussians (MoG) admits the same form as the variational posterior in \VSBQ. The Adam optimizer~\citep{kingma2014adam} is used for optimizing the \ELBO with stochastic gradients. For each variational distribution choice, we applied grid search on the learning rate hyperparameter in $\{ 0.01, 0.001\}$ and the number of Monte Carlo samples for gradient estimation in $\{1, 10, 100\}$ and reported the best result, according to the estimated \ELBO value. Since \BBVI typically requires a large number of target density evaluations for convergence, in all the experiments we assign \textit{ten times more} evaluation budget to \BBVI than \VSBQ to make it a stronger baseline for reference. Moreover, note that this is not a post-process method, but we include it as a reasonable performance reference. For more details, see Supplementary Material C.3.

\paragraph{Neural network regression} 
For a direct comparison with our method, we develop a post-process inference algorithm based on a deep neural network regression surrogate (\NNR) instead of a sparse GP surrogate, otherwise leaving the post-process procedure (Figure~\ref{fig:algo_overview}) as much as possible the same. This is a competitive baseline since deep neural networks exhibit strong regression performance in the presence of a large number of data points \citep{Goodfellow-et-al-2016}. For the network architecture, we use a multilayer perceptron (MLP) with an input layer of dimension $D$, four hidden layers of $1024$ units, and an output layer for predicting the log-density value. The activation function is chosen to be the rectified linear units (ReLU; \citealp[Chapter 6]{Goodfellow-et-al-2016}). In addition, we add a negative quadratic mean function to the neural network output to ensure that it represents a valid log-density surrogate function.\footnote{An MLP with ReLU activations is a \textit{continuous piecewise affine} function~\citep{aroraUnderstandingDeepNeural2018}, and therefore adding a trainable negative quadratic mean function ensures the integrability of the exponentiated surrogate.}
The negative quadratic mean function is the same as the one used for the (sparse) GP (see Eq.~\ref{eq:neg_quadratic_mean}). In total, the surrogate function $g$ is:
\begin{equation}
    g(\x; \mathbf{w}) = m_0 - \frac{1}{2} \sum_{i=1}^{\nparams} \frac{\left(x_i - \mu_i\right)^2}{\omega_i^2} + \text{MLP}(\x),
\end{equation}
where $\mathbf{w}$ denotes the neural network parameters (weights and biases), including additional surrogate model parameters (i.e., for the quadratic mean).

We adopt the observation noise model delineated in Section~\ref{sec:noiseshaping}, yielding the loss:
\begin{equation}
\label{eq:nnr_objective}
    \mathcal{L}(\mathbf{w}) = \sum_{n=1}^{N} 
     \frac{\left(g(\x_n; \mathbf{w}) - y_n \right)^2}{\sigma_\text{tot}^2(\x_n, y_n)},
\end{equation}
where $\sigma_\text{tot}^2(\x_n, y_n) = \sigma_{\text{obs}}^2(\x_n) + \sigma_{\text{shape}}^2(\Delta y_n)$, as per noise shaping.\footnote{We found that noise shaping empirically helped stabilize the neural network training.}
We optimize the neural network parameters by minimizing the objective in Eq.~\ref{eq:nnr_objective} with the AdamW optimizer~\citep{loshchilovDecoupledWeightDecay2019}.
For regularization, we considered the `weight decay' hyperparameter $\alpha \in \{0, 0.01, 0.1\}$, selecting the best neural network surrogate based on the loss on a split validation dataset. Finally, we use stochastic variational inference with the reparameterization trick~\citep{kingma2013auto} to compute the approximate posterior. This part is the same as \VSBQ except that the expected log joint of the surrogate in Eq.~\ref{eq:supp_elbo} is approximated via Monte Carlo samples rather than calculated exactly via sparse Bayesian quadrature. For more details, see Supplementary Material C.4.

\paragraph{Laplace approximation} 
The Laplace approximation method (\LA) computes a multivariate normal approximation of the posterior centered at the MAP location in the unbounded parameter space (see ``Inference space'' in Section \ref{sec:procedure_metrics} below), providing both a posterior approximation and an estimate of the marginal likelihood \citep{mackay2003information}. Despite its simplicity, the Laplace approximation is often used in practice for its efficiency and can yield reasonable posterior approximations~\citep{piray2019hierarchical, daxbergerLaplaceReduxEffortless2021}. Note that \LA requires additional log-density evaluations for numerically estimating the Hessian and is not easily applicable to noisy evaluations, so this is not a fully post-process inference method, but we also include it as a popular baseline and performance reference.

\subsection{Procedure and metrics}
\label{sec:procedure_metrics}
In this section, we describe the procedure and metrics for the experiments.

\paragraph{MAP estimation} 
It is worth noting that the MAP estimate (the mode of the posterior density) is not parameterization invariant. 
To closely align with real-world scenarios, we find the MAP estimate in the space in which the target model parameters are originally defined, as this is the approach practitioners would typically use. If the parameter space is bounded, we let the optimization algorithm handle the bound constraints. For each problem, we allocate a total budget of 3000$D$ target evaluations across multiple MAP optimization runs, where $D$ is the dimension of the problem.
In the main text, we report results obtained using the \cmaes optimization algorithm for MAP estimation. Covariance matrix adaptation evolution strategy (\cmaes) is a stochastic, derivative-free evolutionary algorithm for continuous optimization, widely adopted and very effective in black-box and potentially noisy objective settings~\citep{hansenCMAEvolutionStrategy2016}.
Further analysis with another black-box optimization algorithm, based on a hybrid Bayesian optimization technique (\bads; \citealp{acerbi2017practical,singh2024pybads}), is provided in Supplementary Material C.5.

\paragraph{Inference space} 
\VSBQ, \NNR, \LA and \BBVI all operate in an \emph{unbounded} parameter space also known as \emph{inference space}. The unbounded inference space is necessary to define and manipulate the multivariate normals (and mixtures thereof) used by all our algorithms. Parameters that are originally subject to bound constraints are mapped to the inference space via a shifted and rescaled probit transform, with an appropriate Jacobian correction to the log-density values.
A similar approach is common in probabilistic inference software \citep{carpenter2017stan, acerbi2018variational}. The approximate posteriors are transformed back to the original space via the corresponding inverse transform, for computing the metrics against the ground truth posterior.

\paragraph{Procedure}

 The optimization trace points and corresponding log-density values are collected as the training dataset for \VSBQ and \NNR.  For each problem, we repeated the entire optimization procedure -- each with \textit{multiple} MAP estimation runs, as explained above -- ten times with different random seeds. This yielded ten different training sets per problem, used to assess the robustness and reliability of the methods. The number of inducing points for the sparse GP is set to $100D$. The number of mixture components $K$ is 50 for both \VSBQ and \NNR. For \BBVI, as mentioned in Section~\ref{sec:baseline}, a budget of $10 \times 3000D = 30000D$ target density evaluations per random seed is allocated for stochastic optimization. In the case of a noisy target, we further vary the observation noise level to study the noise sensitivity of the methods. For the Laplace approximation, we first find the MAP point by transforming the parameter space to unbounded if needed, via a nonlinear mapping.\footnote{For the Laplace approximation to be valid, it is necessary to find the MAP estimate in the unbounded space subsequently used to compute the multivariate normal approximation; this is particularly important if the mode is close to the bounds.} We subsequently compute the Hessian matrix via adaptive numerical differentiation~\citep{numdifftools}. Finally, we compute the performance metrics as detailed below.

\paragraph{Metrics}
We use multiple metrics for assessing the quality of different aspects of the posterior approximation: the absolute difference between the true and estimated log marginal likelihood ($\Delta$LML), the mean marginal total variation distance (\MMTV), and the ``Gaussianized'' symmetrized KL divergence (\gskl) between the approximate and the true posterior \citep{acerbi2020variational}. For all metrics, lower is better. We describe the three metrics below:
\begin{itemize}
    \item $\Delta$LML is the absolute difference between true and estimated log marginal likelihood. The true log marginal likelihood is computed analytically, via numerical quadrature methods, or estimated from extensive \MCMC sampling via Geyer’s reverse logistic regression~\citep{geyer1994estimating}, depending on the structure of each specific problem. Differences in log model evidence $\ll$ 1 are considered negligible for model selection \citep{burnham2003model}, and therefore for practical usability of a method we aim for an LML loss $<$ 1.
    \item The \MMTV quantifies the (lack of) overlap between true and approximate posterior marginals, defined as
    \begin{equation} \label{eq:mmtv}
        \text{\MMTV}(p,q) = 
        \sum_{d = 1}^\nparams\! \int_{-\infty}^{\infty} \!\frac{\left|p^\text{M}_d(x_d) - q^\text{M}_d(x_d) \right|}{2\nparams} dx_d,
    \end{equation}
    where $p^\text{M}_d$ and $q^\text{M}_d$ denote the marginal densities of $p$ and $q$ along the $d$-th dimension. Eq. \ref{eq:mmtv} has a direct interpretation in that, for example, an \MMTV metric of 0.5 implies that the posterior marginals overlap by 50\% (on average across dimensions). As a rule of thumb, we consider a threshold for a reasonable posterior approximation to be \MMTV $< 0.2$, which is more than $80\%$ overlap.

    \item The \gskl metric is sensitive to differences in means and covariances, being defined as
\begin{equation} \label{eq:GsKL}
\begin{split}
\text{GsKL}(p, q) = &\frac{D_\text{KL}\left(\mathcal{N}[p]||\mathcal{N}[q]\right)}{2D}\\
    &+ \frac{D_\text{KL}(\mathcal{N}[q]|| \mathcal{N}[p])}{2D},
\end{split}
\end{equation}
    where $D_\text{KL}\left(p||q\right)$ is the Kullback-Leibler divergence between distributions $p$ and $q$ and $\mathcal{N}[p]$ is a multivariate normal distribution with mean equal to the mean of $p$ and covariance matrix equal to the covariance of $p$ (and same for $q$).\footnote{In contrast to the definition in \citet{acerbi2020variational}, we normalize the \gskl metric by the number of dimensions $D$.} Eq. \ref{eq:GsKL} can be expressed in closed form in terms of the means and covariance matrices of $p$ and $q$. For reference, two Gaussians with unit variance and whose means differ by $\sqrt{2}$ (resp., $\frac{1}{2}$) have a \gskl of 1 (resp., $\frac{1}{8}$). As a rule of thumb, we consider a desirable target to have \gskl less than $\frac{1}{8}$.
\end{itemize}

For each metric, we report the median and bootstrapped 95\% confidence interval (CI) of the median over the ten different training datasets and random seeds. Further details for performance evaluation can be found in Supplementary Material C.2.

\subsection{Synthetic problems}
We begin our analysis with two synthetic problems with known log-density functions.

\paragraph{Two Moons bimodal posterior}
%
%
To evaluate how our method deals with multimodality, we first consider a synthetic bimodal posterior consisting of two `moons' with different weights in $D = 2$. The bimodal posterior admits an analytic log-density function:
\begin{align}
    p(\x) =& \log ( \exp(\kappa x_1/r)/3 + 2\exp(-\kappa x_1/r)/3 ) \nonumber\\
    &-\frac{1}{2} \left(\frac{r - 1/\sqrt{2}}{0.1} \right)^2,
\end{align} 
where $r= \Vert \x \Vert_2$ with $\x=(x_1,x_2)$, $\kappa=8$. The posterior corresponds to an angle following a von Mises distribution and a normal radius in the polar coordinate system.
This density is not defined for $\x=0$, but remains defined for almost all $\x$ with respect to the Lebesgue measure. 

\begin{figure*}
    \centering
    \subcaptionbox{\VSBQ \label{fig:bimodal_svbmc}}{\includegraphics[scale=1]{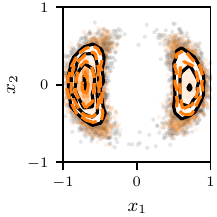}}\hspace{0em}
    \subcaptionbox{\NNR \label{fig:bimodal_nnr}}{\includegraphics[scale=1]{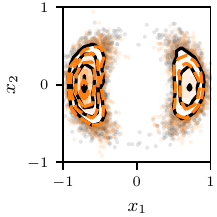}}\hspace{0em}
    \subcaptionbox{\LA \label{fig:bimodal_la}}{\includegraphics[scale=1]{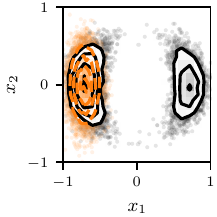}}
    \subcaptionbox{\BBVI \label{fig:bimodal_mog-50}}{\includegraphics[scale=1]{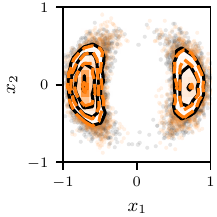}}
    \caption{\textbf{Two Moons.} The black density contours and points denote ground truth samples. The orange density contours and points represent the posterior samples from (a) \VSBQ; (b) \NNR; (c) \LA; (d) \BBVI with MoG ($K=50$). \VSBQ, \NNR and \BBVI with MoG ($K=50$) perfectly recover the bimodal density.}
    \label{fig:twomoons}
\end{figure*}
As shown in Figure~\ref{fig:twomoons} and Table~\ref{tab:twomoons}, \VSBQ, \NNR and \BBVI with MoG($K=50$) reconstruct the bimodal target almost perfectly. By contrast, the Laplace approximation can only cope with unimodal posteriors and thus unsurprisingly fails in this case, despite its otherwise relative simplicity. \BBVI with a diagonal Gaussian, a full-rank Gaussian, and a MoG($K=5$) also reveal inferior performance compared to the other methods.

\begin{table*}
    \centering
    \caption{\textbf{Two Moons posterior ($D=2$).} The method performance is measured using the metrics $\Delta$LML, \MMTV, and \gskl. For all metrics, lower values indicate better performance. We bold the best results based on the 95\% confidence interval (CI) of the median. If there are overlaps between CIs, we bold all overlapping values. Note that \LA is deterministic, hence its CI has zero length and is not displayed.\label{tab:twomoons}}
    \begin{tabular}{c c c c c}
        \toprule
        & \textbf{$\Delta$LML ($\downarrow$)} & \textbf{MMTV} ($\downarrow$) & \textbf{GsKL} ($\downarrow$) \\
    \toprule
Gaussian (diagonal) & 0.56 $\scriptstyle{[0.50, 1.2]}$ & 0.22 $\scriptstyle{[0.21, 0.37]}$ & 7.6 $\scriptstyle{[6.6, 15.]}$\\ 
Gaussian (full-rank) & 1.1 $\scriptstyle{[0.57, 1.2]}$ & 0.36 $\scriptstyle{[0.21, 0.37]}$ & 14. $\scriptstyle{[7.3, 16.]}$\\ 
MoG ($K=5$) & 0.44 $\scriptstyle{[0.28, 0.79]}$ & 0.21 $\scriptstyle{[0.13, 0.28]}$ & 5.7 $\scriptstyle{[0.16, 8.7]}$\\ 
MoG ($K=50$) & \textbf{0.0061} $\scriptstyle{[0.0021, 0.015]}$ & \textbf{0.022} $\scriptstyle{[0.018, 0.031]}$ & \textbf{0.00033} $\scriptstyle{[0.00012, 0.00088]}$\\
\LA & 0.43 & 0.19 & 8.1\\
\NNR & 0.0085 $\scriptstyle{[0.0038, 0.015]}$ & \textbf{0.021} $\scriptstyle{[0.020, 0.023]}$ & \textbf{0.00033} $\scriptstyle{[0.00016, 0.00050]}$\\
\VSBQ & \textbf{0.0017} $\scriptstyle{[0.00057, 0.0026]}$ & \textbf{0.020} $\scriptstyle{[0.018, 0.021]}$ & \textbf{8.5e-05} $\scriptstyle{[4.4e-05, 0.00019]}$\\
    \toprule
    \end{tabular}
\end{table*}

\paragraph{Multivariate Rosenbrock-Gaussian}

We now experiment with a complex synthetic target of known shape to demonstrate the flexibility of our algorithm.
Here we consider a target likelihood in $D = 6$ which consists of the direct product of two exponentiated Rosenbrock (`banana') functions $\mathcal{R}(x, y)$ and a two-dimensional normal density. We apply a Gaussian prior to all dimensions. The target density is thus:
\begin{equation}
\begin{split}
    p(\mathbf{x}) \propto &~ e^{\mathcal{R}(x_1, x_2)} e^{\mathcal{R}(x_3, x_4)} \mathcal{N}([x_5, x_6]; \mathbf{0}, \mathbb{I}) \\
    & \cdot \mathcal{N}(\mathbf{x}; \mathbf{0}, 3^2 \mathbb{I}),
\end{split}
\end{equation}
where $\mathcal{R}(x_1,x_2) = -(x_1^2 - x_2)^2 - \frac{(x_2-1)^2}{100}$. 

As shown in Table~\ref{tab:banana}, both \VSBQ and \NNR approximate this complex posterior well, with \VSBQ performing slightly better than \NNR in terms of metrics. \LA does not give a satisfactory approximation either, due to the heavily non-Gaussian nature of the underlying posterior. Among the \BBVI methods, \BBVI with MoG ($K=50$) achieves the best results. However, it still underperforms compared to \VSBQ, even when allocated a tenfold higher density evaluation budget. The poor metrics further indicate challenges in fitting a full-rank Gaussian using \BBVI.
A visualization of the approximate posteriors and the ground-truth posterior is provided in Supplementary Material C.6.

\begin{table*}
    \centering
    \caption{\textbf{Multivariate Rosenbrock-Gaussian ($D=6$).} See Table~\ref{tab:twomoons} for a detailed description of metrics and bolding criteria.\label{tab:banana}}
    \begin{tabular}{c c c c c}
        \toprule
        & \textbf{$\Delta$LML} ($\downarrow$) & \textbf{MMTV} ($\downarrow$) & \textbf{GsKL} ($\downarrow$) \\
    \toprule
Gaussian (diagonal) & 1.2 $\scriptstyle{[1.1, 1.3]}$ & 0.23 $\scriptstyle{[0.23, 0.23]}$ & 0.55 $\scriptstyle{[0.54, 0.57]}$\\ 
Gaussian (full-rank) & 1.1e+03 $\scriptstyle{[3.8e+02, 2.2e+03]}$ & 0.66 $\scriptstyle{[0.64, 0.75]}$ & 3.2e+05 $\scriptstyle{[1.6e+04, 1.4e+06]}$\\ 
MoG ($K=5$) & 0.91 $\scriptstyle{[0.83, 0.98]}$ & 0.16 $\scriptstyle{[0.16, 0.17]}$ & 0.32 $\scriptstyle{[0.29, 0.35]}$\\ 
MoG ($K=50$) & \textbf{0.30} $\scriptstyle{[0.21, 0.36]}$ & 0.058 $\scriptstyle{[0.057, 0.060]}$ & 0.049 $\scriptstyle{[0.046, 0.051]}$\\ 
\LA & 1.3 & 0.24 & 0.91\\ 
\NNR & \textbf{0.20} $\scriptstyle{[0.12, 0.28]}$ & 0.062 $\scriptstyle{[0.054, 0.073]}$ & 0.047 $\scriptstyle{[0.037, 0.066]}$\\
\VSBQ & \textbf{0.20} $\scriptstyle{[0.20, 0.20]}$ & \textbf{0.037} $\scriptstyle{[0.035, 0.038]}$ & \textbf{0.018} $\scriptstyle{[0.017, 0.018]}$\\
    \toprule
    \end{tabular}
\end{table*}

\subsection{Real-world models}

In this section, we perform experiments on two real-world problems from computational neuroscience, focusing on both noiseless and noisy likelihood evaluations.

\paragraph{Noisy likelihood evaluations}

In many computational models, the likelihood may not be available in closed form, but an estimate of the (log) likelihood may still be obtained via stochastic estimators, yielding a `noisy' likelihood -- or log-likelihood -- evaluation \citep{wood2010statistical, van2020unbiased}. These estimators work by drawing multiple synthetic data samples from the model, and the number of samples or `repetitions' amounts to a hyperparameter governing the precision of the estimate, which trades off with computational complexity \citep{van2020unbiased}. Moreover, these estimates are often approximately normally distributed and approximately -- or exactly -- unbiased  \citep{van2020unbiased, jarvenpaa2021parallel}.
\VSBQ is particularly useful when dealing with noisy log-likelihood evaluations, since the sparse GP can effectively compress a large number of noisy evaluations into a more precise estimate. Notably, many common alternative inference methods are unable to handle noisy evaluations.

In this section, we study the performance and robustness of post-process inference methods (\VSBQ and \NNR) by varying the noise in the log-likelihood evaluations (or observations) in two benchmark problems. A noise standard deviation $\sigma_\text{obs}=0$ corresponds to noiseless target evaluations, obtained through a closed-form or numerical solution of the likelihood. 
A standard deviation $\sigma_\text{obs}$ from 1 to 7 amounts to mild-to-substantial estimation noise in log-likelihood space \citep{acerbi2020variational}, corresponding to cheaper estimates (fewer model samples).
Note that \LA only supports noiseless log-density evaluations ($\sigma_\text{obs} = 0$), and its results are reported for reference.

\paragraph{Bayesian timing model}
We consider a popular Bayesian observer model of time perception from cognitive neuroscience \citep{jazayeri2010temporal,acerbi2012internal,acerbi2020variational}. The key premise of Bayesian observer modeling in perception is that the participant of a psychophysical experiment -- the participant being the system being modeled -- is herself performing Bayesian inference over the sensory stimuli, and employs Bayesian decision theory to report their perception \citep{pouget2013probabilistic, ma2023bayesian}. 

In this specific sensorimotor timing experiment, in each trial human participants had to reproduce the time interval $\tau$ between a mouse click and a screen flash, with $\tau \sim \text{Uniform}$[0.6, 0.975] s \citep{acerbi2012internal}. We assume participants had only access to a noisy sensory measurement $t_\text{s} \sim \mathcal{N}\left(\tau, w_\text{s}^2 \tau^2 \right)$, and their reproduced time $t_\text{m}$ was affected by motor noise, $t_\text{m} \sim \mathcal{N}\left(\tau_\star, w_m^2 \tau_\star^2 \right)$, where $w_\text{s}$ and $w_\text{m}$ are \emph{Weber's fractions}, a psychophysical measure of perceptual and motor variability. We assume participants estimated $\tau_\star$ by combining their sensory likelihood with an approximate Gaussian prior over time intervals, $\normpdf{\tau}{\mu_\text{p}}{\sigma^2_\text{p}}$, and took the mean of the resulting Bayesian posterior. For each trial we also consider a probability $\lambda$ of a `lapse' (e.g., a misclick) producing a response $t_\text{m} \sim \text{Uniform}$[0, 2] s.  Model parameters are $\vtheta = (w_\text{s},w_\text{m},\mu_\text{p},\sigma_\text{p},\lambda)$, so $D=5$. We infer the posterior of a representative participant using published data from \citet{acerbi2012internal}.

\begin{figure*}
    \centering
    \includegraphics[scale=1]{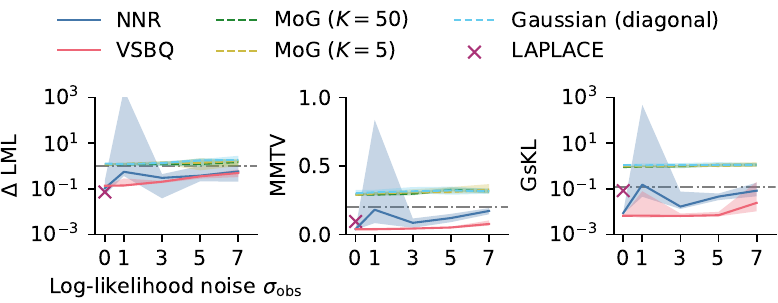}

    \caption{\textbf{Bayesian timing model.} Median $\Delta$LML loss (left), \MMTV (middle), and \gskl (right) as a function of the log-likelihood noise $\sigma_{\text{obs}}$ for the Bayesian Timing model. Shaded areas are 95\% CI of the median and grey dash-dotted horizontal lines are the rule-of-thumb thresholds for good performance ($\Delta$LML $=1$, \MMTV $=0.2$, \gskl $=1/8$). \VSBQ performs well across all noise levels and can outperform (noiseless) \LA even under high log-likelihood noise, whereas \NNR demonstrates less robustness with several failed runs. \BBVI methods exhibit similar performance to each other and are above the metric thresholds.
    }
    \label{fig:timing}
\end{figure*}

As shown in Figure~\ref{fig:timing}, \VSBQ consistently gives a good posterior approximation across different levels of log-likelihood evaluation noise. In this case, even with a large noise $\sigma_{\text{obs}}=7$, \VSBQ surpasses the performance of \LA. Note that \LA is computed based on \emph{noiseless} target evaluations since it does not support noisy evaluations. In this problem, \NNR is considerably less robust, with several failed runs, and performs generally worse compared to \VSBQ. Since the target posterior is close to a Gaussian, the \BBVI methods exhibit similar performance to each other  (as well as across noise levels), all fairly unsatisfactorily hovering above the desired metric thresholds. The insensitivity of \BBVI performance to log-likelihood noise suggests that the variance of the score function estimator is the dominating factor behind the suboptimal results. \BBVI with a Gaussian with full-rank covariance is excluded from the figure, as it yields poor results due to fitting challenges.

Plots of approximate posteriors and the ground-truth posteriors are visualized in Supplementary Material C.6.

    

\paragraph{Multisensory causal inference}

Perceptual causal inference -- inferring whether two sensory cues have the same common source -- comprises a variety of models and tasks of major interest in computational and cognitive neuroscience \citep{kording2007causal,cao2019causal,shams2022bayesian}. Here we consider a visuo-vestibular causal inference experiment representative of this class of models \citep{acerbi2018bayesian,acerbi2020variational}.
In this experiment, participants were seated in a moving chair and asked to determine whether the direction of their movement ($s_\text{vest}$) corresponded to the direction of a looming visual field ($s_\text{vis}$) on a trial by trial basis. It is assumed that the participants can only access noisy sensory measurements, denoted as $z_\text{vest} \sim \mathcal{N}\left(s_\text{vest}, \sigma^2_\text{vest} \right)$ for vestibular information and $z_\text{vis} \sim \mathcal{N}\left(s_\text{vis}, \sigma^2_\text{vis}(c) \right)$ for visual information. Here, $\sigma_\text{vest}$ represents the vestibular noise, while $\sigma_\text{vis}(c)$ represents the visual noise, with $c$ being one of three distinct levels of visual coherence ($c_\text{low}, c_\text{med}, c_\text{high}$) used in the experiment.
To model the participants' responses, we use a heuristic `Fixed' rule, which determines the source to be the same if the absolute difference between the visual and vestibular measurements is less than a threshold $\kappa$, i.e., $|z_\text{vis} - z_\text{vest}| < \kappa$. Additionally, the model incorporates a probability $\lambda$ of the participant providing a random response \citep{acerbi2018bayesian}. The model parameters are $\theta = (\sigma_\text{vis}(c_\text{low}),\sigma_\text{vis}(c_\text{med}),\sigma_\text{vis}(c_\text{high}),\sigma_\text{vest},\lambda, \kappa)$, with a total of $D=6$ parameters. Here we fit data from participant S0 of \citet{acerbi2018bayesian}.

\begin{figure*}
    \centering
    \includegraphics[scale=1]{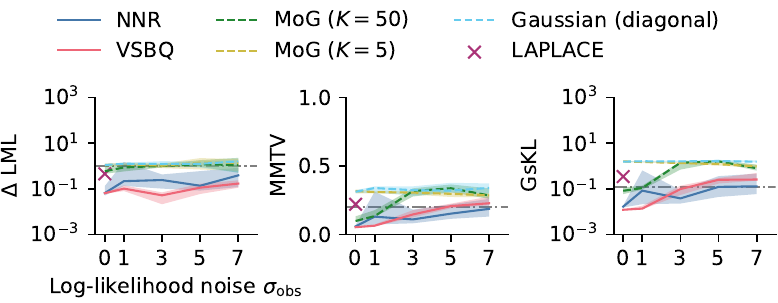}
    \caption{\textbf{Multisensory causal inference model.} Median $\Delta$LML loss (left), \MMTV (middle), and \gskl (right) as a function of the log-likelihood noise $\sigma_{\text{obs}}$ for the multisensory causal inference model. Shaded areas are 95\% CI of the median and grey dash-dotted horizontal lines are the rule-of-thumb thresholds for good performance ($\Delta$LML $=1$, \MMTV $=0.2$, \gskl $=1/8$). \VSBQ performs well across low to moderate noise levels and outperforms (noiseless) \LA even at high noise levels. \NNR also demonstrates strong performance across all noise levels. \BBVI methods perform similarly to each other and are above the thresholds, except for the MoG($K=50$) at low noise levels.}
    \label{fig:multisensory}
\end{figure*}

The performance metrics of all tested methods are plotted in Figure~\ref{fig:multisensory}, as a function of log-likelihood observation noise. \VSBQ consistently outputs a good posterior approximation and only exceeds the desirable metrics thresholds at large observation noise. \LA works reasonably for the noiseless case, still worse than \VSBQ, and slightly above the thresholds for \MMTV and \gskl. In this problem, \NNR performs well, slightly outperforming \VSBQ in some metrics in a higher observation noise regime. \BBVI with MoG ($K=50$) performs well for low noise cases and becomes similar to \BBVI with a diagonal Gaussian or MoG ($K=5$) for high noise levels. \BBVI with a full-rank Gaussian is excluded from the figure, as it yields poor results due to optimization challenges also in this case. We visualize the approximate and ground-truth posteriors in Supplementary Material C.6.

\subsection{Additional analyses}

We summarize here the results of a number of additional analyses, with further details reported in the Supplementary Material.

\paragraph{MAP estimation with Bayesian Adaptive Direct Search (BADS)} 

We ran again all our experiments using optimization traces from a different optimization algorithm based on a hybrid Bayesian optimization approach \citep{garnett2023bayesian}, namely Bayesian Adaptive Direct Search (\bads;  \citealp{acerbi2017practical}). \bads is a state-of-the-art optimization algorithm with wide application in computational neuroscience and other fields.
We found the performance of \VSBQ with \bads is almost identical to \cmaes for three out of four benchmark problems (Two Moons, multivariate Rosenbrock-Gaussian, and Bayesian timing model), and slightly worse for the multisensory causal inference problem. See Supplementary Material C.5 for the full results and their discussion.

\paragraph{Runtime analysis}

Ideally, running a post-process inference method should only take a relatively short time (e.g., a few minutes), so we performed a detailed comparison of the runtimes of different algorithms on different computational architectures (CPU and GPU). Overall, we found that \VSBQ takes several minutes on CPU and 1-3 minutes on GPU across the various problems we considered, meeting the speed desiderata of a post-process technique. \NNR can take considerably longer, mainly due to multiple training runs required for hyperparameter selection. \BBVI methods are not intended as post-processing approaches, making their runtime less directly relevant. The runtime of \BBVI and Laplace approximation methods depends on the cost and the number of target density evaluations. Full results are reported in Supplementary Material C.7.

\begin{figure*}
    \centering
    \subcaptionbox{Bayesian timing model ($\sigma_{\text{obs}}=3$).}{
        \includegraphics[scale=1]{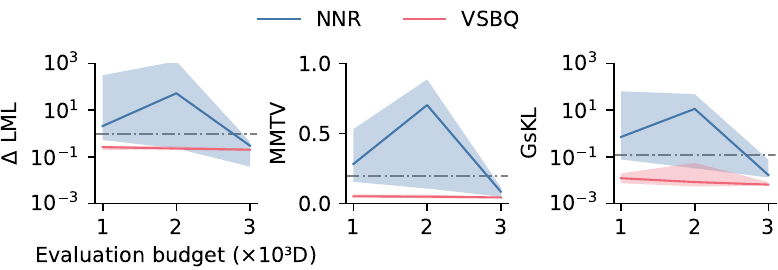}
    } \\
    \subcaptionbox{Multisensory causal inference model ($\sigma_{\text{obs}}=0$).}{
        \includegraphics[scale=1]{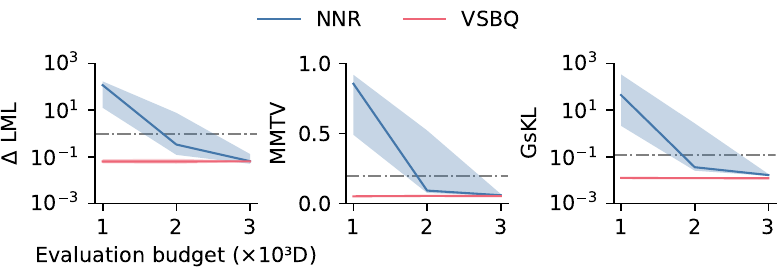}
    } \\
    \caption{\textbf{Sensitivity to the number of target evaluations.} Median $\Delta$LML loss (left), \MMTV (middle), and \gskl (right) as a function of the number of target evaluations, for two benchmark problems. Shaded areas are 95\% CI of the median and grey dash-dotted horizontal lines are the rule-of-thumb thresholds for good performance ($\Delta$LML $=1$, \MMTV $=0.2$, \gskl $=1/8$). Across both tasks, \VSBQ demonstrates better robustness to the number of evaluations compared to \NNR.}
    \label{fig:varying_N_evaluations}
\end{figure*}

\paragraph{Sensitivity to the number of target evaluations}

The number of target density evaluations is a critical factor in determining coverage over the posterior distribution, and therefore, would naturally impact the performance of both \NNR and \VSBQ. 
Throughout the experiments, we used a fixed number of target density evaluations, $3000D$, which corresponds to at least two complete MAP optimization trajectories for all benchmark problems.
To investigate the sensitivity of our method to the number of target density evaluations $N$, we further conducted experiments across three evaluation budgets: $N \in \{1000D, 2000D, 3000D\}$, where $D$ is the dimensionality of the parameter space. 

Figure~\ref{fig:varying_N_evaluations} summarizes the results on two representative benchmarks: the noisy Bayesian timing model ($\sigma_{\text{obs}}=3$) and the multisensory causal inference model ($\sigma_{\text{obs}}=0$). We find that \VSBQ is robust to changes in the number of evaluations, whereas \NNR is more sensitive and tends to perform significantly worse with fewer evaluations. Additional results for the \bads optimizer are reported in Supplementary Material Section~C.5. In contrast to \cmaes, \bads can lead to poorer coverage of the posterior, as discussed in Supplementary Section~C.5, making \VSBQ more sensitive to the number of evaluations in this case. Nonetheless, \VSBQ consistently outperforms \NNR across most tested settings.

\paragraph{Posterior estimation: MCMC or variational inference?} In the earlier paragraphs, we presented results obtained by approximating the target posterior density from the surrogate of the log density (Figure~\ref{fig:density_modeling}) via variational inference. A natural alternative is to instead run \MCMC on the surrogate log-density \citep{rasmussen2003gaussian,nemeth2018merging, jarvenpaa2021parallel, gammalFastRobustBayesian2022}. In Supplementary Material C.8, we provide experimental results showing that \MCMC can underperform in this setting and discuss the reasons.

\section{Discussion}
\label{sec:discussion}

In this paper, we introduced the framework of post-process, black-box Bayesian inference, and proposed a specific post-process algorithm, Variational Sparse Bayesian Quadrature (\VSBQ). By recycling evaluations from previous MAP optimization runs, \VSBQ enables full Bayesian inference at a limited additional cost.  In this section, we first discuss why surrogate-based approaches like \VSBQ can be more effective than \BBVI, followed by an exploration of the limitations of our method and potential directions for future work.

\subsection{Why are surrogate-based approaches more effective than direct variational inference?}
As demonstrated across a series of benchmarks, in our black-box setting surrogate-based methods like \VSBQ and \NNR produce high-quality solutions, whereas \BBVI -- which performs black-box variational inference directly on the target -- struggles to effectively fit the target posterior, even when given \emph{$10\times$} target density evaluations. This disparity arises primarily from the high variance associated with the score function estimator in \BBVI. While gradient-based variational inference using the reparameterization trick (such as automatic differentiation variational inference, or \ADVI) empirically reduces variance and is more widely adopted~\citep{titsiasDoublyStochasticVariational2014,kucukelbirAutomaticDifferentiationVariational2017,xuVarianceReductionProperties2019}, it requires the target density to be differentiable with respect to the parameters. The black-box nature of the target model makes \ADVI inapplicable in our considered scenarios.
Surrogate-based methods address these challenges effectively. Fitting a surrogate model (e.g., a sparse Gaussian process for \VSBQ or a neural network for \NNR) to the target density offers two key advantages:
\begin{enumerate}[nosep]
    \item[a.] These methods can fully leverage all existing target density evaluations from various sources, effectively interpolating and smoothing across (noisy) observations.
    \item[b.] The surrogate model resolves the non-differentiability issue, allowing gradient-based variational inference to be applied to the surrogate density, which is differentiable (e.g., via automatic differentiation) by construction.
\end{enumerate}

\subsection{Limitations and future work}

Arguably, post-process inference via \VSBQ is constrained to low-dimensional problems (e.g., up to ten parameters) due to several reasons. As the dimension increases, pre-existing evaluations from MAP optimizations become less likely to cover the majority of the posterior mass~\citep[Chapter~3.3.3]{vershyninHighdimensionalProbabilityIntroduction2018}, while sufficient coverage of the posterior is essential for \VSBQ to perform well. Moreover, the MAP estimate may become not meaningful or even ill-defined in certain settings, e.g., when the likelihood is unbounded~\citep{gelman2013bayesian}. While a larger number of evaluations might expand the range of tractable dimensions, the curse of dimensionality poses challenges in approximating a function without additional structural assumptions. A potential solution to address the issue of lack of coverage is active sampling, that is, the ability to acquire new log-density evaluations where needed~\citep{bliznyukBayesianCalibrationUncertainty2008,osborne2012active, acerbi2020variational, jarvenpaa2021parallel, desouza2022parallel}. Active sampling would need one or multiple rounds of ad-hoc function evaluations and subsequent surrogate model updates, which would substantially increase the post-processing time~\citep{desouza2022parallel}. Finding efficient methods for active learning for post-process inference is left for future research.
Despite the above limitations, the relatively cheap cost of our method makes it still valuable for quickly constructing a tractable initial approximation of the posterior, potentially useful to inform subsequent runs of \MCMC or other inference methods~\citep{zhangPathfinderParallelQuasiNewton2022}. 

An important feature of approximate inference methods is the availability of diagnostics to assess the reliability of inference \citep{vehtariRankNormalizationFoldingLocalization2021, yao2018yes}. A general-purpose inference diagnostic consists of posterior-predictive checks, i.e., testing that synthetic data generated from parameters sampled from the posterior are compatible with the actual data \citep{gelman2013bayesian}.
As an additional diagnostic, \VSBQ provides the standard deviation of the \ELBO, $\ELBO_\text{sd}$, which can be calculated via sparse Bayesian quadrature (Eq.~\ref{eq:cov_sbq}). $\ELBO_\text{sd}$ reflects the uncertainty of the sparse GP prediction in regions where the variational posterior has non-negligible mass. Therefore, it can serve as a useful diagnostic in that solutions with $\ELBO_\text{sd} \gg 1$ should not be trusted. However, as is often the case with inference diagnostics, a small $\ELBO_\text{sd}$ does \emph{not} guarantee the validity of the approximate variational posterior~\citep{acerbi2018variational}. As a practical validation approach, we recommend running \VSBQ multiple times by dropping for example 20\% of the training set and checking the consistency of the approximate posteriors via a form of cross-validation. Visualization of the posterior together with the locations of the evaluated points is also helpful for validating that the approximate posterior is supported by actual evaluations, as opposed to escaping from the training points region, leading to `hallucinated' posterior regions \citep{desouza2022parallel}. We further discuss this `hallucination' problem in Supplementary Material C.8. Finally, if additional exact log-density evaluations are possible, one can leverage Pareto smoothed importance sampling \citep{vehtariParetoSmoothedImportance2024} for correcting and validating the approximate posterior~\citep{yao2018yes}.

Noise shaping, a principled heuristic introduced in Section \ref{sec:noiseshaping}, is an important component of \VSBQ. Noise shaping effectively downweighs the low-density observations, providing a straightforward probabilistic explanation. This approach is closely related to, but distinct from, the weighted KL divergence method in \citet{mcintire2016sparse} and the inducing points allocation strategy in \citet{mossInducingPointAllocation2023}. We chose the noise shaping function out of theoretical and empirical considerations (see Supplementary Material C.1), and further work is needed to make it a general tool for surrogate modeling, e.g., via adaptive techniques, and to provide a sounder theoretical grounding.


Finally, in this work we explored sparse GP surrogates via \SGPR due to its numerical convenience (i.e., closed-form solutions for the sparse GP posterior), but \SGPR is also somewhat limited in scalability and restricted to Gaussian observations. A natural extension of our work would consist of extending \VSBQ to \emph{stochastic} variational GPs (\SVGP; \citealp{hensman2013gaussian,hensman2015scalable}), which are able to handle nearly arbitrarily large datasets and non-Gaussian observations of the target density. Additionally, as our experiments with neural network regression (\NNR) suggested, deep neural networks can be competitive surrogates for log-density function modeling in the regime of large datasets. Exploring better regularization and uncertainty quantification for neural networks~\citep{daxbergerLaplaceReduxEffortless2021,immerEffectiveBayesianHeteroscedastic2023} is a promising future direction with the potential to enhance the effectiveness of deep learning within the framework of post-process inference.

\subsection{Conclusions}

In this paper, we showed the application of post-process approximate Bayesian inference and \VSBQ as a valuable tool for quickly constructing a posterior approximation at a low cost by recycling existing log-density evaluations. With further developments in diagnostics, theoretical analysis, and scalability, the framework of post-process inference has the potential to make Bayesian inference more accessible and efficient for a wide range of applications.

\section*{Acknowledgments}

This work was supported by the Research Council of Finland Flagship programme: Finnish Center for Artificial Intelligence FCAI.  The authors wish to thank the Finnish Computing Competence Infrastructure (FCCI) for supporting this project with computational and data storage resources.

Gr\'egoire Clart\'e completed part of the work when he was at the Department of Computer Science, University of Helsinki, Finland, where he was supported by the Research Council of Finland Flagship programme: Finnish Center for Artificial Intelligence FCAI.

Chengkun Li, Martin Jørgensen and Luigi Acerbi were supported by the Research Council of Finland (grants number 356498 and 358980 to Luigi Acerbi).

\section*{Declarations}

\subsection*{Supplementary information}

The supplementary material contains mathematical proofs, implementation details, additional results, and extended explanations omitted from the main text. 

\subsection*{Data availability}
The data for benchmark problems and an implementation of our algorithm are available at \url{https://github.com/acerbilab/vsbq}.

\subsection*{Author contribution}
Conceptualization: C.L, G.C, L.A; 
Methodology: C.L, G.C, L.A;
Theoretical analysis: C.L, G.C;
Experiment design: all authors;
Experiment investigation: C.L;
Writing: C.L, G.C, L.A; 
Review and editing: all authors;
Supervision: L.A.

\bibliographystyle{plainnat}
\bibliography{vsbq}

\setcounter{footnote}{0}
\setcounter{figure}{0}
\setcounter{table}{0}
\setcounter{equation}{0}
\renewcommand{\theequation}{S\arabic{equation}}
\renewcommand{\thetable}{S\arabic{table}}
\renewcommand{\thefigure}{S\arabic{figure}}

\clearpage

\part*{Supplementary Material}
\appendix
\section{Analytical formulae}
\label{supp_sec:formulae}
In this section, we provide analytical formulae and derivations omitted from the main paper. In addition to the notation used in the main text, we denote here with $\mathbf{f}$ the vector $f(\xx)$, and $\m_\u$ and $\mathbf{R}_{\u\u}$ are, respectively, the mean and covariance matrix of the optimal variational distribution at inducing points $\zz$, summarized by $\vgp$. We use $\kappa$ for the Gaussian process (GP) kernel, and the following notations for the matrices $\K_{\xx,\xx} \equiv \kappa(\xx,\xx)$ (and similarly for $\zz$). We write the matrix of observation noise at each point of $\xx$ as $\S = \text{diag}\left(\sigma^2_\text{obs}(\xx)\right) = \text{diag} \left( s_n^2\right)$, $n = 1, 2, \cdots, N$, where $N$ is the number of training points.

\subsection{Optimal variational parameters, heteroskedastic case}
\label{supp_sec:SGPR-derivation}
Here we describe how to derive the optimal variational parameters $\m_\u$ and $\mathbf{R}_{\u\u}$ for the heteroskedastic observation noise case. Our derivations mostly follow \citet{buiPaperVariationalLearning}. The only difference is that we consider the heteroskedastic case instead of the homoskedastic case.

\paragraph{Zero-mean case.}
First, we compute the optimal parameters for a sparse GP with a zero-mean function. From \citet{buiPaperVariationalLearning}, the quantities that need to be adapted for heteroskedastic noise are $\mathcal{M}(\y, \u)$ and $H(\mathbf{y},\mathbf{u}) \equiv \exp \mathcal{M}(\y, \u)$. 
\begin{align*}
\mathcal{M}(\y, \u) & = \int p(\mathbf{f} \mid \textbf{u})\log p(\mathbf{y} \mid \mathbf{f}) \mathrm{d}\mathbf{f} \\
 & = \int \mathcal{N}\left( \mathbf{f};\K_{\xx,\zz}\K_{\zz,\zz}^{- 1}\textbf{u} ,\K_{\xx,\xx} - \K_{\xx,\zz}\K_{\zz,\zz}^{- 1}\K_{\zz,\xx} \right)\log\left\lbrack \mathcal{N}\left( \mathbf{y};\mathbf{f},\mathbf{\mathbf{S}} \right) \right\rbrack \mathrm{d}\mathbf{f}\\
 &\hspace{-1cm} \text{(Define $\mathbf{A}=\K_{\xx,\zz}\K_{\zz,\zz}^{- 1}\textbf{u} $ and $\mathbf{B}=\K_{\xx,\xx} - \K_{\xx,\zz}\K_{\zz,\zz}^{- 1}\K_{\zz,\xx}$ for brevity.)} \\
 & = \int \mathcal{N}(\mathbf{f};\mathbf{A},\mathbf{B})\left\lbrack - \frac{N}{2}\log(2\pi) - \frac{1}{2}\sum\limits_{n = 1}^{N}\log s_n^{2} - \frac{1}{2}(\mathbf{y} - \mathbf{f})^{\top}\mathbf{\mathbf{S}}^{- 1}(\mathbf{y} - \mathbf{f}) \right\rbrack \mathrm{d}\mathbf{f}\\
 & = \int \mathcal{N}(\mathbf{f};\mathbf{A},\mathbf{B})\left\lbrack - \frac{N}{2}\log(2\pi) - \frac{1}{2}\sum\limits_{n = 1}^{N}\log s_n^{2} - \frac{1}{2}{\Tr}\left( \left( \mathbf{y}\mathbf{y}^{\top} - 2\mathbf{y}\mathbf{f}^{\top} + {\mathbf{f}\mathbf{f}}^{\top} \right)\mathbf{\mathbf{S}}^{- 1} \right) \right\rbrack \mathrm{d}\mathbf{f}\\
 & = - \frac{N}{2}\log(2\pi) - \frac{1}{2}\sum\limits_{n = 1}^{N}\log s_n^{2} - \frac{1}{2}{\Tr}\left( \left( \mathbf{y}\mathbf{y}^{\top} - 2\mathbf{y}\mathbf{A}^{\top} + \mathbf{A}\mathbf{A}^{\top} + \mathbf{B} \right)\mathbf{\mathbf{S}}^{- 1} \right) \\
 & = - \frac{1}{2}{\Tr}(\mathbf{B}\mathbf{\mathbf{S}}^{- 1}) + \log\left\lbrack \mathcal{N}\left( \mathbf{y};\mathbf{A},\mathbf{\mathbf{S}} \right) \right\rbrack. 
\end{align*}
According to \citet{buiPaperVariationalLearning}, the optimal variational distribution at inducing points $\zz$ is:
\begin{equation*}
\begin{split}
{\tilde{p}(\textbf{u})} \propto & \; { p(\textbf{u})\exp(\mathcal{M}(\y, \u))} \\
 \propto & \; { \exp\left\lbrack - \frac{1}{2}\u^{\top}(\K_{\zz,\zz}^{- 1}\K_{\zz,\xx}\mathbf{\mathbf{S}}^{- 1}\K_{\xx,\zz}\K_{\zz,\zz}^{- 1} + \K_{\zz,\zz}^{- 1})\mathbf{u} + \mathbf{u}^{\top}\K_{\zz,\zz}^{- 1}\K_{\zz,\xx}\mathbf{\mathbf{S}}^{- 1}\mathbf{y} \right\rbrack} \\
 = & \; { \mathcal{N}(\K_{\zz,\zz}\bf{\Sigma} \K_{\zz,\xx}\mathbf{\mathbf{S}}^{- 1}\mathbf{y},\K_{\zz,\zz}\bf{\Sigma} \K_{\zz,\zz})} \\
 \triangleq & \; \mathcal{N}(\m_\u,\mathbf{R}_{\u\u}),
\end{split}
\end{equation*}
where $\bf{\Sigma} = (\K_{\zz,\xx}\mathbf{\mathbf{S}}^{- 1}\K_{\xx,\zz} + \K_{\zz,\zz})^{- 1}$.
The evidence lower bound is $\text{GP-\ELBO}=\log(\mathcal{Z})$, where:
\begin{align*}
\mathcal{Z} & {= \int H(\mathbf{y},\mathbf{u})p(\mathbf{u})\mathrm{d}\mathbf{u}} \\
 & {= \int\exp\left( - \frac{1}{2}{\Tr}(\mathbf{B}\mathbf{\mathbf{S}}^{- 1}) \right)\mathcal{N}\left( \mathbf{y};\mathbf{A},\mathbf{\mathbf{S}} \right)\mathcal{N}\left( \mathbf{u};\mathbf{0},\K_{\zz,\zz} \right)\mathrm{d}\mathbf{u}} \\
 & {= \mathcal{N}(\mathbf{y};\mathbf{0},\mathbf{\mathbf{S}} + \K_{\xx,\zz}\K_{\zz,\zz}^{- 1}\K_{\zz,\xx})\exp\left( - \frac{1}{2}{\Tr}(\mathbf{B}\mathbf{\mathbf{S}}^{- 1}) \right)}.
\end{align*}
Thus, the \ELBO writes:
\begin{align*}
\text{GP-\ELBO}& {= \log \mathcal{Z}} \\
 & {= \log\mathcal{N}(\mathbf{y};\mathbf{0},\mathbf{\mathbf{S}} + \K_{\xx,\zz}\K_{\zz,\zz}^{- 1}\K_{\zz,\xx}) - \frac{1}{2}{\Tr}(\mathbf{B}\mathbf{\mathbf{S}}^{- 1})}. 
\end{align*}

\paragraph{Non-zero mean case.} \label{supp_par:non-zero-mean}

In the case of a non-zero mean function, we have:
\[\mathbf{A} = m(\xx) + \K_{\xx,\zz}\K_{\zz,\zz}^{- 1}(\mathbf{u} -m(\zz)),\]
which leads to:
\begin{align*}
\tilde{p}(\textbf{u}) & \propto p(\textbf{u})\exp(\mathcal{M}) \\
 & \propto \exp\left\lbrack - \frac{1}{2}\mathbf{u}^{\top}(\K_{\zz,\zz}^{- 1}\K_{\zz,\xx}\mathbf{\mathbf{S}}^{- 1}\K_{\xx,\zz}\K_{\zz,\zz}^{- 1} + \K_{\zz,\zz}^{- 1})\mathbf{u} \right. \\
&  \hspace{1cm}+\mathbf{u}^{\top}\lbrack \K_{\zz,\zz}^{- 1}\K_{\zz,\xx}\mathbf{\mathbf{S}}^{- 1}(\mathbf{y} - m(\xx) + \K_{\xx,\zz}\K_{\zz,\zz}^{- 1}m(\zz)) + \K_{\zz,\zz}^{- 1}m(\zz)\rbrack \biggr]\\
 & = \mathcal{N}(\K_{\zz,\zz} \mathbf{\Sigma} \lbrack \K_{\zz,\xx}\mathbf{\mathbf{S}}^{- 1}(\mathbf{y} - m(\xx) + \K_{\xx,\zz}\K_{\zz,\zz}^{- 1}m(\zz)) + m(\zz)\rbrack,\K_{\zz,\zz} \mathbf{\Sigma}  \K_{\zz,\zz}), 
\end{align*}
and
\begin{align}
\label{supp_eq:GP-ELBO-nonzero}
\begin{aligned}
\text{GP-\ELBO} 
 & = \log\mathcal{N}(\mathbf{y};m(\xx),\mathbf{\mathbf{S}} + \K_{\xx,\zz}\K_{\zz,\zz}^{- 1}\K_{\zz,\xx}) - \frac{1}{2}{\Tr}(\mathbf{B}\mathbf{\mathbf{S}}^{- 1}). \\
\end{aligned}
\end{align}

\subsection{Numerical implementation of the \text{GP-ELBO}}\label{marginal-likelihood-bound}
\label{supp_sec:numerical-ELBO}
In this section, we derive formulae for efficient and numerically stable computation of the \text{GP-\ELBO} \citep{GPflow2017} in the heteroskedastic case. 
We first define $\Q_{\xx,\xx} \equiv \K_{\xx,\zz}\K_{\zz,\zz}^{- 1}\K_{\zz,\xx}$. Then, Eq. \ref{supp_eq:GP-ELBO-nonzero} can be written as:
\begin{align*}
\text{GP-\ELBO} = \log\mathcal{N}(\mathbf{y}; m(\xx),\Q_{\xx,\xx} + \mathbf{\mathbf{S}}) - \frac{1}{2}{\Tr}((\K_{\xx,\xx} - \Q_{\xx,\xx})\mathbf{\mathbf{S}}^{- 1}).
\end{align*}
To obtain an efficient and stable evaluation of the \text{GP-\ELBO}, we apply the Woodbury identity to the effective
covariance matrix:
\[\lbrack \Q_{\xx,\xx} + \mathbf{\mathbf{S}}\rbrack^{- 1} = \mathbf{\mathbf{S}}^{- 1} - \mathbf{\mathbf{S}}^{- 1}\K_{\xx,\zz}\lbrack \K_{\zz,\zz} + \K_{\zz,\xx}\mathbf{\mathbf{S}}^{- 1}\K_{\xx,\zz}\rbrack^{- 1}\K_{\zz,\xx}\mathbf{\mathbf{S}}^{- 1}.\]
To obtain a better conditioned matrix for inversion, we introduce in the previous formula the matrix $\mathbf{L}$, the Cholesky decomposition of  $\K_{\zz,\zz}$, i.e., $\mathbf{L}\mathbf{L}^{\top} = \K_{\zz,\zz}$:
\begin{align*}
\lbrack \Q_{\xx,\xx} + \mathbf{\mathbf{S}}\rbrack^{- 1} & = \mathbf{\mathbf{S}}^{- 1} - \mathbf{\mathbf{S}}^{- 1}\K_{\xx,\zz}{{\mathbf{L}^{- \top}\mathbf{L}^{\top}\lbrack \K_{\zz,\zz} + \K_{\zz,\xx}\mathbf{\mathbf{S}}^{- 1}\K_{\xx,\zz}\rbrack^{- 1}{{\mathbf{L}\mathbf{L}^{- 1}}\K_{\zz,\xx}}}\mathbf{\mathbf{S}}^{- 1}} \\
 & = \mathbf{\mathbf{S}}^{- 1} - \mathbf{\mathbf{S}}^{- 1}\K_{\xx,\zz}{{\mathbf{L}^{- \top}\lbrack \mathbf{L}^{- 1}
 ( \K_{\zz,\zz} + \K_{\zz,\xx}\mathbf{\mathbf{S}}^{- 1}\K_{\xx,\zz})
 \mathbf{L}^{- \top}\rbrack^{- 1}\mathbf{L}^{- 1}\K_{\zz,\xx}}\mathbf{\mathbf{S}}^{- 1}} \\
 & = \mathbf{\mathbf{S}}^{- 1} - \mathbf{\mathbf{S}}^{- 1}\K_{\xx,\zz}{{\mathbf{L}^{- \top}\lbrack(\mathbf{I} + \mathbf{L}^{- 1}(\K_{\zz,\xx}\mathbf{\mathbf{S}}^{- 1}\K_{\xx,\zz})\mathbf{L}^{- \top}\rbrack^{- 1}\mathbf{L}^{- 1}\K_{\zz,\xx}}\mathbf{\mathbf{S}}^{- 1}}.
\end{align*}
For notational convenience, we define
$\mathbf{U} \equiv \mathbf{L}^{- 1}\K_{\zz,\xx}\mathbf{\mathbf{S}}^{- \frac{1}{2}}$,
and
$\mathbf{V} \equiv \lbrack\mathbf{I} + \mathbf{U}\mathbf{U}^{\top}\rbrack $:
\[{\lbrack \Q_{\xx,\xx} + \mathbf{\mathbf{S}}\rbrack^{- 1}} = \mathbf{\mathbf{S}}^{- 1} - \mathbf{\mathbf{S}}^{- \frac{1}{2}}\mathbf{U}^{\top}\mathbf{V}^{- 1}\mathbf{U}\mathbf{\mathbf{S}}^{- \frac{1}{2}}.\]
By the matrix determinant lemma, we have:
\begin{align*}
|\Q_{\xx,\xx} + \mathbf{\mathbf{S}}| & = |\K_{\zz,\zz} + \K_{\zz, \xx}\mathbf{\mathbf{S}}^{- 1}\K_{\xx, \zz}|\,|\K_{\zz,\zz}^{- 1}|\,|\mathbf{\mathbf{S}}| \\
 & = |{\mathbf{L}\mathbf{L}}^{\top} + \K_{\zz,\xx}\mathbf{\mathbf{S}}^{- 1}\K_{\xx,\zz}|\,|\mathbf{L}^{- \top}|\,|\mathbf{L}^{- 1}|\,|\mathbf{\mathbf{S}}| \\
 & = |\mathbf{I} + \mathbf{L}^{- 1}\K_{\zz,\xx}\mathbf{\mathbf{S}}^{- 1}\K_{\xx,\zz}\mathbf{L}^{- \top}|\,|\mathbf{\mathbf{S}}| \\
 & = |\mathbf{V}||\mathbf{\mathbf{S}}|.
\end{align*}
With these two definitions, the \text{GP-\ELBO} can be written as:
\begin{align*}
\mathcal{L} & = \log\mathcal{N}(\mathbf{y};m(\xx),\Q_{\xx,\xx} + \mathbf{\mathbf{S}}) - \frac{1}{2}{\Tr}((\K_{\xx,\xx} - \Q_{\xx,\xx})\mathbf{\mathbf{S}}^{- 1}) \\
 & = - \frac{N}{2}\log{2\pi} - \frac{1}{2}\log|\Q_{\xx,\xx} + \mathbf{\mathbf{S}}| - \frac{1}{2}\mathbf{\bar{y}}^{\top}\lbrack \Q_{\xx,\xx} + \mathbf{\mathbf{S}}\rbrack^{- 1}\mathbf{\bar{y}} \\
 & \quad - \frac{1}{2}{\Tr}((\K_{\xx,\xx} - \Q_{\xx,\xx})\mathbf{\mathbf{S}}^{- 1}) \\
 & = - \frac{N}{2}\log{2\pi} - \frac{1}{2}\log|\mathbf{V}||\mathbf{\mathbf{S}}| - \frac{1}{2}\mathbf{\bar{y}}^{\top}(\mathbf{\mathbf{S}}^{- 1} - \mathbf{\mathbf{S}}^{- \frac{1}{2}}\mathbf{U}^{\top}\mathbf{V}^{- 1}\mathbf{U}\mathbf{\mathbf{S}}^{- \frac{1}{2}})\mathbf{\bar{y}} \\ 
 & \quad - \frac{1}{2}{\Tr}((\K_{\xx,\xx} - \Q_{\xx,\xx})\mathbf{\mathbf{S}}^{- 1}) \\
& = - \frac{N}{2}\log{2\pi} - \frac{1}{2}\log|\mathbf{V}| - \frac{1}{2}\log|\mathbf{\mathbf{S}}| - \frac{1}{2}\mathbf{\bar{y}}^{\top}\mathbf{\mathbf{S}}^{- 1}\mathbf{\bar{y}} - \frac{1}{2}c^\top c \\
& \quad - \frac{1}{2}{\Tr}(\K_{\xx,\xx}\mathbf{\mathbf{S}}^{- 1}) + \frac{1}{2}{\Tr}(\mathbf{U}\mathbf{U}^{\top}),
\end{align*}
where $\mathbf{\bar{y}} \equiv \mathbf{y} - m(\xx)$ and we have defined  $c \equiv \mathbf{L}_{\mathbf{V}}^{- 1}\mathbf{U}\mathbf{\mathbf{S}}^{- \frac{1}{2}}\mathbf{\bar{y}}$,  with the Cholesky decomposition $\mathbf{L}_{\mathbf{V}}\mathbf{L}_{\mathbf{V}}^{\top} = \mathbf{V}$.
%

\subsection{Predictive distribution of SGPR}
\label{supp_sec:prediction}

In this section, we derive the predictive latent distribution of \SGPR and its numerically stable implementation, given the variational GP posterior $\vgp$, i.e., $p(\f^\star | \vgp)$, from a sparse GP with heteroskedastic observation noise. From Section \ref{supp_sec:SGPR-derivation}, the optimal variational distribution $\tilde{p}$ on $\u$ writes:
\[\tilde{p}(\mathbf{u}) = \mathcal{N}(\mathbf{u}\mid\mathbf{m}_\mathbf{u},\mathbf{R}_{\mathbf{uu}}),\]
with:
\begin{align*}
\mathbf{R}_{\mathbf{uu}} &= \K_{\zz,\zz} (\K_{\zz,\xx}\mathbf{\mathbf{S}}^{- 1}\K_{\xx,\zz} + \K_{\zz,\zz})^{-1}\K_{\zz,\zz} \\
    \mathbf{R}_{\mathbf{uu}}^{-1} &= \K_{\zz,\zz}^{- 1} + \K_{\zz,\zz}^{- 1}\K_{\zz,\xx}\mathbf{\mathbf{S}}^{- 1}\K_{\xx,\zz}\K_{\zz,\zz}^{- 1} 
\end{align*} 
\[
{\m_\u = \mathbf{R}_{\mathbf{uu}}\K_{\zz,\zz}^{- 1}\lbrack \K_{\zz,\xx}\mathbf{\mathbf{S}}^{- 1}(\mathbf{y} - m(\xx) + \K_{\xx,\zz}\K_{\zz,\zz}^{- 1}m(\zz)) + m(\zz)\rbrack}.
\]

The predictive distribution at $\x^\star$ is:
\[p(\mathbf{f}^{\star}| \vgp) = \int p(\mathbf{f}^{\star}\,|\,\mathbf{u}) \tilde{p}(\mathbf{u})\text{d}\mathbf{u},\]
with:
\[p(\mathbf{f}^{\star}\,|\,\mathbf{u}) = \mathcal{N}(\mathbf{f}^{\star}\,|\,m(\x^\star) + \K_{\star \zz}\K_{\zz,\zz}^{- 1}(\mathbf{u} - m(\zz)),\,\K_{\star \star} - \K_{\star \zz}\K_{\zz,\zz}^{- 1}\K_{\zz \star});\]
therefore:
\begin{align*}
    p(\mathbf{f}^{\star}|\vgp) = \mathcal{N}(\mathbf{f}^{\star}\, | & \,m(\x^\star) - \K_{\star \zz}\K_{\zz,\zz}^{- 1}m(\zz) + \K_{\star \zz}\K_{\zz,\zz}^{- 1}\mathbf{m}_\u, \\
    & \K_{\star \star} - \K_{\star \zz}\K_{\zz,\zz}^{- 1}\K_{\zz \star} + \K_{\star \zz}\K_{\zz,\zz}^{- 1}\mathbf{R}_{\mathbf{uu}}\K_{\zz,\zz}^{- 1}\K_{\zz \star}).
\end{align*}
%
For the numerical implementation, we define the same notations as in Section \ref{supp_sec:numerical-ELBO}: $\mathbf{L}\mathbf{L}^{\top} = \K_{\zz,\zz}$, $\mathbf{U} \equiv \mathbf{L}^{- 1}\K_{\zz,\xx}\mathbf{\mathbf{S}}^{- \frac{1}{2}}$,  $\mathbf{V} \equiv \lbrack\mathbf{I} + \mathbf{U}\mathbf{U}^{\top}\rbrack $, and $\mathbf{c} \equiv \mathbf{L}_{\mathbf{V}}^{- 1}\mathbf{U}\mathbf{\mathbf{S}}^{- \frac{1}{2}}(\mathbf{y} - m(\xx))$,  with the Cholesky decomposition $\mathbf{L}_{\mathbf{V}}\mathbf{L}_{\mathbf{V}}^{\top} = \mathbf{V}$. This leads to:

\[\K_{\zz,\zz}^{- 1}\mathbf{R}_{\mathbf{uu}}\K_{\zz,\zz}^{- 1} = \mathbf{L}^{- \top}\mathbf{V}^{- 1}\mathbf{L}^{- 1},\]
and further:
\[\K_{\zz,\zz}^{- 1}\m_\u = \mathbf{L}^{- \top}\mathbf{L}_{\mathbf{V}}^{- \top}\mathbf{c} + \mathbf{L}^{- \top}\mathbf{L}_{\mathbf{V}}^{- \top}\mathbf{L}_{\mathbf{V}}^{- 1}\mathbf{U}\mathbf{U}^{\top}\mathbf{L}^{- 1}m(\zz) + \mathbf{L}^{- \top}\mathbf{L}_{\mathbf{V}}^{- \top}\mathbf{L}_{\mathbf{V}}^{- 1}\mathbf{L}^{- 1}m(\zz).\]
Finally, we obtain:
\begin{align*}
p(\mathbf{f}^{\star}|\vgp) = \mathcal{N}(\mathbf{f}^{\star} \mid & m(\x^\star) + \K_{\star\zz}(\mathbf{L}^{- \top}\mathbf{L}_{\mathbf{V}}^{- \top}\mathbf{c} + \mathbf{L}^{- \top}\mathbf{L}_{\mathbf{V}}^{- \top}\mathbf{L}_{\mathbf{V}}^{- 1}\mathbf{U}\mathbf{U}^{\top}\mathbf{L}^{- 1}m(\zz) + \\
& \hspace{1cm}\mathbf{L}^{- \top}\mathbf{L}_{\mathbf{V}}^{- \top}\mathbf{L}_{\mathbf{V}}^{- 1}\mathbf{L}^{- 1}m(\zz) - \mathbf{L}^{-\top}\mathbf{L}^{-1}m(\zz)  ),\\
&\K_{\star \star} - \K_{\star\zz}\mathbf{L}^{- \top}(\mathbf{I} - \mathbf{V}^{- 1})\mathbf{L}^{- 1}\K_{\zz\star}). 
\end{align*}

\paragraph{Sanity check with $\zz = \xx$.}

\label{supp_par:sanity-check-with-u-f}

If $\zz = \xx$, i.e., all training points are selected as inducing points, the \SGPR posterior should be exactly the same as the exact GP posterior. In this case, 
by substituting $\K_{\zz,\zz}$ with $\K_{\xx,\xx}$, the mean and covariance matrix of $\tilde{p}(\mathbf{u})$ become:
\begin{align*}
    \m_\u &= \mathbf{R}_{\u\u}^{-1}(\mathbf{\mathbf{S}}^{- 1}\mathbf{y} + \K_{\xx,\xx}^{- 1}m(\xx)) = \K_{\xx,\xx}(\K_{\xx,\xx} + \mathbf{S})^{- 1}(\mathbf{y} - m(\xx)) + m(\xx) \\
    \mathbf{R}_{\u\u} &= (\K_{\xx,\xx}^{- 1} + \mathbf{\mathbf{S}}^{- 1})^{-1} = \K_{\xx,\xx} - \K_{\xx,\xx} (\K_{\xx,\xx} + \mathbf{S})^{-1}\K_{\xx,\xx},
\end{align*}
which matches the predictive distribution of the exact GP at $\xx$.

Derivations and numerical implementations for \SGPR in the homoskedastic and heteroskedatic cases, as stated in Section \ref{supp_sec:numerical-ELBO} and \ref{supp_sec:prediction}, are also considered and discussed in some other work \citep{GPflow2017, Maddox2021ConditioningSV} but differ slightly in the details. 

\subsection{Sparse Bayesian quadrature}
\label{supp_sec:sparse-bq}

To compute the variational posterior of \VSBQ, we need to compute integrals of Gaussian distributions against the \SGPR posterior with heteroskedastic noise. We provide here analytical formulae for this purpose. The following formulae are written in the zero-mean case; the non-zero mean case can be straightforwardly derived from this one (i.e., by considering $f - m$). We follow \citet{rasmussenBayesianMonteCarlo2002}.
Here, we denote with $\bm{\Sigma}_\ell$ the diagonal matrix of the parameters in the covariance kernel and with $\sigma_f$ the output scale, so that $\kappa(\x,\x') = \sigma_f^2 \Lambda \mathcal{N}(\x ; \x',\bm{\Sigma}_\ell)$, with 
$\Lambda \equiv \sqrt{(2\pi)^D \left|\bm{\Sigma}_\ell \right|}$.\footnote{This formulation of the squared exponential kernel is equivalent to the main paper, but makes it easier to apply Gaussian identities.}

We are interested in Bayesian quadrature formulae for the sparse GP integrated over Gaussian distributions of the form $\mathcal{N}(\cdot \, ; \vmu_j,\bm{\Sigma}_j)$, for $1 \le j \le K$. 
The integrals of interest are Gaussian random variables which depend on the sparse GP and take the form:
\begin{equation*}
\begin{split}
\mathcal{I}_j[\u] = & \int \mathcal{N}(\tilde{\x} ; \vmu_j,\bm{\Sigma}_j) f(\tilde{\x} \mid \textbf{u}) \mathrm{d} \tilde{\x}. 
\end{split}
\end{equation*}
Denoting with $\psi(\textbf{u})$  the optimal variational distribution of $\u$ in \SGPR, the posterior mean of each integral is:
\begin{equation*}
\begin{split}
\mathbb{E}\left[\mathcal{I}_j\right] = & \int \int \mathcal{N}(\tilde{\x} ; \vmu_j,\bm{\Sigma}_j) f(\tilde{\x} \mid \textbf{u}) \psi(\textbf{u}) \mathrm{d} \tilde{\x} \mathrm{d}\textbf{u} \\
= & \int \mathcal{N}(\tilde{\x} ; \vmu_j,\bm{\Sigma}_j) \mu_{\vgp}(\tilde{\x}) \mathrm{d} \tilde{\x}, \\
= & \int \mathcal{N}(\tilde{\x} ; \vmu_j,\bm{\Sigma}_j) \kappa(\tilde{\x},\zz)\bm{\Sigma} \K_{\zz,\xx}\mathbf{S}^{-1} \y \; \mathrm{d} \tilde{\x} \\
 =& \;\bm{w}_j^T \left[ \bm{\Sigma}\K_{\zz,\xx}\mathbf{S}^{-1} \right] \y,
\end{split}
\end{equation*}
where $(\bm{w}_j)_p \equiv \sigma_f^2 \Lambda \mathcal{N}(\bm{\mu}_j ; \z_p, \mathbf{\Sigma}_j + \bm{\Sigma}_\ell)$.
We compute then the posterior covariance between integrals $\mathcal{I}_j$ and $\mathcal{I}_k$,
\begin{align*}
\mathrm{Cov}(\mathcal{I}_j,\mathcal{I}_k) =& \int \int \mathcal{N}(\tilde{\x} ; \vmu_j, \bm{\Sigma}_j)\mathcal{N}(\tilde{\x}' ; \vmu_k, \bm{\Sigma}_k) \mathrm{Cov}(f(\tilde{\x}),f(\tilde{\x}')) \mathrm{d}\tilde{\x}\mathrm{d}\tilde{\x}' \\
=& \int \int \mathcal{N}(\tilde{\x} ; \vmu_j, \bm{\Sigma}_j)\mathcal{N}(\tilde{\x}' ; \vmu_k, \bm{\Sigma}_k) \kappa_{\vgp}(\tilde{\x},\tilde{\x}^\prime) \mathrm{d}\tilde{\x}\mathrm{d}\tilde{\x}' \\
=&  \int \int \sigma_f^2 \Lambda\mathcal{N}(\tilde{\x};\tilde{\x}',\bm{\Sigma}_\ell)) \mathcal{N}(\tilde{\x}';\vmu_j,\bm{\Sigma}_j) \mathcal{N}(\tilde{\x};\vmu_k,\bm{\Sigma}_k) \mathrm{d}\tilde{\x}\mathrm{d}\tilde{\x}' \\
& - \int\int  \mathcal{N}(\tilde{\x}';\vmu_j,\bm{\Sigma}_j) \mathcal{N}(\tilde{\x};\vmu_k,\bm{\Sigma}_k) \sigma_f^2 \Lambda \mathcal{N}(\tilde{\x};\zz,\bm{\Sigma}_\ell) \\
& \cdot \left[ \K_{\zz,\zz}^{-1} - \bm{\Sigma} \right] \sigma_f^2 \Lambda \mathcal{N}(\tilde{\x}';\zz,\bm{\Sigma}_\ell) \mathrm{d}\tilde{\x}\mathrm{d}\tilde{\x}' \\
=& \sigma_f^2 \Lambda \mathcal{N}(\vmu_j ; \vmu_k,\bm{\Sigma}_\ell + \bm{\Sigma}_j + \bm{\Sigma}_k) - \bm{w}_j^T \left[  \K_{\zz,\zz}^{-1} - \bm{\Sigma}\right] \bm{w}_k.
\end{align*}

\section{Proofs}
\label{supp_sec:proofs}
In this section, we provide proofs for the lemmas in the main paper. We note $p(\mathbf{f},\mathbf{u} \mid \vgp))$ the sparse GP posterior associated with $\mathbf{f}$ and $\mathbf{u}$, given the optimal variational parameters $\vgp$.

\begin{lemma}[Lemma 3.1]
\label{supp_lemma:dist_predict}
Assume that $D_\text{KL}(p(\mathbf{f},\mathbf{u} \mid \vgp) \Vert  p(\mathbf{f},\mathbf{u} \mid \mathbf{y}))  < \gamma$. 
%
Then, for any $\ell>0$ there exists $K_\ell$ such that, for any $\x^*$, $\vert \mathbb{E}[{f}(\x^*)^\ell] - \mathbb{E}[{f}_e(\x^*)^\ell] \vert < K_\ell \sqrt{\gamma/2}$. There also exists $K_\mathrm{e}$ such that, for any $\x^*$, $\vert \mathbb{E}[\mathrm{exp}({f}(\x^*))] - \mathbb{E}[\mathrm{exp}({f}_e(\x^*))] \vert < K_\mathrm{e} \sqrt{\gamma/2}$.
\end{lemma}

\begin{proof}[Proof of Lemma 3.1]

For this lemma, we only need to study the predictive distribution at a single point $\x^*$, with associated value $f^* \equiv f(\x^*)$.
\begin{align*}
   & D_\text{KL}(p(\textbf{f},\textbf{u},f^* \mid \vgp) \Vert p(\textbf{f},\textbf{u},f^* \mid \mathbf{y}))  \\
   =& \int\int\int p(\textbf{f},\textbf{u},f^* \mid \vgp) \log \left( \frac{p(\textbf{f},\textbf{u},f^* \mid \vgp)}{p(\textbf{f},\textbf{u},f^*\mid \y)} \right) \mathrm{d}\textbf{f}\mathrm{d}\textbf{u}\mathrm{d}f^* \\
   =& \int\int\int  p(f^*,\textbf{f} \mid \textbf{u})\tilde{p}(\textbf{u})\log\left( \frac{p(\textbf{f},\textbf{u},f^*\mid \vgp)}{p(\textbf{f},\textbf{u},f^*\mid \y)} \right)\mathrm{d}\textbf{f}\mathrm{d}\textbf{u}\mathrm{d}f^* \\
   =& \int\int\int p(f^*,\textbf{f} \mid \textbf{u})\tilde{p}(\textbf{u})\log\left( \frac{p(\mathbf{y})p(\textbf{f},f^* \mid \textbf{u})\tilde{p}(\textbf{u})}{p(\mathbf{y}\mid \textbf{f} )p(\textbf{f},f^*\mid \textbf{u})p(\textbf{u})} \right)\mathrm{d}\textbf{f}\mathrm{d}\textbf{u}\mathrm{d}f^* \\
   =& \int\int p(\textbf{f} \mid \textbf{u})\tilde{p}(\textbf{u})\log\left( \frac{p(\mathbf{y})p(\textbf{f} \mid \textbf{u})\tilde{p}(\textbf{u})}{p(\mathbf{y}\mid \textbf{f} )p(\textbf{f}\mid \textbf{u}) p(\textbf{u})} \right)\mathrm{d}\textbf{f}\mathrm{d}\textbf{u} \\
   =& D_\text{KL}(p(\textbf{f},\textbf{u}\mid \vgp) \Vert p(\textbf{f},\textbf{u} \mid \mathbf{y})).
\end{align*}
By Pinsker inequality \citep{tsybakov2003introduction}, we know that: \[ \Vert p(\textbf{f},\textbf{u},f^* \mid \mathbf{y}) - p(\textbf{f},\textbf{u},f^*\mid\vgp) \Vert_{TV} \leq \sqrt{ D_\text{KL}(p(\textbf{f},\textbf{u},f^* \mid \vgp) \Vert p(\textbf{f},\textbf{u},f^* \mid \mathbf{y}))/2} < \sqrt{\gamma/2}, \]
using one of the assumptions of the Lemma.
We further have that \begin{equation*}
    \Vert p(f^* \mid \mathbf{y}) - p(f^*\mid\vgp) \Vert_{TV} \leq \Vert p(\textbf{f},\textbf{u},f^* \mid \mathbf{y}) - p(\textbf{f},\textbf{u},f^*\mid\vgp) \Vert_{TV}, 
\end{equation*} as a coupling of the joint distribution is a coupling of the marginal distributions, and by definition, the total variation distance is $\Vert \pi_a - \pi_b \Vert_{TV} = \inf_{\omega \in \Omega(\pi_a,\pi_b)} P_\omega(X \neq Y)$, where $(X,Y) \sim \omega$ and $ \Omega(\pi_a,\pi_b)$ is the set of all the couplings between $\pi_a$ and $\pi_b$.

To find the inequalities of the Lemma, we will first show that:
\begin{equation} \label{supp_eq:lemma_power}
\left\vert \int f^{*\ell} (p(f^* \mid \mathbf{y}) - p(f^* \mid \vgp)) \mathrm{d}f^* \right\vert \leq K_\ell \sqrt{\gamma/2}.
\end{equation}
As both $p(\cdot \mid \vgp )$ and $p(\cdot \mid \mathbf{y})$ are normal, there exists $A\subset \mathbb{R}$ compact such that $\int_{\mathbb{R} \backslash A} f^\ell (p(f^* \mid \mathbf{y}) - p(f^* \mid \vgp)) \mathrm{d}f^* \leq K_\ell'' \sqrt{\gamma/2}$ for $K_\ell''>0$ small enough. We can then study on $A$ the integral, using the fact that the total variation distance is the $L_1$ norm and the fact that $x \mapsto x^\ell$ is continuous, and thus bounded by some $K_\ell'$ on $A$:
\begin{align*}
    \left\vert \int f^{*\ell} (p(f^*\mid \mathbf{y}) - p(f^* \mid \vgp)) \mathrm{d}f^* \right\vert &\leq \left\vert \int_{\mathbb{R} \backslash A} f^{*\ell} (p(f^* \mid \mathbf{y}) - p(f^* \mid \vgp)) \mathrm{d}f^* \right\vert \\
    & \hspace{1cm}+\left\vert \int_A f^{*\ell} (p(f^* \mid \mathbf{y}) - p(f^* \mid \vgp)) \mathrm{d}f^* \right\vert \\
    &\leq K_\ell'' \sqrt{\gamma/2}+ K'_\ell\int_A \vert p(f^* \mid \mathbf{y}) - p(f^* \mid \vgp)\vert \mathrm{d}f^* \\
    &\leq K_\ell'' \sqrt{\gamma/2}+ K'_\ell\int_\mathbb{R} \vert p(f^* \mid \mathbf{y}) - p(f^* \mid \vgp)\vert \mathrm{d}f^* \\
    &\leq K_\ell'' \sqrt{\gamma/2}+ K_\ell'\Vert p(f^* \mid \mathbf{y}) - p(f^*\mid\vgp) \Vert_{TV} \\
    &\leq K_\ell\sqrt{\gamma/2}
\end{align*}
for $K_\ell = K_\ell'' + K_\ell'$, proving the first inequality of the Lemma.

The second inequality of the Lemma follows from an identical proof as above, except that in Eq. \ref{supp_eq:lemma_power} we would use an exponential instead of the power function.\footnote{An alternative derivation for these results would use uniform integrability and conclude that the distance between the parameters of the two normal distributions is controlled. We preferred to provide here a longer but more explicit proof.}
\end{proof}
\begin{lemma}[Lemma 3.2]
 \label{supp_lemma:lemma_int_exp}
 Let $a$ and $b$ be two functions associated with two distributions defined on $\mathcal{X}$, $\pi_a \propto \exp(a(\cdot)) $ and $\pi_b \propto \exp(b(\cdot))$. If $\forall x, \ \vert a(x) - b(x) \vert < K$, then:
 \[ \Vert \pi_a - \pi_b \Vert_{TV} \leq  1-\exp(-K).\]
 \end{lemma}
 \begin{proof}[Proof of Lemma 3.2]
 
    By definition, $\Vert \pi_a - \pi_b \Vert_{TV} = \inf_{\omega \in \Omega(\pi_a,\pi_b)} P_\omega(X \neq Y)$, where $(X,Y) \sim \omega$ and $ \Omega(\pi_a,\pi_b)$ is the set of all the couplings between $\pi_a$ and $\pi_b$.
    
    To find an upper bound on the TV distance it is sufficient to find a particular coupling $\omega$ such that $P_\omega(X \neq Y)$ is small enough. Here, we propose the following coupling for joint sampling of $X$ and $Y$, derived from the rejection sampling algorithm. Note that $a\vee b$ is the maximum of $a$ and $b$ and $a\wedge b$ is the minimum:
    \begin{itemize} 
    \item Sample $Z \in \mathbb{R}^{d+1}$ under the curve $\exp(a\vee b)$, i.e., $Z[1:d] \sim \exp(a(\cdot)\vee b(\cdot))$ and $Z[d+1] \sim \mathcal{U}(0,\exp(a(Z[1:d])\vee b(Z[1:d])))$;
    
    \item If $Z[d+1]< \exp(a)$ then $X=Z[1:d]$, and if $Z[d+1]<\exp(b)$ then $Y=Z[1:d]$.
    
    \item Otherwise, we have either $Z[d+1]> \exp(a)$ and $Z[d+1]<\exp(b)$ or the opposite, i.e., $Z[d+1]> \exp(b)$ and $Z[d+1]<\exp(a)$. If $Z[d+1]>\exp(a)$ (resp. $b$), then resample $Z'$ under the curve $\exp(a\vee b)$ until $Z'[d+1]<\exp(a)$ (resp. $b$), then $X=Z'[1:d]$ (resp. $Y=Z'[1:d]$).
    
    \item Return $(X,Y)$.
    \end{itemize}
    
    Under this coupling $P_\gamma(X \neq Y) \leq P(Z[d+1]>\exp(b \wedge a))$. Which leads to the following bound, using that $a(x) \vee b(x) - a(x) \wedge b(x)\leq K$:
    \begin{align*}
        \Vert \pi_a - \pi_b \Vert_{TV} & \leq \frac{  \int \exp(a(x) \vee b(x)) - \exp(a(x) \wedge b(x)) \mathrm{d}x}{ \int \exp(a(x)\vee b(x)) \mathrm{d}x} \\
        &=  \frac{\int \exp(a(x)\vee b(x))(1 - \exp(a(x)\wedge b(x)-a(x)\vee b(x)))\mathrm{d}x}{\int \exp(a(x)\vee b(x))\mathrm{d}x} \\
        &\leq  \frac{\int \exp(a(x)\vee b(x))(1 - \exp(-K))\mathrm{d}x}{\int \exp(a(x)\vee b(x))\mathrm{d}x} \\
        &= 1- \exp(-K),
    \end{align*}
where we used that $a(x)\vee b(x) - a(x)\wedge b(x) = |a(x) - b(x)|$.
 \end{proof}

\section{Experiment details and additional results}

\subsection{Implementation details}
\label{supp_sec:implementation_details}
In this section, we describe the implementation details of variational sparse Bayesian quadrature (\VSBQ), including noise shaping, choice of hyperparameters, and variational inference details. Algorithm~\ref{alg:vsbq} summarizes the complete procedure of \VSBQ.

\begin{algorithm}[H]
\caption{Variational Sparse Bayesian Quadrature (VSBQ)}
\label{alg:vsbq}
\KwIn{Evaluation traces $\left( \mathbf{X}, \y, \s \right) = (\x_n, y_n, s_n)_{n=1}^N$ from MAP optimizations}
\KwOut{Posterior approximation $q_\phi$, estimated surrogate \ELBO mean $\overline{\ELBO}$ and its standard deviation $\ELBO_\text{sd}$}
\vspace{0.5em}
\textbf{Step 1: Trimming of evaluations}

Compute \textsc{lcb}$(\x_n) = y_n - \beta s_n$, \textsc{ucb}$(\x_n) = y_n + \beta s_n$ for all $n$\;
Set $\textsc{lcb}_\text{max} = \max_n(\textsc{lcb}(\x_n))$\;
Discard $\x_n$ where $\textsc{lcb}_\text{max} - \textsc{ucb}(\x_n) > \eta_\text{trim}$\;
Retain remaining evaluations as $\left( \mathbf{X}, \y, \s \right)$;

\vspace{0.5em}
\textbf{Step 2: Sparse GP fitting}

Initialize sparse GP hyperparameters by fitting an exact GP to a stratified $K$-means subset of $\left( \mathbf{X}, \y, \s \right)$\;
\Repeat{no improvement in GP-\ELBO}{
  Select $M$ inducing points $\zz$ via greedy variance selection (see Section 3.2 and \citep{burt2020jmlr})\;
  Update sparse GP hyperparameters via maximizing GP-\ELBO in Eq.~24\;
}
Compute the sparse GP posterior given observations $\left( \mathbf{X}, \y, \s \right)$ (see Eq.~18 and Eq.~19)\;
Obtain a sparse GP surrogate $f$ for the target log-joint density function $f_0$;

\vspace{0.5em}
\textbf{Step 3: Variational inference with sparse Bayesian quadrature}

Initialize variational posterior $q_\phi$ as a mixture of $K$ multivariate Gaussians in Eq.~20\;
\Repeat{convergence of $\overline{\ELBO}$ or max iterations reached}{
  Compute analytically the expected log joint $\mathbb{E}_{f}\left[\mathbb{E}_{\qparams} \left[f \right]\right]$ and its variance $\mathrm{Var}_{f}\left[\mathbb{E}_{\qparams} \left[f \right]\right]$, via sparse Bayesian quadrature (see Eq.~22 and Eq.~23) \;
  Maximize $\overline{\ELBO}$: $\mathbb{E}_{f}\left[\mathbb{E}_{\qparams} \left[f \right]\right] + \mathcal{H}[q_{\qparams}]$\ using reparameterized stochastic gradients with the Adam optimizer~\citep{kingma2014adam};\footnote{The entropy $\mathcal{H}[q_\phi]$ for Gaussian mixtures lacks a closed-form; we follow \citet{acerbi2018bayesian} for stochastic estimation.}
  
    Update $q_\phi$ parameters\;
}

\Return{$q_\phi$, $\overline{\ELBO}$, $\ELBO_\text{sd} = \sqrt{\mathrm{Var}_{f}\left[\mathbb{E}_{\qparams} \left[f \right]\right]}$}
\end{algorithm}

\paragraph{Design principles for the noise shaping formula.}

We recall that noise shaping increases the total likelihood variance for observation $(\x_n, y_n, \sigma_\text{obs}(\x_n))$,
\begin{align*}
\sigma^2_\text{tot}(\x_n, y_n) &= \sigma^2_\text{obs}(\x_n) + \sigma^2_\text{shape}(\Delta y_n),
\end{align*}
where $\sigma^2_\text{obs}(\x_n)$ is the estimated measurement variance at $\x_n$, and $\Delta y_n \equiv y_\text{max} - y_n$, with $y_\text{max}$ the maximum observed log-density.
We design $\sigma^2_\text{shape}(\Delta y)$ according to the following principles: 
\begin{enumerate}
\item[a.] 
Noise shaping should be a monotonically increasing function of $\Delta y$ (larger shaping noise for lower-density points);
\item[b.] 
Below a threshold $\theta_\sigma$, noise shaping should be `small', up to a quantity $\sigma_\text{med}$ (noise shaping should be small in high-density regions);
\item[c.] 
Asymptotically, the noise shaping standard deviation should increase \emph{linearly} in $\Delta y$, as any other functional form would make the noise shape contribution disappear (for sublinear functions) or dominate (superlinear) for extremely low values of the log-density.
\end{enumerate}
Following these principles, we propose the form used in the main text (Eq. 21),
\begin{equation*} \label{supp_eq:noiseshaping_appendix}
\sigma_\text{shape}(\Delta y) = \exp((1-\rho) \log \sigma_{\text{min}} + \rho \log \sigma_{\text{med}}) + \mathbf{1}_{\Delta y  \geq \theta_\sigma} {\lambda_\sigma} (\Delta y - \theta_\sigma),
\end{equation*}
where $\rho = \min(1,\Delta y/\theta_\sigma )$;  $\theta_\sigma$ is a threshold for `very low density' points, at which we start the linear increase; $\lambda_\sigma$ is the slope of the increase; and $\sigma^2_\text{min}$ and $\sigma^2_{\text{med}}$ are two shape parameters. $\sigma_{\text{med}}$ is the added noise at the low density threshold $\theta_\sigma$. 

\paragraph{Trimming and noise shaping hyperparameters.}

Both the trimming stage and noise shaping involve the selection of hyperparameters for what is considered a `low-density threshold'. The trimming stage consists of removing from the initial set points with log posterior density lower than the threshold $\eta_\mathrm{trim}$, relative to the maximum observed value. Noise shaping begins to linearly increase the added shaping noise starting from the low-density threshold $\theta_\sigma$. 

\begin{figure}[htb!]
    \centering
    \includegraphics{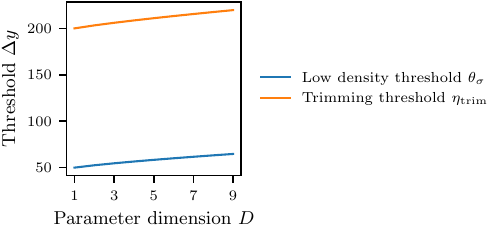}
    \caption{Log density threshold value versus the parameter dimension $D$.}
    \label{supp_fig:nstd_thresholds}
\end{figure}

As discussed by \citet{gammalFastRobustBayesian2022}, we can set a reasonable threshold in $D$ dimensions by considering a multivariate normal distribution in dimension $D$. The log density of a multivariate normal distribution is proportional to the sum of $D$ independent standard 1D Gaussian random variables. By defining $\Delta_y = 2\left[ \max(\log p) - \log p \right]$, we have $\Delta_y \sim \chi^2_D$. Further, the threshold for a ``$n$-$\sigma$ contour'' \citep{gammalFastRobustBayesian2022} is,
\begin{equation}
    [\Delta_y] (n) = F_{D}^{-1} \left[ \erf (n / \sqrt{2}) \right],
\end{equation}
where $F_{D}$ is the $\chi^2$ cumulative distribution function for $D$ degrees of freedom. In other words, we choose as a `low-density' threshold the density of a multivariate normal at $n$ standard deviations from the center, for $n \gg 1$.
In the experiments, we use $\eta_\mathrm{trim} = [\Delta_y] (20)$ and $\theta_\sigma = [\Delta_y] (10) $. The values of $\eta_\mathrm{trim}$ and $\theta_\sigma$ are plotted in Figure \ref{supp_fig:nstd_thresholds}.

As for the confidence interval parameter $\beta$, we used 1.96 (the 97.5th percentile point of a normal distribution). For the other noise shaping hyperparameters, we used $\sigma_{\mathrm{med}}=1$, $\lambda_\sigma=0.05$ throughout the experiments.

\paragraph{Stochastic variational inference.} The number of variational components $K$ in \VSBQ is 50, which is also the default maximum number of components in \textsc{vbmc} \citep{acerbi2018variational}. For initialization, we first perform K-means clustering on a subset of training points and initialize each component location $\mmu_k^{(i)}$ around the cluster centers, adding Gaussian noise with a standard deviation $10^{-6}$, for $1 \le k \le K$. The initial scale of the component $\sigma_k \bm{\lambda}^{(i)}$ is set to $10^{-3}$, for each dimension $i$, where $1 \le i \le D$. For the two moons bimodal problem, we set the number of clusters to 50, using the top $80\%$ high-density points for K-means clustering. For all other problems, the number of clusters is set to 1, with K-means applied to the top $1\%$ of high-density points -- effectively computing the mean of the selected points.

As described in the main text, after fitting the sparse GP surrogate, variational inference is conducted by optimizing the surrogate \ELBO with Bayesian quadrature. In addition, we impose a soft penalty loss during the \ELBO optimization for bounding the variational parameters (means and scales of the mixture components), as done in \citet{acerbi2020variational}, to help constrain the variational distribution in the local trust region of the surrogate. The lower and upper bounds are computed based on the training points $\xx$. For each dimension $i$, let $\xx^{(i)}_{\text{min}}$ and $\xx^{(i)}_{\text{max}}$ denote the minimum and maximum value of this dimension. The lower bounds and upper bounds for the $k^{\text{th}}$ component's mean $\mmu_k^{(i)}$ and log scale $\log \sigma_k \bm{\lambda}^{(i)}$ are provided in Table~\ref{supp_tab:vp_bound}.
\begin{table}[ht]
\begin{tabular}{clcccc}
Parameter & Description & Lower bound & Upper bound \\
\hline
$\mmu_k^{(i)}$ & mixture component mean & $\xx^{(i)}_{\text{min}}$ & $ \xx^{(i)}_{\text{max}}$ 
\\
$\log \sigma_k \bm{\lambda}^{(i)}$ & mixture component scale & $\log \left[10^{-6} \left( \xx^{(i)}_{\text{max}} - \xx^{(i)}_{\text{min}} \right)\right] $ & $\log \left( \xx^{(i)}_{\text{max}} - \xx^{(i)}_{\text{min}} \right) $  \\
\hline
\end{tabular}
  \centering
\caption{The soft bounds for variation posterior parameters.}
\label{supp_tab:vp_bound}
\end{table}

For both the mean and scale, the soft penalty loss can be written as,
\begin{equation}
    \mathbf{1}_{\theta_k^{(i)} \leq \text{LB}(\theta_k^{(i)}) \text{ or } \theta_k^{(i)} \geq \text{UB}(\theta_k^{(i)})} \cdot\frac{1}{2} \left[ \frac{\max \left( \theta_k^{(i)} - \text{LB}(\theta_k^{(i)}), \text{UB}(\theta_k^{(i)}) -  \theta_k^{(i)}\right)}{\tau \left(\text{UB}(\theta_k^{(i)}) - \text{LB}(\theta_k^{(i)}) \right)} \right]^2,
\end{equation}
where $ \theta_k^{(i)} $ represents either the mean $ \mmu_k^{(i)} $ or the log scale $ \log \sigma_k \bm{\lambda}^{(i)} $, and $ \tau = 0.01 $.  $\text{LB}(\theta_k^{(i)})$ and $\text{UB}(\theta_k^{(i)})$ denote the lower and upper bounds, respectively.

\subsection{Further details of procedure and metrics}
\label{supp_sec:details_of_procedure}
\paragraph{MAP estimation.}
To find the global mode, we launch multiple MAP optimization runs in parallel and independently, using different random seeds and initial starting points. The initialization strategy proceeds as follows: we randomly sample a small batch of candidate points (e.g., $20D$) from the prior distributions and plausible parameter ranges, where the latter are guided by prior distributions or domain expertise. We then evaluate the log-density at each of these points and select the one with the highest value as the initial starting point for optimization. Each MAP optimization run produces a trace of evaluated points. To meet the total evaluation budget of 3000$D$ for the benchmark experiments, we sequentially add the optimization traces until the total number of evaluations exceeds the budget. The last trace is then truncated to ensure that the total number of evaluations precisely equals the budget. The entire MAP estimation procedure was repeated ten times for each problem, with different seeds, to yield ten different training datasets over which we computed statistics (see below).

\paragraph{Computing the metrics.} 

For $\Delta$LML, the true marginal likelihood is computed analytically, via numerical quadrature methods, or estimated from extensive \MCMC sampling via Geyer’s reverse logistic regression \citep{geyer1994estimating}, depending on the structure of each specific problem. The estimated log marginal likelihood of \VSBQ and \NNR are taken as the \ELBO computed in variational inference. For the Laplace method, the log normalization constant of the approximation can be computed analytically given the (numerically estimated) Hessian at the mode.
We computed the posterior metrics (\MMTV and \gskl) based on samples from the variational posteriors of \VSBQ and \NNR, samples from the Laplace approximation. The ground-truth posterior is represented by samples from well-tuned and extensive \MCMC sampling.

\paragraph*{Statistical analyses.} 
We ran the \VSBQ and \NNR algorithm with ten different random seeds, also corresponding to ten different training datasets, and computed the triplet of metrics ($\Delta$LML, \MMTV, \gskl) for each run. We report the median and 95\% confidence interval of the median obtained via bootstrap ($n_{\text{bootstrap}}=10^4$). For the Laplace method, we report the estimate obtained by running numerical differentiation of the Hessian from the MAP estimate (the mode). The output of the Laplace approximation is deterministic given the global mode, so there is a single estimate per problem.

\subsection{Black-box variational inference}
\label{supp_sec:bbvi}
For black-box variational inference, the score function estimator for the \ELBO gradient can be written as:
\begin{equation}
    \begin{split}
    \nabla_{\qparams} \mathbb{E}_{\qparams} \left[\log \frac{\like p(\x)}{q_{\qparams}(\x)} \right] &= \mathbb{E}_{\qparams} \left[ \nabla_{\qparams} \log q_{\qparams}(\x) \left( \log \like p(\x) - \log q_{\qparams}(\x) \right) \right] \\
    &= \mathbb{E}_{\qparams} \left[ \nabla_{\qparams} \log q_{\qparams}(\x) h_{\qparams}(\x) \right],
\end{split}
\end{equation}
where $h_{\qparams}(\x) = \log \like p(\x) - \log q_{\qparams}(\x)$. As shown by~\citet{ranganath2014black}, there are two possible ways to reduce the variance of the score function estimator: the Rao-Blackwellization technique and control variates. The Rao-Blackwellization technique cannot be used since we only have access to the target density while the dependency structure of the black-box target model is assumed unknown. Conversely, we can use control variates. In particular, noting that for any constant $b$, $\mathbb{E}_{\qparams}[ b \nabla_{\qparams} \log q_{\qparams} ] = 0$, $b\nabla_{\qparams} \log q_{\qparams}$ can be used as the control variate for reducing the stochastic gradient variance, i.e.,
\begin{equation}
    \mathbb{E}_{\qparams} \left[ \nabla_{\qparams} \log q_{\qparams}(\x) h_{\qparams}(\x) \right] = \mathbb{E}_{\qparams} \left[ \nabla_{\qparams} \log q_{\qparams}(\x) (h_{\qparams}(\x) - b) \right].
\end{equation}
An optimal choice of $b$ would require estimating the covariance between $\nabla_{\qparams} \log q_{\qparams}(\x) h_{\qparams}(x)$ and $\nabla_{\qparams} \log q_{\qparams}$, and the variance of $\nabla_{\qparams} \log q_{\qparams}$, which would require extra evaluations on the target density function. To simplify the implementation and comparison to other methods, we instead take an exponential moving average of $h_{\qparams}(\x)$ as the value of $b$, as suggested in the probabilistic programming framework Pyro~\citep{binghamPyroDeepUniversal2018}. The smoothing factor for the exponential moving average is 0.9.

As stated in the main text, we allocate $10 \times 3000D$ target density evaluation budget for BBVI. Denoting the number of Monte Carlo samples for gradient estimation with $M_g \in \{1, 10, 100\}$, the total number of optimization iterations is then set to $\frac{10 \times 3000D}{M_g}$. For all variational distribution families---a Gaussian with \emph{diagonal} covariance matrix, a Gaussian with \emph{full-rank} covariance matrix, and a mixture of Gaussians with $K=5$ and $K=50$ components---the Gaussian distribution mean is initialized near the origin, by adding Gaussian noise with a standard deviation $10^{-6}$ and the scales are set to $10^{-3}$.
 
\subsection{Neural network regression}
\label{supp_sec:nnr}
For neural network regression, we followed as closely as possible the same procedure as in \VSBQ, while substituting the sparse GP surrogate with a deep neural network. This means that we used techniques such as a global mean function, noise shaping, and trimming (removal of very low-density points), exactly as done in \VSBQ.
Empirically, we found that noise shaping helps stabilize the training of the neural network by avoiding exploding gradients.

We report below the neural network setup as described in the main text, with additional implementation and training details.

\paragraph{Neural network details.} For the neural network, we use a multilayer perceptron (MLP) with an input layer of dimension $D$, four hidden layers of 1024 units, an output layer for scalar prediction, and ReLu activation functions. The dataset is randomly split into training and validation sets, with a ratio of $4:1$. In addition, we add a negative quadratic mean function to the neural network output to ensure integrability, the same as the (sparse) GP. Thus, the surrogate function $g$ is:
\begin{equation*}
    g(\x; \mathbf{w}) = m_0 - \frac{1}{2} \sum_{i=1}^{\nparams} \frac{\left(x_i - \mu_i\right)^2}{\omega_i^2} + \text{MLP}(\x),
\end{equation*}
where $\mathbf{w}$ denotes the free parameters to optimize, including the parameters in the negative quadratic mean function and MLP parameters.

\paragraph{Loss.} The loss is the average log-likelihood under the heteroskedastic noise model, which is equivalent to the mean squared error (MSE) weighted by the noise variance,
\begin{equation*}
\label{supp_eq:nnr_objective}
    \mathcal{L}(\mathbf{w}) = \frac{1}{N} \sum_{n=1}^{N} 
     \frac{\left(g(\x_n) - y_n \right)^2}{\sigma_\text{tot}^2(\x_n, y_n)},
\end{equation*}
where $\sigma_n^2 = \sigma_{\text{obs}}^2(\x_n) + \sigma_{\text{shape}}^2(\Delta y_n)$.

\paragraph{Optimization.} We use AdamW~\citep{loshchilovDecoupledWeightDecay2019} to optimize the parameters, with a learning rate of 0.001 and batch size of 32. The optimization is stopped when the loss on the validation set does not decrease for 20 epochs.

\paragraph{Weight decay.} We optimize the neural network by trying three different values of weight decay hyperparameter $\alpha \in \{0, 0.01, 0.1\}$. The neural network with the lowest validation loss is used. Note that for adaptive gradient algorithms like AdamW, weight decay is different from $L_2$ regularization. Using weight decay for regularization is standard practice in modern neural network training~\citep{loshchilovDecoupledWeightDecay2019,zhangThreeMechanismsWeight2019}. As per standard practice, we apply weight decay only to the MLP weights, excluding the biases and the trainable quadratic mean parameters.

\paragraph{Stochastic variational inference.} After neural network fitting, we run stochastic variational inference via automatic differentiation (\ADVI; \citealp{kucukelbirAutomaticDifferentiationVariational2017}) to compute a tractable posterior approximation from the neural network surrogate. The variational distribution is the same as that used for \VSBQ, i.e., a mixture of $K=50$ multivariate normal distributions. Unlike \VSBQ where we can compute the expected log joint in the \ELBO analytically, with a neural network surrogate, we have to estimate the value and gradient of the expected log joint. We use reparametrization tricks~\citep{kingma2013auto} to get an unbiased estimate for the gradient of the expected log joint value. Apart from the difference in computing the expected log joint, all the other steps (optimization iterations, variational distribution initialization, etc.) stay the same as the variational inference part in \VSBQ.

\subsection{MAP estimates via Bayesian Adaptive Direct Search (BADS)}
\label{supp_sec:bads}
A popular choice for black-box optimization is Bayesian optimization (BO; \citealp{garnett2023bayesian}). BO can also deal with noisy observations like \cmaes and is known for its efficiency in finding the optimum. We therefore applied \VSBQ to optimization traces obtained from a (hybrid) BO optimization method named Bayesian Adaptive Direct Search (\bads;  \citealp{acerbi2017practical}), a state-of-the-art BO optimization algorithm with wide application in computational neuroscience and other fields.

\begin{figure}[tb]
    \centering
    \includegraphics[scale=1]{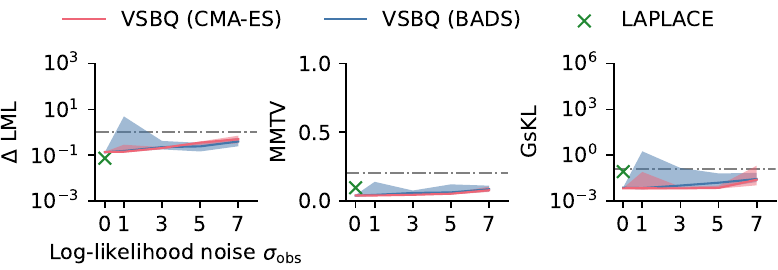}
    \caption{\textbf{Performance on Bayesian timing model with traces from different optimization algorithms.} Median $\Delta$LML loss (left), \MMTV (middle), and \gskl (right) as a function of the log-likelihood noise $\sigma_{\text{obs}}$ for the Bayesian Timing model. Shaded areas are 95\% CI of the median and grey dash-dotted horizontal lines are the rule-of-thumb thresholds for good performance ($\Delta$LML=1, \MMTV=0.2, \gskl=$1/8$). The performance of \VSBQ is virtually identical, regardless of the source of evaluations, whether \cmaes or \bads.}
    \label{supp_fig:timing_pybads}
\end{figure}
\begin{figure}[thb]
    \centering
    \includegraphics[scale=1]{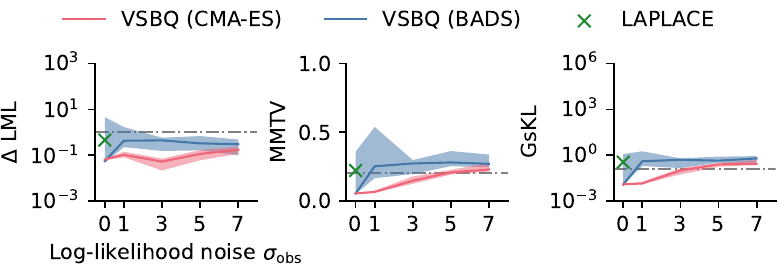}
    
    \caption{\textbf{Performance on multisensory causal inference model with traces from different optimization algorithms.} Median $\Delta$LML loss (left), \MMTV (middle), and \gskl (right) as a function of the log-likelihood noise $\sigma_{\text{obs}}$ for the Bayesian Timing model. Shaded areas are 95\% CI of the median and grey dash-dotted horizontal lines are the rule-of-thumb thresholds for good performance ($\Delta$LML=1, \MMTV=0.2, \gskl=$1/8$). \VSBQ with traces from \bads performs worse than \VSBQ with \cmaes, but still comparable to the (noiseless) \LA.}
    \label{supp_fig:multisensory_pybads}
\end{figure}

We provide the results for \VSBQ with \cmaes and \bads in Table \ref{supp_tab:twomoons_pybads}, \ref{supp_tab:banana_pybads} and Figure \ref{supp_fig:timing_pybads}, \ref{supp_fig:multisensory_pybads}. From the tables and figures, we can see that the performance of \VSBQ with \bads and \cmaes is almost identical for three out of four benchmark problems (Two Moons, multivariate Rosenbrock-Gaussian, and Bayesian timing model). Instead, we found that \VSBQ performs less effectively with traces from \bads in the multisensory causal inference model, compared to \cmaes. Still, \VSBQ (\bads) is comparable to the noiseless Laplace approximation (\LA) even in the presence of large amounts of log-density evaluation noise.

We hypothesize that \bads can be a worse choice than \cmaes for our purpose of post-process inference, exactly for the reasons that make \bads a better optimization algorithm on these problems \citep{acerbi2017practical}. Specifically, an optimization algorithm that converges to the global optimum quickly and efficiently is not ideal for post-process inference, in that a more exploratory population-based algorithm, such as \cmaes, provides better coverage and more information about the shape of the posterior landscape. For this reason, we recommend \cmaes over \bads for the purpose of post-process inference. Future work could explore strategies to augment the initial set of evaluations to improve coverage and enhance approximation quality.

\begin{table}[t]
    \centering
    \caption{\textbf{Two Moons posterior ($D=2$).} The method performance is measured using the metrics $\Delta$LML, \MMTV, and \gskl. For all metrics, lower values indicate better performance. We bold the best results based on the 95\% confidence interval (CI) of the median. If there are overlaps between CIs, we bold all overlapping values.\label{supp_tab:twomoons_pybads}}
    \begin{tabular}{c c c c c}
        \toprule
        & \textbf{$\Delta$LML} ($\downarrow$) & \textbf{MMTV} ($\downarrow$) & \textbf{GsKL} ($\downarrow$) \\
    \toprule
\VSBQ (\cmaes) & \textbf{0.0017} $\scriptstyle{[0.00057, 0.0026]}$ & \textbf{0.020} $\scriptstyle{[0.018, 0.021]}$ & \textbf{8.5e-05} $\scriptstyle{[4.4e-05, 0.00019]}$\\ 
\VSBQ (\bads) & \textbf{0.0010} $\scriptstyle{[0.00041, 0.0017]}$ & \textbf{0.020} $\scriptstyle{[0.019, 0.022]}$ & \textbf{0.00018} $\scriptstyle{[0.00013, 0.00020]}$\\ 
    \toprule
    \end{tabular}
\end{table}

\begin{table}[t]
    \centering
    \caption{\textbf{Multivariate Rosenbrock-Gaussian ($D=6$).} See Table~\ref{supp_tab:twomoons_pybads} for a detailed description of metrics and bolding criteria.\label{supp_tab:banana_pybads}}
    \begin{tabular}{c c c c c}
        \toprule
        & \textbf{$\Delta$LML} ($\downarrow$) & \textbf{MMTV} ($\downarrow$) & \textbf{GsKL} ($\downarrow$) \\
    \toprule
\VSBQ (\cmaes) & \textbf{0.20} $\scriptstyle{[0.20, 0.20]}$ & \textbf{0.037} $\scriptstyle{[0.035, 0.038]}$ & \textbf{0.018} $\scriptstyle{[0.017, 0.018]}$ \\ 
\VSBQ (\bads) & \textbf{0.19} $\scriptstyle{[0.19, 0.20]}$ & \textbf{0.038} $\scriptstyle{[0.037, 0.039]}$ & \textbf{0.018} $\scriptstyle{[0.017, 0.018]}$\\  
    \toprule
    \end{tabular}
\end{table}

\begin{figure*}
    \centering
    \subcaptionbox{Bayesian timing model ($\sigma_{\text{obs}}=3$).}{
        \includegraphics[scale=1]{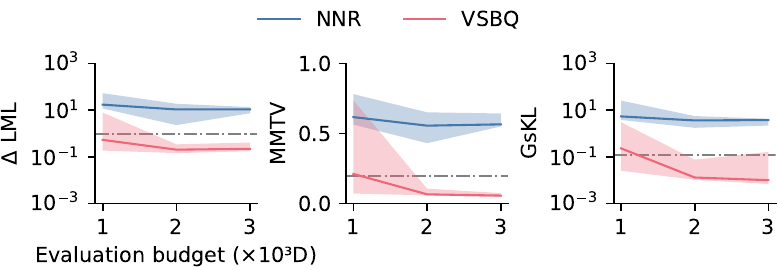}
    } \\
    \subcaptionbox{Multisensory causal inference model ($\sigma_{\text{obs}}=0$).}{
        \includegraphics[scale=1]{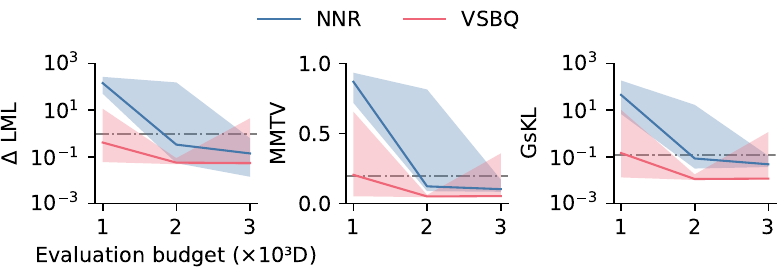}
    } \\
    \caption{
    \textbf{Sensitivity to the number of target evaluations, with BADS optimizer.} Median $\Delta$LML loss (left), \MMTV (middle), and \gskl (right) as a function of the number of target evaluations, for two benchmark problems. Shaded areas are 95\% CI of the median and grey dash-dotted horizontal lines are the rule-of-thumb thresholds for good performance ($\Delta$LML $=1$, \MMTV $=0.2$, \gskl $=1/8$). Compared to \NNR, \VSBQ achieves consistently better performance across most settings.
    }
    \label{supp_fig:varying_N_evaluations_bads}
\end{figure*}

Furthermore, as in the main text, we study the sensitivity to the number of target evaluations when using the \bads optimizer. As shown in Figure~\ref{supp_fig:varying_N_evaluations_bads}, in contrast to \cmaes, \VSBQ is more sensitive to the number of evaluations under \bads, likely due to poorer coverage of the posterior. Nonetheless, \VSBQ consistently outperforms \NNR across most tested settings.

\clearpage
\subsection{Visualization of posteriors}
\label{supp_sec:visualization}
We visualize posterior distributions as `corner plots', i.e., a plot with 1D and all pairwise 2D marginals. For visualization of individual posteriors obtained by the algorithms, for all problems, we report example solutions obtained from a run with the same random seed (see Figure \ref{supp_fig:twomoons_corner}, \ref{supp_fig:multi_banana_D_6_N_2_corner}, \ref{supp_fig:noisy_timing_st_std-3_corner}, and \ref{supp_fig:noisy_multisensory_6D_std-3_corner}).

\begin{figure}[!hb]
    \centering
    \subcaptionbox{\VSBQ \label{supp_fig:bimodal_svbmc}}{\includegraphics[scale=1]{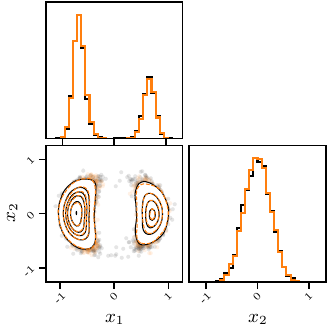}}
    \subcaptionbox{\NNR \label{supp_fig:bimodal_nnr}}{\includegraphics[scale=1]{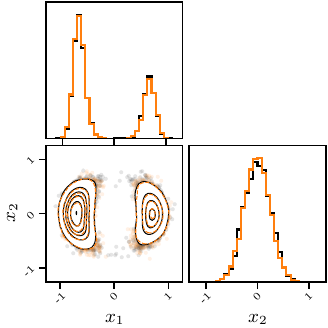}}
    \subcaptionbox{\LA \label{supp_fig:bimodal_la}}{\includegraphics[scale=1]{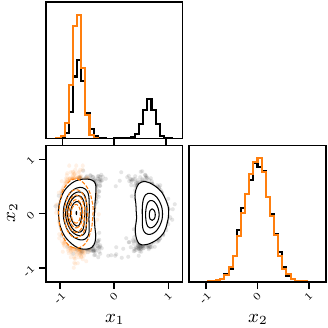}}
    \subcaptionbox{\BBVI, Gaussian (diagonal)}{\includegraphics[scale=1]{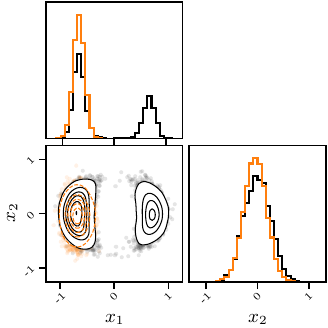}}
\end{figure}

\begin{figure}
    \centering    
    \subcaptionbox{\BBVI, Gaussian (full-rank)}{\includegraphics[scale=1]{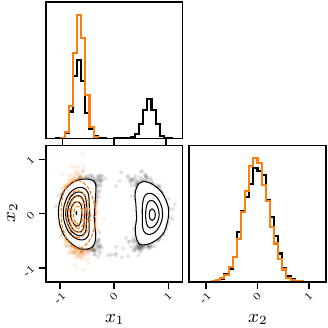}}
    \subcaptionbox{\BBVI, MoG ($K=5$)}{\includegraphics[scale=1]{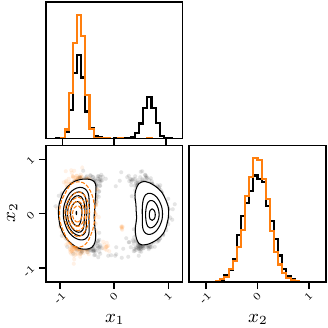}}

    \subcaptionbox{\BBVI, MoG ($K=50$)}{\includegraphics[scale=1]{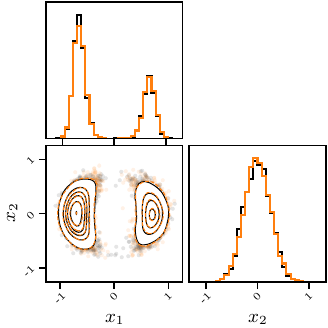}}
    \caption{\textbf{Two Moons bimodal posterior visualization.} The orange density contours and points in the sub-figures represent the posterior samples from \VSBQ, \NNR, and \LA, while the black contours and points denote ground truth samples. \VSBQ and \NNR perfectly capture the ground-truth bimodal posterior, while \LA is limited to one mode. Among the \BBVI methods, the MoG (K=50) configuration performs best.}
    \label{supp_fig:twomoons_corner}
\end{figure}

\begin{figure}
    \centering
    \subcaptionbox{\VSBQ \label{supp_fig:multi_banana_D_6_N_2_svbmc}}{\includegraphics[scale=1]{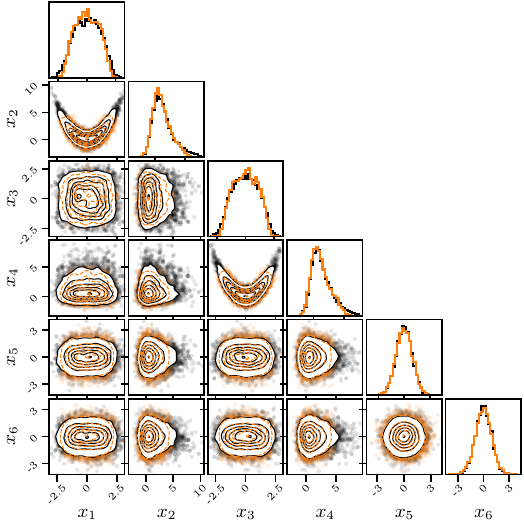}}
    \subcaptionbox{\NNR \label{supp_fig:multi_banana_D_6_N_2nnr}}{\includegraphics[scale=1]{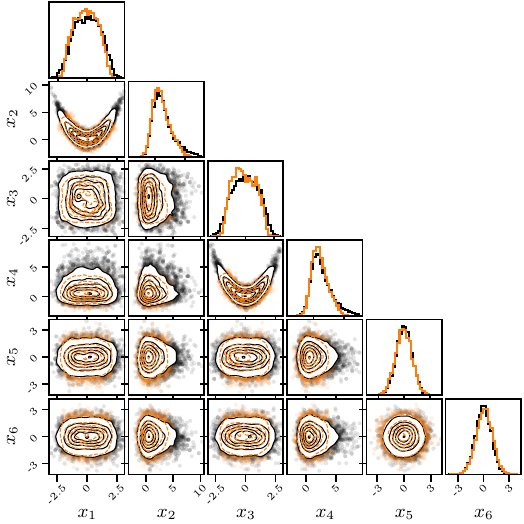}}
\end{figure}

\begin{figure}\ContinuedFloat
    \centering
    \subcaptionbox{\LA \label{supp_fig:multi_banana_D_6_N_2la}}{\includegraphics[scale=1]{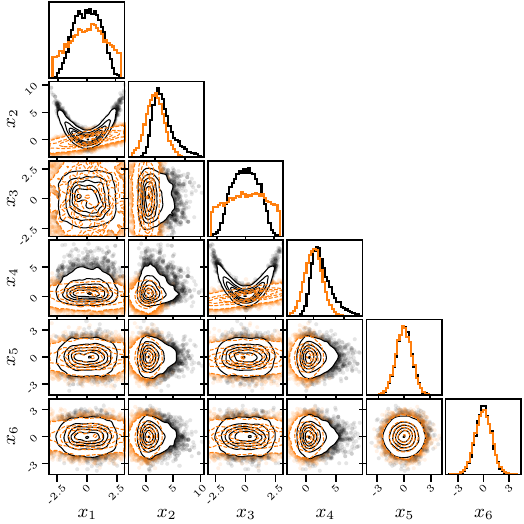}}
    \subcaptionbox{\BBVI, Gaussian (diagonal)}{\includegraphics[scale=1]{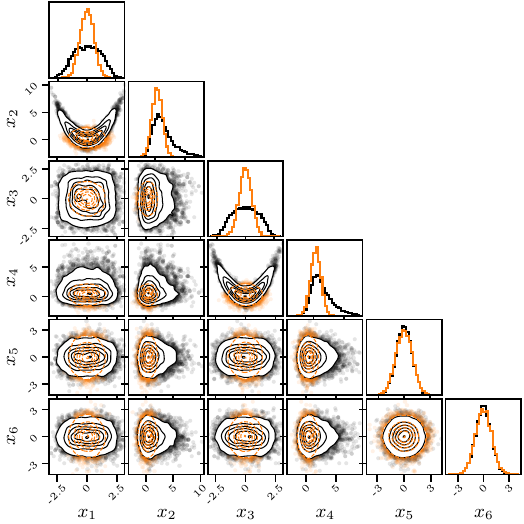}}
\end{figure}

\begin{figure}\ContinuedFloat
    \centering
    \subcaptionbox{\BBVI, Gaussian (full-rank)}{\includegraphics[scale=1]{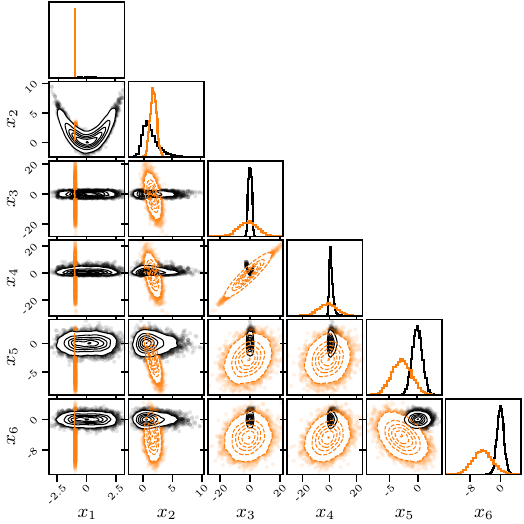}}
    \subcaptionbox{\BBVI, MoG ($K=5$)}{\includegraphics[scale=1]{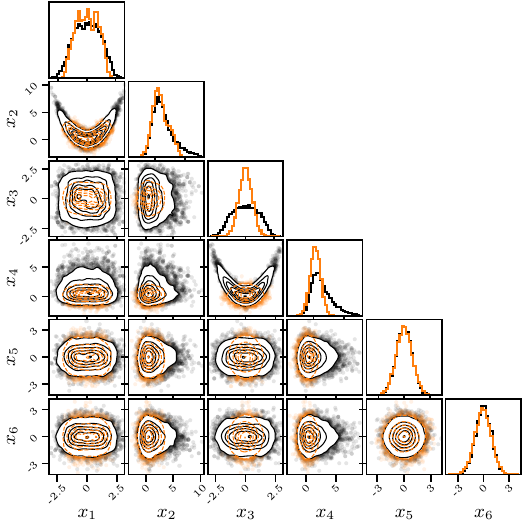}}
\end{figure}

\begin{figure}\ContinuedFloat
    \centering
    \subcaptionbox{\BBVI, MoG ($K=50$)}{\includegraphics[scale=1]{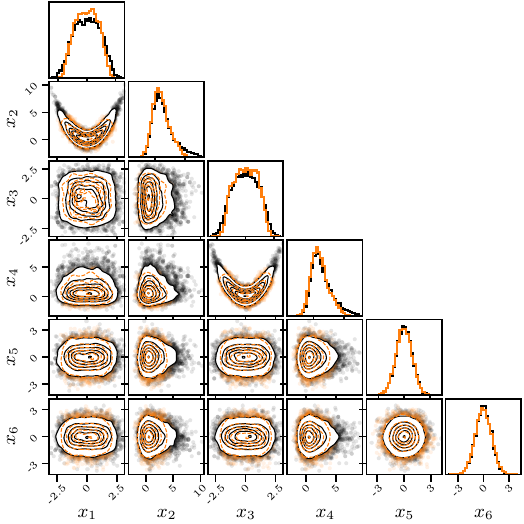}}
    \caption{\textbf{Multivariate Rosenbrock-Gaussian posterior visualization.} The orange density contours and points in the sub-figures represent the posterior samples from \VSBQ, \NNR, and \LA, while the black contours and points denote ground truth samples. Both \VSBQ and \NNR capture very well the complex shape of the distribution, while in this example \LA fails. Among the \BBVI methods, the MoG (K=50) configuration performs best.}
    \label{supp_fig:multi_banana_D_6_N_2_corner}
\end{figure}

\begin{figure}
    \centering
    \subcaptionbox{\VSBQ \label{supp_fig:noisy_timing_st_std-3_svbmc}}{\includegraphics[scale=1]{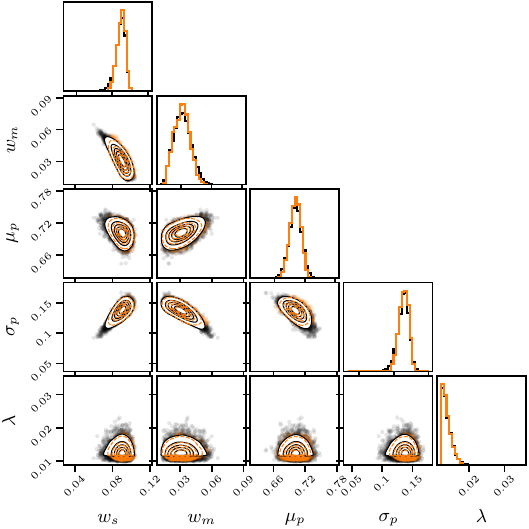}}
    \subcaptionbox{\NNR \label{supp_fig:noisy_timing_st_std-3nnr}}{\includegraphics[scale=1]{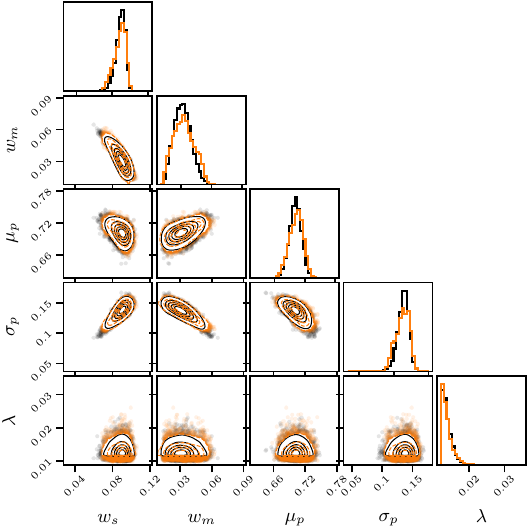}}
\end{figure}

\begin{figure}\ContinuedFloat
    \centering
    \subcaptionbox{\LA \label{supp_fig:noisy_timing_st_std-3la}}{\includegraphics[scale=1]{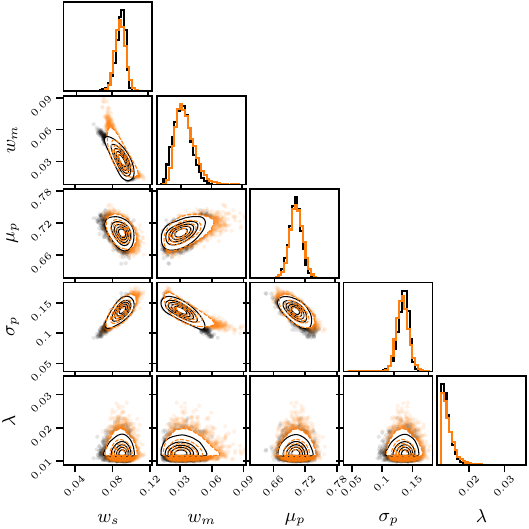}}
    \subcaptionbox{\BBVI, Gaussian (diagonal)}{\includegraphics[scale=1]{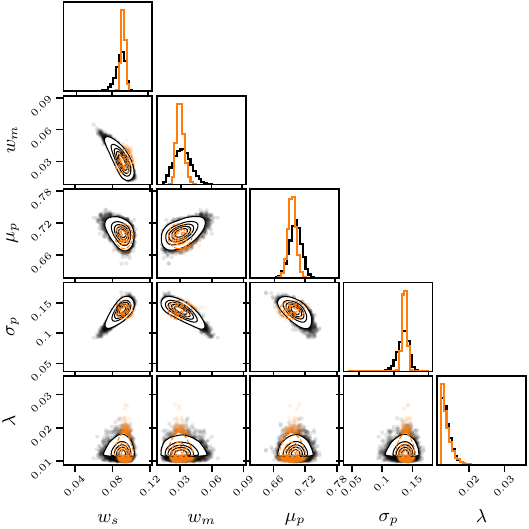}}
\end{figure}

\begin{figure}\ContinuedFloat
    \centering
    \subcaptionbox{\BBVI, Gaussian (full-rank)}{\includegraphics[scale=1]{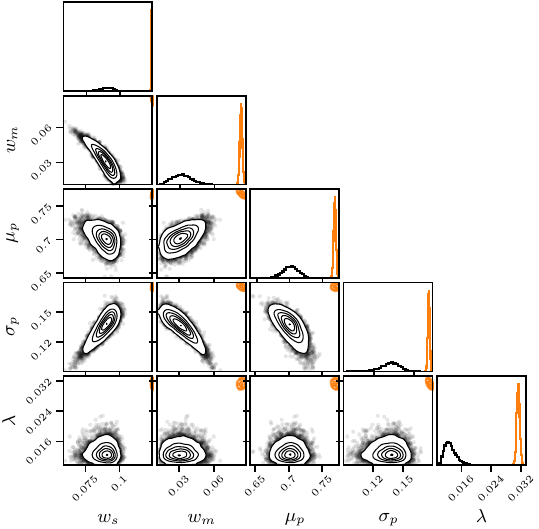}}
    \subcaptionbox{\BBVI, MoG ($K=5$)}{\includegraphics[scale=1]{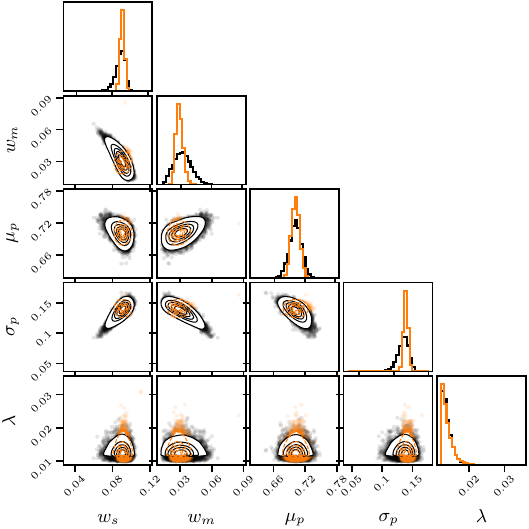}}
\end{figure}

\begin{figure}\ContinuedFloat
    \centering
    \subcaptionbox{\BBVI, MoG ($K=50$)}{\includegraphics[scale=1]{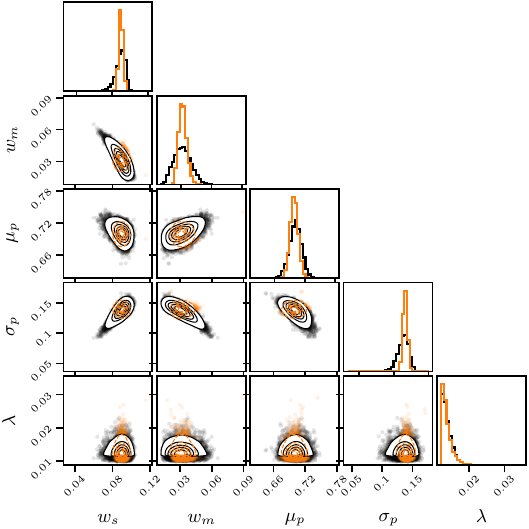}}
    \caption{\textbf{Bayesian timing model posterior visualization.} $\sigma_\textrm{obs} = 3$ for \VSBQ and \NNR, $\sigma_\textrm{obs} = 0$ for \LA. The orange density contours and points in the sub-figures represent the posterior samples from \VSBQ, \NNR, and \LA, while the black contours and points denote ground truth samples. Both \VSBQ and \NNR capture the shape of the posterior in the presence of observation noise. \LA obtains a reasonable approximation, for the noiseless case. Among the \BBVI methods, the Gaussian (diagonal), MoG (K=5), and MoG (K=50) configurations produce visually similar results and underperform compared to \VSBQ and \NNR.}
    \label{supp_fig:noisy_timing_st_std-3_corner}
\end{figure}

\begin{figure}
    \centering
    \subcaptionbox{\VSBQ \label{supp_fig:noisy_multisensory_6D_std-3_svbmc}}{\includegraphics[scale=1]{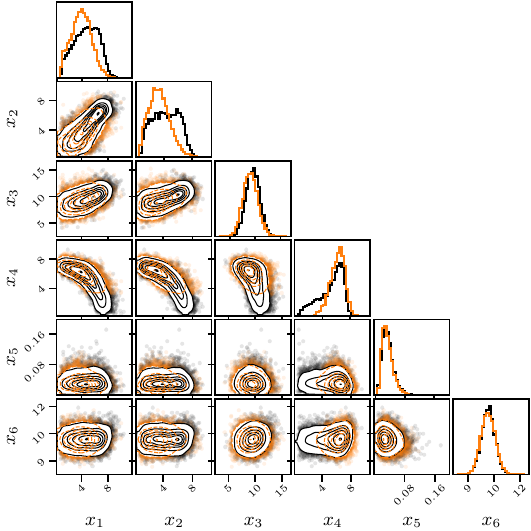}}
    \subcaptionbox{\NNR \label{supp_fig:noisy_multisensory_6D_std-3nnr}}{\includegraphics[scale=1]{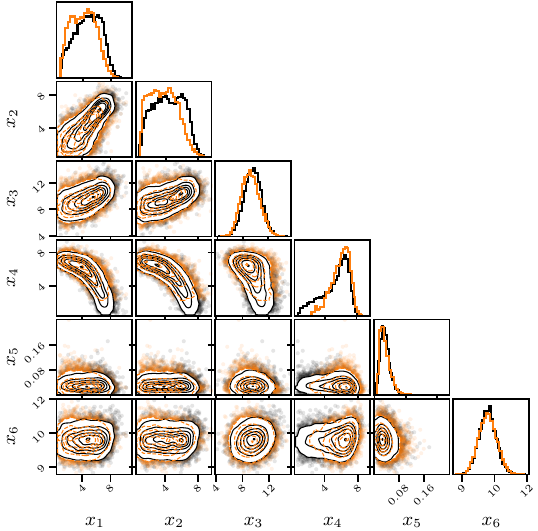}}
\end{figure}

\begin{figure}\ContinuedFloat
    \centering
    \subcaptionbox{\LA \label{supp_fig:noisy_multisensory_6D_std-3la}}{\includegraphics[scale=1]{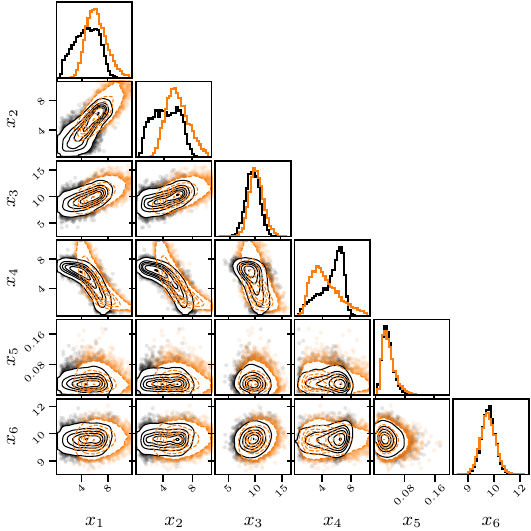}}
    \subcaptionbox{\BBVI, Gaussian (diagonal)}{\includegraphics[scale=1]{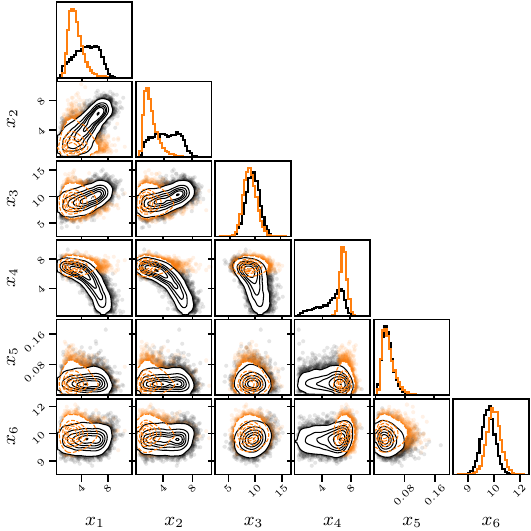}}
\end{figure}

\begin{figure}\ContinuedFloat
    \centering
    \subcaptionbox{\BBVI, Gaussian (full-rank)}{\includegraphics[scale=1]{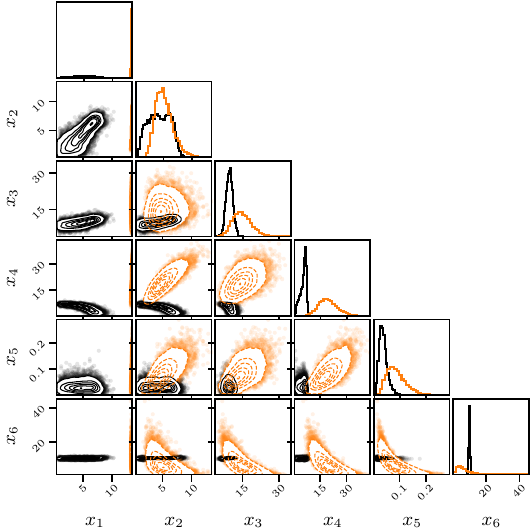}}
    \subcaptionbox{\BBVI, MoG ($K=5$)}{\includegraphics[scale=1]{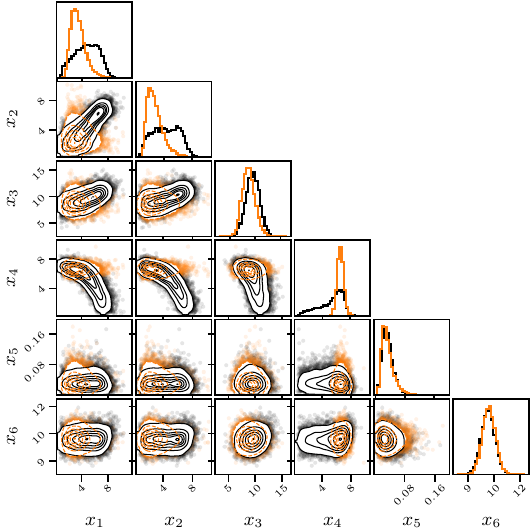}}
\end{figure}
\begin{figure}\ContinuedFloat
    \centering
    \subcaptionbox{\BBVI, MoG ($K=50$)}{\includegraphics[scale=1]{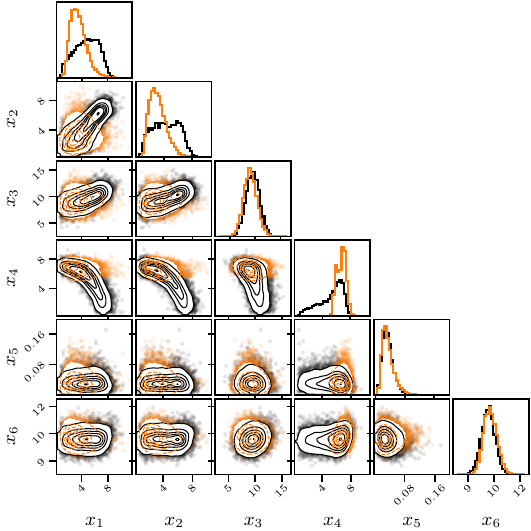}}
    \caption{\textbf{Multisensory causal inference model posterior visualization.} $\sigma_\textrm{obs} = 3$ for \VSBQ and \NNR, $\sigma_\textrm{obs} = 0$ for \LA. The orange density contours and points in the sub-figures represent the posterior samples from \VSBQ, \NNR, and \LA, while the black contours and points denote ground truth samples. $(x_1, x_2, x_3, x_4, x_5, x_6) = (\sigma_\text{vis}(c_\text{low}),\sigma_\text{vis}(c_\text{med}),\sigma_\text{vis}(c_\text{high}),\sigma_\text{vest},\lambda, \kappa)$. Both \VSBQ and \NNR obtain a reasonable approximation of the complex posterior under noisy evaluations, with \NNR yielding a more faithful approximation for this random seed. \LA fails to capture the posterior well, despite it being noiseless. Among the \BBVI methods, the Gaussian (diagonal), MoG (K=5), and MoG (K=50) configurations produce visually similar results and underperform compared to \VSBQ and \NNR.}
    \label{supp_fig:noisy_multisensory_6D_std-3_corner}
\end{figure}

\clearpage

\subsection{Runtime analysis}
\label{supp_sec:runtime}
In this section, we present an analysis of the runtime for our method and baselines. Such an analysis is important for a post-process inference method, which ideally should only take a relatively short time (e.g., a few minutes).\footnote{\BBVI methods are not considered post-processing inference techniques and serve only as a reference baseline for posterior approximation accuracy. Their runtime primarily depends on the total number of target function evaluations and the grid search for tuning learning rates.} For \VSBQ and \NNR, we report the wall-clock runtime based on 5 independent runs with different training data sets (i.e., traces from the MAP optimizations). For \LA, the runtime is based on 5 runs for numerically estimating the Hessian matrix. We measure the runtime on both CPU and GPU platforms.\footnote {It is worth noting that, in the case of GPU runtime measurement, only the sparse GP and neural network fitting were carried out with a GPU. Due to our current implementation, the stochastic variational inference optimization part for \VSBQ and \NNR was always conducted using the CPU. Future efforts to port the stochastic variational inference procedure completely to GPU could further enhance the computational performance of both methods.} Specifically, we utilize an AMD EPYC 7452 32-core Processor for CPU computations and an NVIDIA A100 for GPU computations.
 
The dominating algorithm overhead for \VSBQ is the sparse GP fitting, whose efficiency depends on the number of data points and inducing points. 
In general, we recommend utilizing as many inducing points as resources permit to achieve improved approximation accuracy. The variational inference part of \VSBQ is fast due to the (sparse) Bayesian quadrature. For \NNR, the runtime heavily depends on the size of the network and the effort spent on hyperparameter search (e.g., the weight decay). A larger network increases the representation flexibility but also introduces more computational overhead and potentially a higher risk of overfitting. Extensive hyperparameter search generally enhances regression performance but increases computational demand. The runtime for \LA depends on the computational cost associated with evaluating the log-likelihood function when computing the Hessian matrix.

\begin{table}[htbp]
\centering
\caption{\textbf{Algorithm runtime.} The wall-clock runtime for \VSBQ and \NNR is measured on both the CPU and GPU, whereas for \LA, only the CPU runtime for numerically computing the Hessian matrix is reported.} 
\label{supp_tab:runtimes}
\begin{tabular}{@{}llcc@{}}
\toprule
\textbf{Benchmark Task} & \textbf{Algorithm} & \textbf{CPU Runtime (s)} & \textbf{GPU Runtime (s)} \\ \midrule
Two moons                   & \VSBQ        & $164 \pm 6$                     & $110 \pm 5$                      \\
                          & \NNR        & $1935 \pm 148$                  &  $168 \pm 19$                     \\
                          & \LA        & $< 1$                   & \NA                      \\ \midrule
Multivariate                   & \VSBQ        & $1863 \pm 499$                      & $254 \pm 4$                      \\
Rosenbrock-Gaussian                          & \NNR      &  $18155 \pm 1740$                    & $638 \pm 161$                      \\
                          & \LA        & $< 1$                     & \NA                      \\ \midrule
Bayesian timing model                   & \VSBQ        & $538 \pm 66$                      & $198 \pm 4$                      \\
                          & \NNR        & $12287 \pm 2471$                  &  $451 \pm 147$                     \\
                          & \LA        & $38 \pm 0$                     & \NA                      \\ \midrule
Multisensory                   & \VSBQ        & $1167 \pm 178$                     & $260 \pm 3$                      \\
causal inference                          & \NNR        & $13745 \pm 2103$                    & $547 \pm 88$                    \\
                          & \LA        & $2 \pm 0$                     & \NA                      \\
                          \bottomrule
\end{tabular}
\end{table}

As shown in Table~\ref{supp_tab:runtimes}, \VSBQ takes several minutes on a CPU and benefits strongly from GPU acceleration, bringing post-process inference times down to 1-5 minutes. \NNR is slower compared to \VSBQ in the benchmark problems and also greatly benefits from GPU computation. In scenarios where the log-likelihood function is cheap to evaluate, \LA proves to be efficient and straightforward. Its additional requirement is that the target MAP point during optimization runs lies within the unbounded parameter space. Moreover, we recall that the Laplace approximation is not easily obtainable if only noisy likelihood evaluations are available.

Overall, \VSBQ remains efficient across the benchmark problems.
The wall-clock runtime efficiency makes it well-suited as a fast post-processing algorithm to compute the approximate posterior directly from the existing target posterior evaluations.

\subsection{Posterior estimation: MCMC or variational inference?}
\label{supp_sec:mcmc}
After having fitted a surrogate regression model (a sparse GP or a neural network) to the log-density function, the key issue is how to obtain an estimate of the posterior density (see Figure 1 in the main text). 
In the paper, we demonstrated how (stochastic) variational inference (\SVI) on the surrogate can successfully recover the posterior, based on prior work \citep{acerbi2018variational}.\footnote{In this paper, \SVI refers to stochastic variational inference with the reparameterization trick, including \ADVI.}
However, why not directly run \MCMC on the log-density surrogate to obtain approximate posterior samples, as done by other works \citep{rasmussen2003gaussian,nemeth2018merging, jarvenpaa2021parallel}?

The answer is that we attempted to use \MCMC (slice sampling; \citealp{neal2003slice}) and empirically found that, for both sparse GPs and neural networks, \MCMC often gives inferior results compared to \SVI. For example, Table~\ref{supp_tab:mcmc_vs_svi} shows the performance metrics for \SGPR with \SVI (i.e., \VSBQ), \SGPR with \MCMC, \NNR with \SVI, and \NNR with \MCMC, on the Bayesian timing model benchmark problem.  \MCMC is substantially less robust and yields worse results than \SVI.

\begin{table}[ht]
    \centering
    \caption{\textbf{Comparison between MCMC and SVI (Bayesian timing model, $\sigma_{\text{obs}}=3$).} For both \VSBQ and \NNR, MCMC performs poorly while SVI performs well in terms of metrics. For MCMC, the marginal likelihood estimate is not directly available.\label{supp_tab:mcmc_vs_svi}}
    \begin{tabular}{c c c c c}
        \toprule
        & \textbf{$\Delta$LML} ($\downarrow$) & \textbf{MMTV} ($\downarrow$) & \textbf{GsKL} ($\downarrow$) \\
    \toprule
\SGPR (\SVI) & 0.21 $\scriptstyle{[0.18, 0.22]}$  &  0.044 $\scriptstyle{[0.039, 0.049]}$ & 0.0065 $\scriptstyle{[0.0059, 0.0084]}$ \\ 
\SGPR (\MCMC) & \NA  & 0.69 $\scriptstyle{[0.057, 0.92]}$ &  5.2e+03 $\scriptstyle{[0.088, 5.8e+11]}$ \\ \hdashline
\NNR (\SVI) & 0.30 $\scriptstyle{[0.039, 0.44]}$  &  0.086 $\scriptstyle{[0.049, 0.12]}$ & 0.017 $\scriptstyle{[0.013, 0.076]}$ \\ 
\NNR (\MCMC) & \NA &  0.81 $\scriptstyle{[0.75, 0.89]}$ &  2.4e+11 $\scriptstyle{[2.1e+03, 1.2e+14]}$ \\ 
    \toprule
    \end{tabular}
\end{table}

\begin{figure}[ht]
    \centering
    \subcaptionbox{\SGPR (\MCMC) \label{supp_fig:noisy_timing_st_std-3_svbmc_mcmc_failure}}{\includegraphics[scale=0.9]{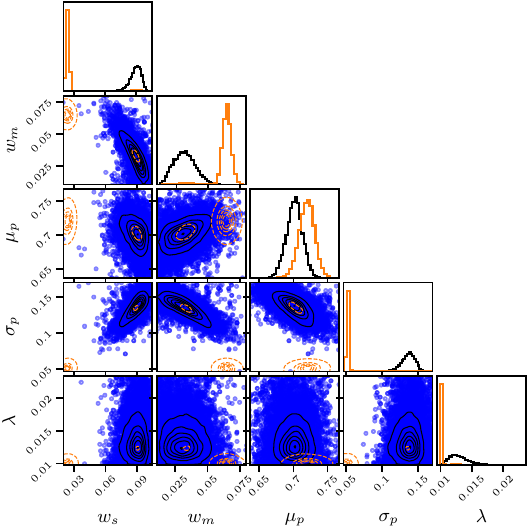}}
    \subcaptionbox{\NNR (\MCMC)\label{supp_fig:noisy_timing_st_std-3_nnr_mcmc_failure}}{\includegraphics[scale=0.9]{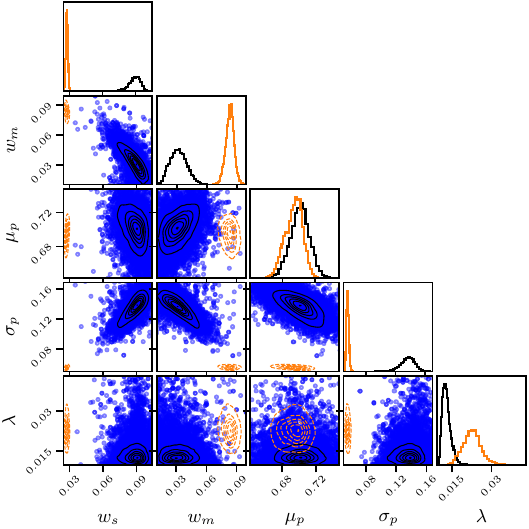}}
    \caption{\textbf{A typical failure with MCMC}. The benchmark problem is the Bayesian timing model ($\sigma_\textrm{obs} = 3$). The orange density contours in the sub-figures represent the posterior samples from \SGPR (\MCMC) and \NNR (\MCMC), while the black contours and points denote ground truth samples. The blue points are (the projection of) training points collected from the MAP optimization runs. In both shown cases, the \MCMC samples `escaped' to a region far from the training points, where the surrogate cannot be trusted.}
    \label{supp_fig:noisy_timing_st_std-3_mcmc_failure}
\end{figure}

We hypothesize that the reason why both the sparse GP and neural network surrogates can perform poorly with \MCMC is that, given the limited set of log-density observations, the surrogate model only approximates the log-density function well in a local region. Outside of this local ``trust region'', the surrogate may end up hallucinating \citep{desouza2022parallel}. Therefore, it is important to be careful not to leave the regions that contain actual observations. Without additional constraints, an \MCMC sampler can escape into the hallucinated regions and return meaningless samples. 
In support of our hypothesis, Figure~\ref{supp_fig:noisy_timing_st_std-3_mcmc_failure} shows a representative failure case where the samples from \MCMC are far from the log-density observations. 

Instead, for variational inference, initializing the variational distribution components around the high-density observation points and imposing an additional penalty loss in the \ELBO optimization for bounding the variational parameters (means and scales of the mixture components), help constrain the variational distribution in the local trust region, see Section~\ref{supp_sec:implementation_details} for details. As a further advantage, for a sparse GP, variational inference with Bayesian quadrature is efficient and affords computation of the \ELBO and its standard deviation, $\ELBO_\text{sd}$, for assessing its uncertainty.

\subsection{Note on package versions}
We implemented \VSBQ in Python, using JAX 0.4.20 \citep{jax2018github} and \NNR using PyTorch 2.1.1 ~\citep{paszkePyTorchImperativeStyle2019}.
We used the following package versions for all experiments in this paper: 
\begin{itemize}
    \item pycma 3.3.0 (\url{https://github.com/CMA-ES/pycma/releases/tag/r3.3.0}) for the \cmaes optimization algorithm;
    \item PyBADS 1.0.3 (\url{https://github.com/acerbilab/pybads/releases/tag/v1.0.3}) for the \bads optimization algorithm;
    \item emcee 3.1.4 (\url{https://github.com/dfm/emcee/releases/tag/v3.1.4}) for generating the ground-truth posterior samples.
\end{itemize}

\end{document}